\def\isarxiv{1} 

\ifdefined\isarxiv
\documentclass[11pt]{article}

\usepackage[numbers]{natbib}

\else
\documentclass{article}
\usepackage{neurips_2022}
\fi

\usepackage{amsmath}
\usepackage{amsthm}
\usepackage{amssymb}
\usepackage{algorithm}
\usepackage{subfig}
\usepackage{algpseudocode}
\usepackage{graphicx}
\usepackage{grffile}
\usepackage{wrapfig,epsfig}
\usepackage{url}
\usepackage{xcolor}
\usepackage{epstopdf}

\usepackage{bbm}
\usepackage{dsfont}

\allowdisplaybreaks

\ifdefined\isarxiv

\usepackage{tikz}
\usepackage{hyperref}  
\hypersetup{colorlinks=true,citecolor=blue,linkcolor=blue} 
\usetikzlibrary{arrows}
\usepackage[margin=1in]{geometry}

\else

\usepackage{microtype}
\usepackage{hyperref}
\definecolor{mydarkblue}{rgb}{0,0.08,0.45}
\hypersetup{colorlinks=true, citecolor=mydarkblue,linkcolor=mydarkblue}

\fi

\newtheorem{theorem}{Theorem}[section]
\newtheorem{lemma}[theorem]{Lemma}
\newtheorem{definition}[theorem]{Definition}

\newtheorem{corollary}[theorem]{Corollary}

\newtheorem{fact}[theorem]{Fact}

\newtheorem{claim}[theorem]{Claim}

\newcommand{\wh}{\widehat}
\newcommand{\wt}{\widetilde}

\newcommand{\R}{\mathbb{R}}
\newcommand{\A}{\mathsf{A}}
\newcommand{\B}{\mathsf{B}}

\renewcommand{\d}{\mathrm{d}}

\renewcommand{\hat}{\wh}

\DeclareMathOperator*{\Z}{\mathbb{Z}}

\DeclareMathOperator{\diag}{diag}

\DeclareMathOperator{\vect}{vec}

\DeclareMathOperator{\mat}{mat}

\DeclareMathOperator{\sparse}{sparse}
\DeclareMathOperator{\cent}{cent}
\DeclareMathOperator{\ent}{ent}

\makeatletter
\newcommand*{\RN}[1]{\expandafter\@slowromancap\romannumeral #1@}
\makeatother

\usepackage{lineno}

\begin{document}

\ifdefined\isarxiv

\date{}

\title{In-Context Learning for Attention Scheme: from Single Softmax Regression to Multiple Softmax Regression via a Tensor Trick}
\author{
Yeqi Gao\thanks{\texttt{a916755226@gmail.com}. The University of Washington.}
\and 
Zhao Song\thanks{\texttt{zsong@adobe.com}. Adobe Research.}
\and 
Shenghao Xie\thanks{\texttt{xsh1302@gmail.com}. The Chinese University of Hong Kong, Shenzhen.}
}

\else

\title{Intern Project} 
\maketitle 
\fi

\ifdefined\isarxiv
\begin{titlepage}
  \maketitle
  \begin{abstract}
 
Large language models (LLMs) have brought significant and transformative changes in human society. 
These models have demonstrated remarkable capabilities in natural language understanding and generation, leading to various advancements and impacts across several domains. 

We consider the in-context learning under two formulation for attention related regression in this work. Given matrices $A_1 \in \mathbb{R}^{n \times d}$, and $A_2 \in \R^{n \times d}$ and $B \in \R^{n \times n}$, the purpose is to solve some certain optimization problems,
\begin{itemize}
    \item Normalized version $\min_{X} \| D(X)^{-1} \exp(A_1 X A_2^\top) - B \|_F^2$
    \item Rescaled version $\| \exp(A_1 X A_2^\top) - D(X) \cdot B \|_F^2$
\end{itemize}
Here $D(X) := \mathrm{diag}( \exp(A_1 X A_2^\top) {\bf 1}_n )$.  
 
Our regression problem shares similarities with previous studies on softmax-related regression. Prior research has extensively investigated regression techniques related to softmax regression.
\begin{itemize}
    \item Normalized version $\| \langle \exp(Ax) , {\bf 1}_n \rangle^{-1} \exp(Ax) - b \|_2^2$ (see \cite{lsx+23})
    \item Resscaled version $\| \exp(Ax) - \langle \exp(Ax), {\bf 1}_n \rangle b \|_2^2 $ (see \cite{gsy23_hyper})
\end{itemize}

In contrast to previous approaches, we adopt a vectorization technique to address the regression problem in matrix formulation. This approach expands the dimension from $d$ to $d^2$, resembling the formulation of the regression problem mentioned earlier.

Upon completing the lipschitz analysis of our regression function, we have derived our main result concerning in-context learning.

  \end{abstract}
  \thispagestyle{empty}
\end{titlepage}

{\hypersetup{linkcolor=black}
\tableofcontents
}
\newpage

\else

\begin{abstract}
 
Large language models (LLMs) have brought significant and transformative changes in human society. 
These models have demonstrated remarkable capabilities in natural language understanding and generation, leading to various advancements and impacts across several domains. 

We consider the in-context learning under two formulation for attention related regression in this work. Given matrices $A_1 \in \mathbb{R}^{n \times d}$, and $A_2 \in \R^{n \times d}$ and $B \in \R^{n \times n}$, the purpose is to solve some certain optimization problems,
\begin{itemize}
    \item Normalized version $\min_{X} \| D(X)^{-1} \exp(A_1 X A_2^\top) - B \|_F^2$
    \item Rescaled version $\| \exp(A_1 X A_2^\top) - D(X) \cdot B \|_F^2$
\end{itemize}
Here $D(X) := \mathrm{diag}( \exp(A_1 X A_2^\top) {\bf 1}_n )$.  
 
Our regression problem shares similarities with previous studies on softmax-related regression. Prior research has extensively investigated regression techniques related to softmax regression.
\begin{itemize}
    \item Normalized version $\| \langle \exp(Ax) , {\bf 1}_n \rangle^{-1} \exp(Ax) - b \|_2^2$ (see \cite{lsx+23})
    \item Resscaled version $\| \exp(Ax) - \langle \exp(Ax), {\bf 1}_n \rangle b \|_2^2 $ (see \cite{gsy23_hyper})
\end{itemize}

In contrast to previous approaches, we adopt a vectorization technique to address the regression problem in matrix formulation. This approach expands the dimension from $d$ to $d^2$, resembling the formulation of the regression problem mentioned earlier.

Upon completing the lipschitz analysis of our regression function, we have derived our main result concerning in-context learning.
\end{abstract}

\fi

\section{Introduction}
Large Language Models (LLMs) such as GPT \cite{rwc+19,o23}, BERT \cite{dclt18}, RoBERTa \cite{log+19}, T5 \cite{rsr+20}, and XLNet \cite{ydy+19} have emerged as significant breakthroughs in natural language processing and artificial intelligence. These models, built on deep learning techniques and advanced architectures like Transformers \cite{vsp+17}, have revolutionized the field of language understanding and generation.

The success of Transformers \cite{vsp+17} in natural language processing has also spurred rapid advancements in the development of "in-context learning"  techniques.
Traditionally, natural language understanding models relied on isolated, context-independent representations of words or sentences \cite{hs97,g13}. However, Transformers introduced the concept of contextual embeddings, which capture the meaning and relationships of words within their surrounding context. This breakthrough enabled models to better understand the nuances and complexities of language.

The large-scale parameters significantly enhances the in-context learning ability of LLMs during both the training and inference stages. This improvement in in-context learning capability has led to successful applications in numerous fields. 
At the same time, in-context learning \cite{mlzh21,wbz+21,khsm22,onr+22,gtlv22,hsd+22} has revolutionized the way we approach various tasks in computer vision and natural language processing. By considering the contextual information and dependencies within a sequence or visual data, models can capture richer and more nuanced representations \cite{bmr+20}. This not only improves the performance of existing tasks but also opens up new possibilities for tackling more complex and challenging problems, which laid the foundation for the widespread application of Transformers \cite{vsp+17}.

The application of Transformers \cite{vsp+17,rns+18,pvu+18,dclt18,sll19,bmr+20,cms+20} in tasks like natural language queries \cite{tdb08,bmr+20,o23}, language translation services \cite{hwl21,izsn19,sll19}, and creative writing \cite{o23} showcased the power of in-context learning (ICL). By incorporating contextual information, models could generate more accurate and coherent translations, generate contextually relevant and fluent text, and better capture sentiment nuances. They have enriched our understanding and brought us closer to the concerns portrayed in books and movies, as many ideas that were once confined to science fiction now hold real-world significance. 

In spite of the success of transformers based on their ICL ability in various fields, a profound understanding of the ICL theory remains imperative for future research endeavors. This aspect is also a focal point of concern in our paper.
Indeed, a comprehensive understanding of in-context learning \cite{asa+22,gtlv22,onr+22,khsm22,zfb23,wzw23}, both in theory and practice, holds significant importance for future research. Our paper presents an analysis of in-context learning phenomenon about the regression problem in attention computation \cite{szks21,zkv+20,lsz+23}, an area extensively researched by some researchers. To aid readers' comprehension, we will introduce several works \cite{gms23,lsz23,ssz23,dls23,lsz+23,gsy23_hyper,zsz+23} that contribute to a better understanding of this topic.

\paragraph{Recent softmax regression}

A number of recent work have focused on the following three formulations: exponential regression \cite{gms23,lsz23}, the softmax regression \cite{dls23,lsx+23,ssz23,wyw+23,zsz+23} and the rescaled softmax regression \cite{gsy23_hyper},
\begin{definition}[]\label{def:intro_exponential_softmax_rescaled}
The objective is to work with a matrix $A \in \R^{n \times d}$ and $b \in \R^n$, aiming to achieve the following
\begin{itemize}
\item Exponential regression 
\begin{align*}
\min_{x \in \R^d} \| \exp(Ax) - b  \|_2^2
\end{align*}
\item Softmax regression
\begin{align*}
\min_{x \in \R^d} \| \langle \exp(Ax), {\bf 1}_n \rangle^{-1} \cdot \exp( A x ) - b \|_2^2
\end{align*} 
\item Rescaled regression
\begin{align*}
 \min_{x \in \R^d} \| \exp(Ax) - \langle \exp(Ax) , {\bf 1}_n \rangle \cdot b \|_2^2
\end{align*}
\end{itemize}
\end{definition}

Based on the aforementioned regression, our focus will be on the matrix formulation for attention regression as follows. Our paper will study the phenomenon of in-context learning based on it.
\paragraph{The matrix formulation for attention regression}

\begin{definition}[Normalized version]\label{def:intro_normalized_matrix}
Given a matrix $B \in \R^{n \times n}$ and matrices $A_1, A_2 \in \R^{n \times d}$, the optimization formulation is
\begin{align*}
\min_{X \in \R^{d \times d}} \| D(X)^{-1} \exp(A_1 X A_2^\top) - B \|_F^2
\end{align*}

Here diagonal matrix $D(X) \in \R^{n \times n}$ can be written as follows $D(X):= \diag( \exp(A_1 X A_2^\top) {\bf 1}_n )$.
\end{definition}

\begin{definition}[Rescaled version]\label{def:intro_rescaled_matrix}
Given a matrix $B \in \R^{n \times n}$ and matrices $A_1, A_2 \in \R^{n \times d}$,  the optimization task is
\begin{align*}
\min_{X \in \R^{d \times d}} \| \exp(A_1 X A_2^\top) - D(X) \cdot B \|_F^2.
\end{align*}
\end{definition}
To simplify our problem formulation, we denote the query matrix and key matrix as $A_1$ and $A_2$, respectively. In this context, $A_1$ represents the matrix $Q$, which contains the query vectors, while $A_2$ represents the matrix $K$, which contains the key vectors.
The techniques utilized as demonstrated in Section~\ref{sec:preli}, rely on vectorization to address the aforementioned problem.

\paragraph{Turning matrix formulation to vector formulation}

The above equation looks more complicated than the original equation, however, we can use a well-known tensor-trick \cite{swz19,dssw18,djs+19,swyz21,szz21,z22} to turn the multiple regression into a single regression with re-ordering/re-grouping all the entries.
\begin{definition}[Vector (equivalence) version of Definition~\ref{def:intro_normalized_matrix}]\label{def:intro_normalized_vector}
Given a matrix $B \in \R^{n \times n}$ and matrices $A_1, A_2 \in \R^{n \times d}$, let $\A = A_1 \otimes A_2$ and let $b = \vect(B)$, the formulation is
\begin{align*}
    \min_{x \in \R^{d^2}} \| D(x)^{-1} \exp(\A x) - b \|_2^2
\end{align*}
Here the diagonal matrix $D(x) \in \R^{n^2 \times n^2}$ can be written as $D(x) := D(X) \otimes I_n$
\end{definition}

\begin{definition}[Vector (equivalence) version of Definition~\ref{def:intro_rescaled_matrix}]\label{def:intro_rescaled_vector}
Given a matrix $B \in \R^{n \times n}$ and matrices $A_1, A_2 \in \R^{n \times d}$, let $\A = A_1 \otimes A_2$ and let $b = \vect(B)$, the formulation is
\begin{align*}
    \min_{x \in \R^{d^2}} \|  \exp(\A x) - D(x) \cdot b \|_2^2
\end{align*}
Here the diagonal matrix $D(x) \in \R^{n^2 \times n^2}$ can be written as $D(x) := D(X) \otimes I_n$
\end{definition}

\section{Our Results}
Our result concerns in-context learning phenomena based on the Lipschitz analysis of the regression problem mentioned earlier (see Definition~\ref{def:intro_normalized_matrix} and Definition~\ref{def:intro_rescaled_matrix}).

To provide readers with a better understanding of our results, we will offer an explanation of our theorem.
The $x \in \R^{d^2}$ here can be considered as the vectorization of $A_1$ and $A_2$. The training process can be viewed as updating $x$. Among in-context learning applications, an update on the gradient of $x$ can be likened to an update on tokenized document $\A$. This implies a high similarity between Transformers and gradient descent.

Our analysis of in-context learning phenomena can be divided into two parts: the normalized version in Section~\ref{sec:our_result:rescaled}  and the rescaled version in Section~\ref{sec:our_result:rescaled}. For the purpose of assisting future research, we present additional results on Lipschitz analysis for some common loss functions in Section~\ref{sec:our_result:gradient}.

\subsection{Normalized Version}\label{sec:our_result:normalized}
\begin{theorem}[Learning in-context for Normalized Version]\label{thm:in_context_normalized:informal}
Provided that the subsequent requirement are satisfied
\begin{itemize}
      \item Let $R \geq 4$.
    \item Let $x_t \in \R^{d^2}, x_{t+1} \in \R^{d^2}$ satisfy $\| x_{t} \|_2 \leq R$ and $\| x_{t+1} \|_2 \leq R$
    \item Let $\A \in \R^{n^2 \times d^2}$
    \item Let $\max_{j_1 \in [n]} \| \A_{[j_1],*} (x_{t}-x_{t+1}) \|_{\infty} < 0.01$
    \item Let $\max_{j_1 \in [n]}\| \A_{[j_1],*} \| \leq R$
    \item Let $\max_{j_1 \in [n]} \| b_{[j_1]} \|_2 \leq 1$
    \item Let $M = \exp(O(R^2 + \log n))$.
\end{itemize}
We consider the matrix formulation for attention regression (Definition~\ref{def:intro_normalized_matrix}) 
\begin{align*}
     \min_{x \in \R^{d^2}} \|  D(x)^{-1}\exp(\A x) -  b \|_2^2.
\end{align*}
\begin{itemize}
\item {\bf Part 1.} 
By transitioning $x_t$ to $x_{t+1}$, we are effectively addressing a fresh normalized softmax regression problem involving
\begin{align*}
      \min_{x \in \R^{d^2}} \| D(x)^{-1} \exp(\A x) -  \wt{b} \|_2^2
\end{align*}
where 
\begin{align*}
    \|  b -\wt{b} \|_2 \leq M \cdot \| x_{t+1} - x_t \|_2
\end{align*}
\item {\bf Part 2.} 
By transitioning $A_t$ to $A_{t+1}$, we are effectively addressing a fresh normalized softmax regression problem involving
\begin{align*}
       \min_{x \in \R^{d^2}} \|D(x)^{-1} \exp(\A x) -  \hat{b} \|_2^2
\end{align*}
where 
\begin{align*}
    \|  b -\wh{b} \|_2 \leq M \cdot \| A_{t+1} - A_t \|
\end{align*}
\end{itemize}
\end{theorem}

See details in Section~\ref{sec:application_normalized}.

\subsection{Rescaled Version}\label{sec:our_result:rescaled}
\begin{theorem}[Learning in-context for Rescaled Version]\label{thm:in_context_rescaled:informal}
Provided that the subsequent requirement are satisfied
\begin{itemize}
     \item Let $R \geq 4$.
    \item Let $x_t \in \R^{d^2}, x_{t+1} \in \R^{d^2}$ satisfy $\| x_{t} \|_2 \leq R$ and $\| x_{t+1} \|_2 \leq R$
    \item Let $\A \in \R^{n^2 \times d^2}$
    \item Let $\max_{j_1 \in [n]} \| \A_{[j_1],*} (x_{t}-x_{t+1}) \|_{\infty} < 0.01$
    \item Let $\max_{j_1 \in [n]}\| \A_{[j_1],*} \| \leq R$
    \item Let $\max_{j_1 \in [n]} \| b_{[j_1]} \|_2 \leq 1$
    \item Let $M = \exp(O(R^2 + \log n))$.
\end{itemize}
We consider the matrix formulation for attention regression (Definition~\ref{def:intro_rescaled_matrix}) problem
\begin{align*}
     \min_{x \in \R^{d^2}} \|  \exp(\A x) - D(x) b \|_2^2.
\end{align*}
\begin{itemize}
\item {\bf Part 1.} 
By transitioning $x_t$ to $x_{t+1}$, we are effectively addressing a fresh rescaled softmax regression problem involving
\begin{align*}
      \min_{x \in \R^{d^2}} \|  \exp(\A x) - D(x) \wt{b} \|_2^2
\end{align*}
where 
\begin{align*}
    \| b -  \wt{b} \|_2 \leq M \cdot \| x_{t+1} - x_t \|_2
\end{align*}
\item {\bf Part 2.} By transitioning $A_t$ to $A_{t+1}$, we are effectively addressing a fresh rescaled softmax regression problem involving
\begin{align*}
       \min_{x \in \R^{d^2}} \|\exp(\A x) -  D(x)\hat{b} \|_2^2
\end{align*}
where 
\begin{align*}
    \| b - \wh{b} \|_2 \leq M \cdot \| A_{t+1} - A_t \|
\end{align*}
\end{itemize}
\end{theorem}
See details in  Section~\ref{sec:application_rescaled}.

\subsection{Lipschitz of Gradient} \label{sec:our_result:gradient}

We have derived additional findings as follows:
\begin{corollary} 
Provided that the subsequent requirement are satisfied
\begin{itemize}
    \item Let $\A, \B \in \R^{n^2 \times d^2}$ satisfy $\max_{j_1 \in [n]}\| \A_{[j_1],*} \| \leq R$, $\max_{j_1 \in [n]} \| \B_{[j_1],*} \| \leq R$ \item Let $\max_{j_1 \in [n]} \| ( \A_{[j_1],*} - \B_{[j_1],*} ) x \|_{\infty} < 0.01$
    \item Let $x, y \in \R^{d^2}$ satisfy that $\| x \|_2 \leq R $, $\| y \|_2 \leq R$
    \item Let $\max{j_1 \in [n]} \| \A_{[j_1],*} (x - y) \|_\infty < 0.01$
    \item Let $\max_{j_1 \in [n]} \| b_{[j_1]} \|_2 \leq 1$
    \item $\| {\bf 1}_n - \frac{f(x)_{j_1}}{f(y)_{j_1}} \|_\infty \leq 0.1$
    \item $\| {\bf 1}_n - \frac{f(\A)_{j_1}}{f(\B)_{j_1}} \|_\infty \leq 0.1$
    \item $R > 4$ 
    \item Let $\| \A - \B \|_{\infty,2} = \max_{j_1 \in [n]} \| \A_{[j_1],*} - \B_{[j_1],*} \| $ 
    \item Let $L_c(\A)$ be defined as Definition~\ref{def:L_c} and $L_q(\A)$ be defined as Definition~\ref{def:L_q}
    \item Let $L_{\sparse}$, $L_{\cent}$ and $L_{\ent}$ be defined in Definition~\ref{def:L_sparse}, Definition~\ref{def:L_cent} and Definition~\ref{def:L_ent} respectively.
    \item Let $M_0 := \exp(O(R^2 +\log(nd))$
\end{itemize}
it follows that
\begin{itemize}
\item Normalize version Loss (See Details in Section~\ref{sec:Lipschitz_L_c}, informal version of Lemma~\ref{lem:lipschitz_grad_L_c:x} and Lemma~\ref{lem:lipschitz_grad_L_c:A})
\begin{itemize}
\item $\| \nabla L_c(x) - \nabla L_c(y) \|_2 \leq M_0 \cdot \| x - y \|_2$
\item $
    \| \nabla L_c(\A) - \nabla L_c(\B) \|_2 \leq M_0 \cdot \| \A - \B \|_{\infty,2}$
\end{itemize}
\item Rescaled version Loss (See Details in Section~\ref{sec:Lipschitz_L_q}, informal version of Lemma~\ref{lem:lipschitz_grad_L_q:x} and Lemma~\ref{lem:lipschitz_grad_L_q:A})
\begin{itemize}
\item $\| \nabla L_q(x) - \nabla L_q(y) \|_2 \leq M_0 \cdot \| x - y \|_2$
\item $ \| \nabla L_q(\A) - \nabla L_q(\B) \|_2 \leq M_0 \cdot \| \A - \B \|_{\infty,2}$
\end{itemize}
\end{itemize}
\begin{itemize}
    \item {
    Sparse Loss (See Details in Section~\ref{sec:lipschitz_L_sparse}, informal version of Lemma~\ref{lem:lipschitz_grad_L_sparse:x} and Lemma~\ref{lem:lipschitz_grad_L_sparse:A})
    \begin{itemize}
        \item $| \nabla L_{\sparse}(x) - \nabla L_{\sparse}(y) | \leq M_0 \cdot \| x - y \|_2$
        \item  $|\nabla L_{\sparse}(\A) - \nabla L_{\sparse}(\B) | \leq M_0 \cdot \| \A - \B \|_{\infty,2}$
    \end{itemize}
    }
    \item {Cross Entropy Loss (See Details in Section~\ref{sec:lipschitz_L_cent}, informal version of Lemma~\ref{lem:lipschitz_grad_L_cent:x} and Lemma~\ref{lem:lipschitz_grad_L_cent:A})
    \begin{itemize}
        \item $\| \nabla L_{\cent}(x) - \nabla L_{\cent}(y) \|_2
    \leq M_0 \cdot \| x - y \|_2$
        \item $\| \nabla L_{\cent}(\A) - \nabla L_{\cent}(\B) \|_2 \leq M_0 \cdot \| \A - \B \|_{\infty,2}$
    \end{itemize}
    
    }
    \item  {
    Entropy Loss (See Details in Section~\ref{sec:lipschitz_L_ent}, informal version of Lemma~\ref{lem:lipschitz_grad_L_ent:x} and Lemma~\ref{lem:lipschitz_grad_L_ent:A})
    \begin{itemize}
        \item $\| \nabla L_{\ent}(x) - \nabla L_{\ent}(y) \|_2 \leq M_0 \cdot \| x - y \|_2$
        \item $\| \nabla L_{\ent}(\A) - \nabla L_{\ent}(\B) \|_2 \leq M_0 \cdot \| \A - \B \|_{\infty,2}$
    \end{itemize}
    }
    
\end{itemize}
\end{corollary}

\paragraph{Roadmap.}
In Section~\ref{sec:preli}, we cite several lemmas from matrix algebra. In Section~\ref{sec:grad}, we define some functions discussed. We prove the equivalence between matrix version and vector version of softmax regression formulation, and we calculate the gradients of Loss functions. In Section~\ref{sec:upper_bounds}, we state some bounds for auxiliary functions. In Section~\ref{sec:lipschitz_basic_fcts}, we calculate the Lipschitz for auxiliary functions. In Section~\ref{sec:Lipschitz_L_c}, we discuss the Lipschitz condition for softmax loss function in normalized version. In Section~\ref{sec:Lipschitz_L_q}, we discuss the Lipschitz condition for softmax loss function in rescaled version. In Section~\ref{sec:lipschitz_L_sparse}, we discuss the Lipschitz condition for sparse loss function. In Section~\ref{sec:lipschitz_L_cent}, we discuss the Lipschitz condition for cross entropy loss function. In Section~\ref{sec:lipschitz_L_ent}, we discuss the Lipschitz condition for entropy loss function. In Section~\ref{sec:application_rescaled} and Section~\ref{sec:application_normalized},  
a significant contribution lies in the application of Lipshcitz analysis to both the rescaled and normalized versions of regression in the field of in-context learning.  

\section{Related Work}
\paragraph{In-Context Learning}
In in-context learning, deep learning models consider the contextual information of input data during training and inference, enabling them to better understand and leverage the relevant relationships within the data. This means that models learn and make decisions based not only on individual samples but also by considering the surrounding context or related tasks.
\cite{asa+22} was demonstrated that transformers utilized as in-context learners can implicitly execute conventional learning algorithms. 
This accomplishment is attained by integrating smaller models into their activations. Models are updated consistently when new examples emerge in the specified context.

\cite{gtlv22} is centered around gaining a comprehensive understanding of in-context learning. They achieve this by exploring tasks in which models are trained under specific in-context requirements to learn specific classes of functions (linear functions). 
The central question they prioritize is whether the model can be trained to master the most of functions within that class, even when provided with data obtained from other functions belonging to the same class.

Transformers' mechanisms are  perceived as in-context according to  \cite{onr+22}. 
There are similarities between in-context tasks and meta-learning formulations that rely on gradient descent.
This suggests that transformers can leverage meta-learning principles to adapt and enhance their performance in in-context learning scenarios. Meta-learning \cite{ks21}, or learning-to-learn, aims to automate the learning of these aspects, reducing the need for manual effort and unlocking greater capabilities. \cite{khsm22} demonstrates that Transformers and other black-box models can be meta-trained to function as general-purpose in-context learners.

Based on transformers trained through gradient flow on linear regression tasks with a single linear self-attention layer, \cite{zfb23} conduct a comprehensive investigation into the dynamics of ICL. The study aims to explore why transformers, when applied to random instances of linear regression problems, exhibit similarities with the predictions of ordinary least squares models.
\cite{wzw23} demonstrates that ICL can be regarded as a procedure of Bayesian selection, enabling it to implicitly infer information relevant to the designated tasks.
\paragraph{Transformer Theory}
\cite{szks21} investigated the behavior about single-head attention mechanism in Seq2Seq model learning and provided instructions on parameter selection for better performance.
The research by \cite{szks21} serves as a inspiration for parameter selection in Seq2Seq model learning. Their investigation specifically focuses on the single-head attention mechanism.
\cite{zkv+20} conducted an investigation into adaptive methods and proposed a specific approach for attention models.
For regularized exponential regression, \cite{lsz23} proposed an algorithm operates in input sparsity time. Its effectiveness has been demonstrated across any datasets.

When exponential activation functions were utilized, \cite{gms23} examined the convergence of over-parameterized neural networks 
An extensive elucidation was presented by \cite{llr23} on how transformers can acquire the "semantic structure" required for identifying patterns of word co-occurrence.
Optimization techniques used in transformers and their weakness and strengths are discussed.

Some studies are dedicated to fast computation methods for attention. \cite{bsz23} examined the dynamic attention computation and presented both favorable and unfavorable outcomes. 
The effectiveness of static attention algorithms across diverse applications is explored by \cite{zhdk23}.
\cite{as23} proposed an algorithm for static attention and a hardness result is presented based on the exponential time hypothesis.
\cite{dms23} put forth both deterministic and randomized algorithms while studying the sparsification of feature dimensions in attention computation. 

\section{Preliminary}\label{sec:preli}
In this section, we discuss the notations and mathematical tools used in this paper. The notations are explained in Section~\ref{sec:preli:notations}, while the mathematical facts and tools are discussed in Section~\ref{sec:preli:facts}.
\subsection{Notations.}\label{sec:preli:notations} 

Given that $x \in \R^n$, $\exp(x) \in \R^n$ denotes the vector that $\exp(x)_i = \exp(x_i)$. $\log(x) \in \R^n$ is used to denote the vector that $\log(x)_i = \log(x_i)$. For all $i \in [n]$, $x^{-1} \in \R^n$ is used to denote the vector $y \in \R^n$ such that $y_i = x_i^{-1}$.

For a matrix $A \in \R^{n \times d}$, $\exp(A) \in \R^{n \times d}$ denotes the matrix where $\exp(A)_{i,j} = \exp(A_{i,j})$.

Let $x \in \R^n$ be a vector, let $t$ be a scalar, we define $\frac{\d x}{\d t} \in \R^n$ to be a vector that $i$-th entry $\frac{ \d x_i }{\d t}$.

Let $f: \R^d \to \R^n $, let $t$ be a scalar, $\frac{\d f}{\d t} \in \R^n$ denotes a vector whose $i$-th entry is $\frac{\d f_i}{\d t}$.

For a matrix $A \in \R^{n \times d}$, we use $\| A \|$ to denote its spectral norm, i.e., $\| A \|:= \max_{x \in \R^d } \| A x \|_2 / \| x \|_2$

Let $x, ~y$ be two vectors $\in \R^n$, let $\langle x,y \rangle$ be $\sum_{i=1}^n x_i \cdot y_i$.

Let $x, ~y$ be two vectors $\in \R^n$, let $x \circ y := z \in \R^n$ where $z_i = x_i \cdot y_i$.

Let ${\bf 1}_n$ denotes a vector with n entries and all of them are $1$.

Let $n \in \Z_+$, $[n]$ denotes a set $\{1,2, \cdots , n\}$.

Let $x \in \R^{n^2}$ be a vector, let $X \in \R^{n \times n}$ be a matrix, we say $x = \vect(X)$ if the $i$-th row of $X$ is equivalent to the $(i-1)n+1$-th term to the $in$-th term of $x$ for all $i \in [n]$. Symmetrically, $X = \mat(x)$.

For matrix $A \in \R^{n_1 \times d_1}$ and a matrix $B \in \R^{n_2 \times d_2}$, $A\otimes B \in \R^{n_1 n_2 \times d_1 d_2}$ denotes a new matrix that $(i_1 - 1) n_2 + i_2$, $(j_1-1)d_2+j_2$-th entry is $A_{i_1,j_1} B_{i_2,j_2}$, where $i_1 \in [n_1], j_1 \in [d_1], i_2 \in [n_2], j_2 \in [d_2]$.

\subsection{Facts}\label{sec:preli:facts}

\begin{fact}[Basic vector properties]\label{fac:vector_properties}

    Provided that the subsequent requirement are satisfied
    \begin{itemize}
        \item Let $d \in \Z_+$.
        \item Let $x, y, z \in \R^d$.
        \item Let $a, b \in \R$.
    \end{itemize}
    it follows that
    \begin{itemize}
        \item Part 1. $\langle x, y \rangle = \langle x \circ y, {\bf 1}_n \rangle$.
        \item Part 2. $a\langle x, z \rangle + b\langle y, z \rangle = \langle ax + by, z \rangle = \langle z, ax + by \rangle = a\langle z, x \rangle + b\langle z, y \rangle$.
    \end{itemize}
    
\end{fact}

\begin{fact}[Basic derivative rules]\label{fac:derivative_rules}

    Provided that the subsequent requirement are satisfied
    \begin{itemize}
        \item Let $n, d \in \Z_+$ and $k \in \Z$.
        \item Let $x \in \R^d$ be a vector.
        \item Let $t \in \R$ be a scalar. 
        \item Let $c$ be independent of $t$.
        \item Let $f: \R^d \to \R^n$.
        \item Let $h: \R^d \to \R^n$.
        \item Let $g: \R^d \rightarrow \R$.
    \end{itemize}
    it follows that    
    \begin{itemize}
        \item Part 1. $\frac{\d (c \cdot f(x))}{\d t} = c \cdot \frac{\d f(x)}{\d t}$ (constant multiple rule).
        \item Part 2. $\frac{\d (g(x)^k)}{\d t} = k \cdot g(x)^{k - 1} \cdot \frac{\d g(x)}{\d t}$ (power rule). 
        \item Part 3. $\frac{\d (h(x) + f(x))}{\d t} = \frac{\d h(x)}{\d t} + \frac{\d f(x)}{\d t}$ (sum rule).
        \item Part 4. $\frac{\d (h(x) \circ f(x))}{\d t} = \frac{\d h(x)}{\d t} \circ f(x) + h(x) \circ \frac{\d f(x)}{\d t}$ (product rule for Hadamard product).
    \end{itemize}
    
\end{fact}

\begin{fact}\label{fac:vector_norm}
For vectors $x,y \in \R^n$, we have
\begin{itemize}
    \item $\| x \circ y \|_2 \leq \| x \|_{\infty} \cdot \| y \|_2$
    \item $\| x \|_{\infty} \leq \| x \|_2 \leq \sqrt{n} \| x \|_{\infty}$
    \item $\| \exp(x) \|_{\infty} \leq \exp(\| x \|_2)$ 
    \item $\| \exp(x) \|_2 \cdot 2 \| x - y \|_{\infty} \geq \| \exp(x) - \exp(y) \|_2$, for any $\| x - y \|_{\infty} \leq 0.01$
\end{itemize}
\end{fact}

\begin{fact}\label{fac:matrix_norm}
For matrices $X,Y$, we have 
\begin{itemize}
    \item $\| X^\top \| = \| X \|$ 
    \item $\| X \| \geq \| Y \| - \| X - Y \|$
    \item $\| X + Y \| \leq \| X \| + \| Y \|$
    \item $\| X \cdot Y \| \leq \| X \| \cdot \| Y \|$ 
    \item If $X \preceq \alpha \cdot Y$, then $\| X \| \leq \alpha \cdot \| Y \|$
\end{itemize}
\end{fact}

\section{Gradient} \label{sec:grad}
In Section~\ref{sec:def}, we provide an overview of some  concepts used for gradient computation, with a specific emphasis on the fundamental idea in this paper regarding vectorization for matrices.
Section~\ref{sec:grad:basic_equivalence} focuses on basic equivalence, while Section~\ref{sec:gra} presents the derivatives.
\subsection{Preliminary}\label{sec:def}

\begin{definition}
Given $A_1 \in \R^{n \times d}$, $A_2 \in \R^{n \times d}$, we define $\mathsf{A} \in \R^{n^2 \times d^2}$ to be the matrix $A_1 \otimes A_2$, where the $(i_1-1) \cdot n + i_2$-th row is  
\begin{align*}
 \underbrace{ \mathsf{A}_{(i_1-1)n + i_2, *} }_{1 \times d^2} :=  \underbrace{ A_{1,i_1,*} }_{1 \times d} \otimes \underbrace{ A_{2,i_2,*} }_{1 \times d}
\end{align*}
for all $i_1 \in [n]$ and $i_2 \in [n]$. Here $A_{1,i_1,*}$ is the $i_1$-th row of matrix $A_1$.
\end{definition}

\begin{definition}\label{def:dX}
Given $A_1, A_2 \in \R^{n \times d}$ and $X \in \R^{d \times d}$, $D(X) \in \R^{n \times n}$ is defined as outlined below
\begin{align*}
    D(X) := \diag( \exp(A_1 X A_2^\top) {\bf 1}_n ).
\end{align*} 

Note that $X \in \R^{d \times d}$ is matrix version of vector $x \in \R^{d^2 \times 1}$, i.e., $X = \mat(x)$.
\end{definition}

\begin{definition}\label{def:dx}
Given matrices $A_1 \in \R^{n \times d}$, $A_2 \in \R^{n \times d}$ and $x \in \R^{d^2 \times 1}$.
Let $D(X) \in \R^{n \times n}$ is defined in Definition~\ref{def:dX}.
We define diagonal matrix $D(x) \in \R^{n^2 \times n^2}$ as outlined below
\begin{align*}
    D(x)_{ (i_1-1) n + i_2, (i_1-1) \cdot n + i_2 } := \exp( A_{1,i_1,*} X A_2^\top ) {\bf 1}_n
\end{align*} 
In other words, $D(x) = D(X) \otimes I_n$. Here $x$ is the vectorization of matrix $X$, i.e., $x = \vect(X)$.
\end{definition}

\begin{definition}[Function $\alpha(x)$]\label{def:alpha}
We also define $\alpha(x) \in \R^n$
\begin{align*}
    \alpha(x)_{j_1}:= \langle \exp( \mathsf{A}_{[j_1],*} x ) , {\bf 1}_n \rangle, ~~~\forall j_1 \in [n]
\end{align*}
 Here $\mathsf{A}_{[j_1],*} \in \R^{n \times d^2}$ denotes the rows from index $(j_1-1) \cdot n+1$ to index $j_1 \cdot n$.
\end{definition}

\begin{definition}[Function $u(x)$]\label{def:u}
For each $j_1 \in [n]$, $u(x)_{j_1} \in \R^n$ is defined as outlined below
\begin{align*}
    u(x)_{j_1} := \exp( \A_{[j_1],*} x )
\end{align*}
\end{definition}

\begin{definition}[Function $f(x)$]\label{def:f}
For each $j_1 \in [n]$, we define $f(x)_{j_1} \in \R^n$ as outlined below
\begin{align*}
    f(x)_{j_1} := \alpha(x)_{j_1}^{-1} \cdot u(x)_{j_1}.
\end{align*}
\end{definition}

\begin{definition}[Function $h(x)$]\label{def:h}
For each $j_1 \in [n]$, we define $h(x)_{j_1} \in \R^n$ as outlined below
\begin{align*}
    h(x)_{j_1} := \log(f(x)_{j_1}).
\end{align*}
\end{definition}

\begin{definition}[Function $c(x)$]\label{def:c}
For each $j_1 \in [n]$, we define $c(x)_{j_1} \in \R^n$ as outlined below
\begin{align*}
    c(x)_{j_1} := f(x)_{j_1} - b_{[j_1]} .
\end{align*} 
Here we use $b_{[j_1]} \in \R^{n}$ to denote the $(j_1-1)\cdot n+1)$-th to $(j_1 \cdot n)$-th entry of $b \in \R^{n^2}$.
\end{definition}

\begin{definition}[Function $q(x)$]\label{def:q}
For each $j_1 \in [n]$, we define $q(x)_{j_1} \in \R^n$ as outlined below
\begin{align*}
    q(x)_{j_1} := u(x)_{j_1} - b_{[j_1]} \cdot \alpha(x)_{j_1} 
\end{align*} 
Here we use $b_{[j_1]} \in \R^{n}$ to denote the $(j_1-1)\cdot n+1$-th to $(j_1 \cdot n)$-th entry of $b \in \R^{n^2}$.
\end{definition}

\begin{definition}[Loss Function $L_c(x)$] \label{def:L_c}
We define the normalized softmax version loss $L_{c}$ as outlined below
\begin{align*}
    L_{c}(x):= 0.5 \sum_{j_1=1}^n \| c(x)_{j_1} \|_2^2.
\end{align*}  
\end{definition}

\begin{definition}[Loss Function $L_q(x)$]\label{def:L_q}
We define the rescaled softmax version loss $L_q(x)$ as outlined below 
\begin{align*}
    L_q(x) := 0.5 \sum_{j_1=1}^n \| q(x)_{j_1} \|_2^2
\end{align*}
\end{definition}

\begin{definition}[Sparse Penalty, a generalization of Definition D.41 in \cite{zsz+23}]\label{def:L_sparse}

We define
\begin{align*}
L_{\sparse} (x) := \sum_{j_1=1}^n \alpha(x)_{j_1}
\end{align*}

\end{definition}

\begin{definition}[Cross entropy, a generalization of Definition C.7 in \cite{wyw+23}]\label{def:L_cent}

We define
\begin{align*}
L_{\cent} (x) :=  - \sum_{j_1=1}^n \langle f(x)_{j_1}, b_{[j_1]} \rangle
\end{align*}
\end{definition}

\begin{definition}[Entropy, a generalization of Definition 4.7 in \cite{ssz23}]\label{def:L_ent}

We define
\begin{align*}
L_{\ent} (x) :=  - \sum_{j_1=1}^n \langle f(x)_{j_1}, \log f(x)_{j_1} \rangle
\end{align*}
\end{definition}

\subsection{Basic Equivalence}\label{sec:grad:basic_equivalence}

\begin{lemma} \label{lem:basic_equivalence_tools}
Given $A = \diag(a_1, \cdots, a_n)$, a matrix $B \in \R^{n \times n}$ and $I_n \in \R^{n \times n}$ be the identity matrix, we have
\begin{align*}
\vect(AB) = A \otimes I_n \vect(B).
\end{align*}
\end{lemma}

\begin{proof}
     It simply follows that $((i-1)n+j)$-th entry of $\vect(AB)$, which is $a_ib_{ij}$, is equal to the corresponding entry of $A \otimes I_n \vect(B)$.
\end{proof}

For help the proof of Claim~\ref{cla:vectorization}, we propose the following definitions.
\begin{definition}\label{def:index_tensor}
    We have 
    \begin{itemize}
        \item We use $a_{ij}$ to denote the $i,j$-th entry of matrix $A_1$. 
        \item We use $A_i$ to denote the $i$-th row of matrix $A_2$.
        \item We use $X_j$ to denote the $j$-th row of matrix $X$, where $i \in [n]$, $j \in [d]$.
    \end{itemize}
\end{definition}

\begin{claim}\label{cla:vectorization}
Provided that the subsequent requirement are satisfied
\begin{itemize}
    \item Let $b \in \R^{n^2}$ denote the vectorization of $B \in \R^{n \times n}$, i.e., $b = \vect(B)$.
\item  Let $a_{ij}$, $A_i$ and $X_j$ be defined in Definition~\ref{def:index_tensor}.
\end{itemize}
It follows that
\begin{itemize}
\item Part 1.
\begin{align*}
\vect( \underbrace{ A_1 }_{n \times d} \underbrace{ X }_{d \times d} \underbrace{ A_2^\top }_{d \times n} ) = \underbrace{ ( A_1 \otimes A_2) }_{n^2 \times d^2} \underbrace{ \vect(X) }_{d^2 \times 1}
\end{align*}
\item Part 2. $\min_{X \in \R^{d \times d}} \| A_1 X A_2^\top - B \|_F^2$ is equivalent to $\min_{x \in \R^{d^2}} \| (A_1 \otimes A_2) x - b \|_2^2$. 
\item Part 3. $\min_{X \in \R^{d \times d}} \| \exp( A_1 X A_2^\top ) - B \|_F^2$ is equivalent to $\min_{x \in \R^{d^2}} \| \exp( (A_1 \otimes A_2^\top ) x ) - b \|_2^2$.
\item Part 4. $\min_{X \in \R^{d \times d}} \| D(X)^{-1} \exp( A_1 X A_2^\top ) - B \|_F^2$ is equivalent to $\min_{x \in \R^{d^2}} \| D(x)^{-1} \exp( (A_1 \otimes A_2) x ) - b \|_2^2$.
\begin{itemize}
    \item The diagonal matrix $D(X) \in \R^{n \times n}$ is defined as $D(X) := \diag( \exp(A_1 X A_2^\top) {\bf 1}_n )$
    \item The diagonal matrix $D(x) \in \R^{n^2 \times n^2}$ is be written as $ D(x)= D(X) \otimes I_n$.
\end{itemize}
\end{itemize}
\end{claim}

\begin{proof}
Our proof is as outlined below.
\paragraph{Proof of Part 1.}
By directly calculating the matrices, we can conclude that
\begin{align*}
    (A_1 X A_2^\top)_{i,j} = & ~ \sum_{k=1}^d a_{ik} X_k A_j^\top \\
    = & ~ ((A_1 \otimes A_2) \vect(X))_{i(d-1)+j}
\end{align*}

Therefore, {\bf Part 1} holds.

\paragraph{Proof of Part 2.}
We have
\begin{align*}
\vect(A_1 X A_2^\top - B) 
= & ~ \vect(A_1 X A_2^\top x) - b\\
= & ~ (A_1 \otimes A_2) x - b
\end{align*}
where the 1st step is due to $\vect(B) = b$ and the 2nd step is because of {\bf Part 1} above.

Hence,  by the definition of Frobenius norm and Euclidean norm, we have
\begin{align*}
    \| A_1 X A_2^\top - B \|_F^2 = \| (A_1 \otimes A_2) x - b \|_2^2
\end{align*}

Therefore, the equivalence is verified.

\paragraph{Proof of Part 3.}

Similar to {\bf Part 2}, the result can be derived from the equation $\vect(\exp(A_1 X A_2^\top) - B) = \exp((A_1 \otimes A_2) x) - b$.

\paragraph{Proof of Part 4.}
We have 
\begin{align}\label{eq:d_x_I}
    D(x)^{-1} = & ~ (D(X) \otimes I_n)^{-1} \notag \\ 
    = & ~ D(X)^{-1} \otimes I_n
\end{align}
According to Lemma~\ref{lem:basic_equivalence_tools}, it follows that
\begin{align*}
    \vect(D(X)^{-1} \exp(A_1 X A_2^\top)) 
    = & ~ D(x)^{-1} \otimes I_n \vect(\exp(A_1 X A_2^\top)) \\
     = & ~ D(x)^{-1} \vect(\exp(A_1 X A_2^\top)) \\
     = & ~ D(x)^{-1} \exp((A_1 \otimes A_2)x) 
\end{align*}
where the 1st step is based on  Lemma~\ref{lem:basic_equivalence_tools}, the 2nd step follows from Eq.~\eqref{eq:d_x_I} and the 3rd step is because of  {\bf Part 1}.

Then, the proof is complete.
\end{proof}

\subsection{Basic Derivatives}\label{sec:gra}
\begin{lemma}\label{lem:basic_derivatives}
Provided that the subsequent requirement are satisfied
\begin{itemize}
    \item Let $A_1, A_2 \in \R^{n \times d}$.
    \item Let $\mathsf{A} = A_1 \otimes A_2$.
    \item Let $X \in \R^{d \times d}$.
    \item Let $i = (i_1-1) d + i_2$
    \item For each $i \in [d^2]$, we use $\A_{*,i} \in \R^{n^2}$ to denote the $i$-th column of $\A \in \R^{n^2 \times d^2}$
    \item Let $D(x)$ be defined as in Definition~\ref{def:dx}.
    \item Let $x \in \R^{d^2}$ and $x = \vect(X)$.
    \item $\mathsf{A}_{[j_1],*} \in \R^{n \times d^2}$ be the matrix formed by the rows from index $(j_1-1) \cdot n+1$ to index $j_1 \cdot n$ of matrix $\A \in \R^{n^2 \times d^2}$.
    \item Let $\A_{[j_1],i}$ be the $i$-th column of matrix $\A_{[j_1],*}$.
\end{itemize}
Then, we can show
\begin{itemize}
\item Part 1. For each $i \in [d^2]$
\begin{align*}
    \frac{\d  \mathsf{A} x}{\d x_i} = \A_{*,i} 
\end{align*} 
\item Part 2. For each $i \in [d^2]$
\begin{align*}
    \frac{\d \exp( \mathsf{A} x)}{\d x_i} = \A_{*,i} \circ \exp( \mathsf{A} x) 
\end{align*}
\item Part 3. For each $j_1 \in [n] $, for each $i \in [d^2]$
\begin{align*}
    \frac{\d \A_{[j_1],*} x }{\d x_i} = \A_{[j_1],i}
\end{align*}
\item Part 4. For each $j_1 \in [n]$, for each $i \in [d^2]$
\begin{align*}
    \frac{\d u(x)_{j_1} }{\d x_i} = u(x)_{j_1} \circ \A_{[j_1],i}
\end{align*}
\item Part 5. For each $j_1 \in [n]$, for each $i \in [d^2]$
\begin{align*}
    \frac{\d \alpha(x)_{j_1}}{\d x_i} = \langle u(x)_{j_1} , A_{[j_1], i} \rangle
\end{align*}
\item Part 6. For each $j_1 \in [n]$, for each $i \in [d^2]$,
\begin{align*}
    \frac{\d q(x)_{j_1}}{ \d x_i} = u(x)_{j_1} \circ \A_{[j_1],i} - b_{[j_1]} \cdot \langle u(x)_{j_1}, \A_{ [j_1], i  } \rangle
\end{align*}
\item Part 7. For each $i \in [d^2]$,
\begin{align*}
    \frac{\d L_{q}(x)}{ \d x_i} = \sum_{j_1=1}^n \langle q(x)_{j_1}, u(x)_{j_1} \circ \A_{[j_1],i}\rangle - \langle q(x)_{j_1}, b_{[j_1]} \rangle \cdot \langle u(x)_{j_1}, \A_{ [j_1], i  } \rangle
\end{align*}
\item Part 8. For each $j_1 \in [n]$, for each $i \in [d^2]$,
\begin{align*}
    \frac{\d \alpha(x)_{j_1}^{-1} }{\d x_i} = - \alpha(x)_{j_1}^{-1} \cdot \langle f(x)_{j_1}, \A_{[j_1],i} \rangle
\end{align*}
\item Part 9. For each $j_1 \in [n]$, for each $i \in [d^2]$,
\begin{align*}
    \frac{\d f(x)_{j_1}}{ \d x_i} = f(x)_{j_1} \circ \A_{[j_1],i} - f(x)_{j_1} \cdot \langle f(x)_{j_1}, \A_{[j_1],i} \rangle
\end{align*}
\item Part 10. For each $j_1 \in [n]$, for each $i \in [d^2]$,
\begin{align*}
    \frac{\d h(x)_{j_1}}{ \d x_i} = \A_{[j_1],i} - \langle f(x)_{j_1}, \A_{[j_1],i}\rangle \cdot {\bf 1}_n
\end{align*}
\item Part 11. For each $j_1 \in [n]$, for each $i \in [d^2]$,
\begin{align*}
    \frac{ \d c(x)_{j_1}}{ \d x_i } = f(x)_{j_1} \circ \A_{[j_1],i} - f(x)_{j_1} \cdot \langle f(x)_{j_1}, \A_{[j_1],i} \rangle.
\end{align*}
\item Part 12. For each $i \in [d^2]$, 
\begin{align*}
    \frac{\d L_c(x)}{\d x_i} = \sum_{j_1=1}^n (\langle c(x)_{j_1}, f(x)_{j_1} \circ \A_{[j_1],i} \rangle - \langle c(x)_{j_1}, f(x)_{j_1} \rangle \cdot \langle f(x)_{j_1}, \A_{[j_1],i} \rangle).
\end{align*}
\item Part 13. For each $i \in [d^2]$
\begin{align*}
    \frac {\d L_{\sparse}(x)}{\d x_i} = \sum_{j_1=1}^n \langle u(x)_{j_1} , A_{[j_1], i} \rangle
\end{align*}
\item Part 14. For each $i \in [d^2]$
\begin{align*}
    \frac { \d L_{\cent}(x)}{\d x_i} = \sum_{j_1=1}^n  (\langle f(x)_{j_1}, b_{[j_1]} \rangle  \cdot \langle f(x)_{j_1}, \A_{[j_1],i}\rangle - \langle f(x)_{j_1} \circ \A_{[j_1],i}, b_{[j_1]} \rangle)
\end{align*}
\item Part 15. For each $i \in [d^2]$
\begin{align*}
    L_{\ent} (x) = & ~ \sum_{j_1=1}^n (\langle f(x)_{j_1}  , h(x)_{j_1} \rangle \cdot \langle f(x)_{j_1}, \A_{[j_1],i}\rangle - \langle f(x)_{j_1} \circ \A_{[j_1],i}, h(x)_{j_1} \rangle ~ + \\
    & ~ \langle f(x)_{j_1}, \A_{[j_1],i}\rangle \cdot \langle f(x)_{j_1}, {\bf 1}_n \rangle - \langle f(x)_{j_1}, \A_{[j_1],i} \rangle)
\end{align*}
\end{itemize}
\end{lemma}
\begin{proof}

{\bf Proof of Part 1.}

We have
\begin{align*}
    \frac{\d  (\mathsf{A} x)}{\d x_i}
    = & ~ \frac{\mathsf{A} \d x}{\d x_i}\\
    = & ~ \mathsf{A}_{*,i},
\end{align*}
where the 1st step is because of {\bf Part 1} of Fact~\ref{fac:derivative_rules}, 
the 2nd step is based on the fact that only the $i$-th entry of $\frac{\d x}{\d x_i}$ is $1$ and other entries of it are $0$.

{\bf Proof of Part 2.}

We have
\begin{align*}
    \frac{\d \exp( \mathsf{A} x)}{\d x_i}
    = & ~ \exp( \mathsf{A} x) \circ \frac{\d ( \mathsf{A} x)}{\d x_i}\\
    = & ~ \exp( \mathsf{A} x ) \circ \mathsf{A}_{*,i},
\end{align*}
where the 1st step is because of the chain rule and the 2nd step is due to {\bf Part 1} of this Lemma.

{\bf Proof of Part 3.}
It follows that
\begin{align*}
    \frac{\d \A_{[j_1],*} x }{\d x_i} 
    = & ~ \A_{[j_1],*} \frac{\d x }{\d x_i} \\
    = & ~ \A_{[j_1],i},
\end{align*}
where the 1st step follows form {\bf Part 1} of Fact~\ref{fac:derivative_rules} and the 2nd step is from the fact that only the $i$-th entry of $\frac{\d x}{\d x_i}$ is $1$ and other entries of it are $0$.

{\bf Proof of Part 4.}
We have
\begin{align*}
    \frac{\d u(x)_{j_1} }{\d x_i} 
    = & ~ \frac{\d \exp( \A_{[j_1],*} x ) }{\d x_i}\\
    = & ~ \exp( \A_{[j_1],*} x ) \circ \frac{\d \A_{[j_1],*} x }{\d x_i}\\
    = & ~ u(x)_{j_1} \circ \A_{[j_1],i},
\end{align*}
where the 1st step is given by Definition~\ref{def:u}, the 2nd step holds because of the chain rule, and the 3rd step is due to {\bf Part 3} of this Lemma. 

{\bf Proof of Part 5.}

Let $j_1 \in [n]$. Let $i \in [d^2]$.

We have 
\begin{align*}
    \frac{\d \alpha(x)_{j_1}}{ \d x_i} 
    = & ~ \frac{ \d \langle \exp( \mathsf{A}_{[j_1],*} x ),  {\bf 1}_n \rangle }{\d x_i} \\
    = & ~ \langle \frac{  \d \exp( \mathsf{A}_{[j_1],*} x ) }{\d x_i} ,  {\bf 1}_n \rangle \\
    = & ~ \langle  u(x)_{j_1}   \circ  \mathsf{A}_{[j_1], i} , {\bf 1}_n \rangle \\
    = & ~ \langle u(x)_{j_1} , \A_{[j_1],i} \rangle,
\end{align*}
where the 1st step is given by Definition~\ref{def:alpha}, the 2nd step uses property of the inner product and from {\bf Part 3} of Fact~\ref{fac:derivative_rules}, the 3rd step follows from {\bf Part 4} of this Lemma, and the last step uses {\bf Part 1} of Fact~\ref{fac:vector_properties}.

{\bf Proof of Part 6.}

We have
\begin{align*}
    \frac{\d q(x)_{j_1}}{ \d x_i}
    = & ~ \frac{\d (u(x)_{j_1} - b_{[j_1]} \cdot \alpha(x)_{j_1} ) }{ \d x_i} \\
    = & ~ \frac{\d u(x)_{j_1}}{ \d x_i} - b_{[j_1]} \cdot \frac{\d \alpha(x)_{j_1} }{ \d x_i}\\
    = & ~ u(x)_{j_1} \circ \A_{[j_1],i} - b_{[j_1]} \cdot \langle u(x)_{j_1}, \A_{ [j_1], i  } \rangle
\end{align*}
where the 1st step is due to Definition~\ref{def:c}, the 2nd step is due to {\bf Part 3} of Fact~\ref{fac:derivative_rules}, and the fact that $b_{[j_1]}$ does not contain $x_i$, the last step follows from {\bf Part 4} and {\bf Part 5} of this lemma.

{\bf Proof of Part 7.}

\begin{align*}
    \frac{\d L_{q}(x)}{ \d x_i} 
    = & ~ 0.5 \cdot \frac{\d}{\d x_i} \sum_{j_1=1}^n \langle q(x)_{j_1}, q(x)_{j_1} \rangle \\
    = & ~ \sum_{j_1 = 1}^n \langle q(x)_{j_1}, \frac{\d q(x)_{j_1}}{\d x_i} \rangle\\
    = & ~ \sum_{j_1=1}^n \langle q(x)_{j_1}, u(x)_{j_1} \circ \A_{[j_1],i} - b_{[j_1]} \cdot \langle u(x)_{j_1}, \A_{ [j_1], i  } \rangle \rangle \\
    = & ~ \sum_{j_1=1}^n \langle q(x)_{j_1}, u(x)_{j_1} \circ \A_{[j_1],i}\rangle - \langle q(x)_{j_1}, b_{[j_1]} \rangle \cdot \langle u(x)_{j_1}, \A_{ [j_1], i  } \rangle
\end{align*}
where the 1st step is owing to the definition of 2-norm, the 2nd step follows from chain rule, the 3rd step is derived from {\bf Part 6} of this lemma, and the last step is due to definition and basic calculation rules of inner product.

{\bf Proof of Part 8.}
We have
\begin{align*}
\frac{\d \alpha(x)_{j_1}^{-1} }{\d x_i} 
= & ~ -1 \cdot \alpha(x)_{j_1}^{-2} \cdot \frac{\d \alpha(x)_{j_1}}{ \d x_i} \\
= & ~ -1 \cdot \alpha(x)_{j_1}^{-2} \cdot \langle u(x)_{j_1} , \A_{[j_1],i} \rangle \\
= & ~ - \alpha(x)_{j_1}^{-1} \cdot \langle f(x)_{j_1}, \A_{[j_1],i} \rangle
\end{align*}
where the 1st step is because of chain rule, the 2nd step follows by {\bf Part 5} of this Lemma, and the 3rd step is given by the definition of $f$ (see Definition~\ref{def:f}). 

{\bf Proof of Part 9.}

We have
\begin{align*}
    \frac{\d f(x)_{j_1}}{ \d x_i} 
    = & ~ \frac{\d ( \alpha(x)_{j_1}^{-1} u(x)_{j_1}) }{ \d x_i} \\
    = & ~ \alpha(x)_{j_1}^{-1} \cdot \frac{\d u(x)_{j_1} }{ \d x_i} + \frac{\d \alpha(x)_{j_1}^{-1}}{\d x_i} \cdot u(x)_{j_1} \\
    = & ~ \alpha(x)_{j_1}^{-1} \cdot  u(x)_{j_1} \circ \A_{[j_1],i} - \alpha(x)_{j_1}^{-1} \cdot \langle f(x)_{j_1}, \A_{[j_1],i} \rangle \cdot u(x)_{j_1} \\
    = & ~ f(x)_{j_1} \circ \A_{[j_1],i} - f(x)_{j_1}  \cdot \langle f(x)_{j_1}, \A_{[j_1],i}\rangle
\end{align*}
where the 1st step is due to Definition~\ref{def:f}, the 2nd step is due to chain rule, the 3rd step is due to {\bf Part 4} and {\bf Part 8}  of this Lemma, and the last step is according to Definition~\ref{def:f}.  

{\bf Proof of Part 10.}

We have
\begin{align*}
    \frac{\d h(x)_{j_1}}{ \d x_i} = & ~ f(x)_{j_1}^{-1} \circ \frac{\d f(x)_{j_1}}{\d x_i} \\
    = & ~ f(x)_{j_1}^{-1} \circ ( f(x)_{j_1} \circ \A_{[j_1],i} - f(x)_{j_1}  \cdot \langle f(x)_{j_1}, \A_{[j_1],i}\rangle) \\
    = & ~ \A_{[j_1],i} - \langle f(x)_{j_1}, \A_{[j_1],i}\rangle \cdot {\bf 1}_n
\end{align*}
where the 1st step is because of Definition~\ref{def:h}, the 2nd step is due to {\bf Part 9} of this lemma, the 3rd step holds since $f(x)_{j_1}^{-1} \circ f(x)_{j_1} = {\bf 1}_n$.

{\bf Proof of Part 11.}

We have
\begin{align*}
    \frac{\d c(x)_{j_1}}{ \d x_i}
    = & ~ \frac{\d (f(x)_{j_1} - b_{[j_1]})}{ \d x_i} \\
    = & ~ \frac{\d f(x)_{j_1}}{ \d x_i} - \frac{\d b_{[j_1]}}{ \d x_i}\\
    = & ~ \frac{\d f(x)_{j_1}}{ \d x_i},
\end{align*}
where the 1st step is because of Definition~\ref{def:c}, the 2nd step is due to Fact~\ref{fac:derivative_rules}, and the last step holds, since $b_{[j_1]}$ does not contain $x_i$. 

{\bf Proof of Part 12.}

\begin{align*}
    \frac{\d L_{c}(x)}{ \d x_i} 
    = & ~ 0.5 \frac{\d}{\d x_i} \sum_{j_1=1}^n \langle c(x)_{j_1}, c(x)_{j_1} \rangle \\
    = & ~ \sum_{j_1 = 1}^n \langle c(x)_{j_1}, \frac{\d c(x)_{j_1}}{\d x_i} \rangle\\
    = & ~ \sum_{j_1=1}^n \langle c(x)_{j_1}, f(x)_{j_1} \circ \A_{[j_1],i} - f(x)_{j_1} \cdot \langle f(x)_{j_1}, \A_{[j_1],i} \rangle \rangle \\
    = & ~ \sum_{j_1=1}^n (\langle c(x)_{j_1}, f(x)_{j_1} \circ \A_{[j_1],i} \rangle - \langle c(x)_{j_1}, f(x)_{j_1} \rangle \cdot \langle f(x)_{j_1}, \A_{[j_1],i} \rangle),
\end{align*}
where the 1st step is due to Definition~\ref{def:L_c}, the 2nd step is becaused of chain rule, the 3rd step is due to {\bf Part 9} and {\bf Part 10} of this Lemma, and the last step is based on Fact~\ref{fac:vector_properties}.

{\bf Proof of Part 13.}
\begin{align*}
    \frac {\d L_{\sparse}(x)}{\d x_i} = & ~ \frac{\d}{\d x_i} \sum_{j_1=1}^n \alpha(x)_{j_1} \\
    = & ~ \sum_{j_1=1}^n \frac{\d \alpha(x)_{j_1}}{\d x_i} \\
    = & ~ \sum_{j_1=1}^n \langle u(x)_{j_1} , A_{[j_1], i} \rangle 
\end{align*}
where the 1st step is Definition~\ref{def:L_sparse}, the 2nd step follows by Fact~\ref{fac:derivative_rules}, the 3rd step is because of {\bf Part 5} of this lemma.

{\bf Proof of Part 14}
\begin{align*}
    \frac { \d L_{\cent}(x)}{\d x_i} = & ~  - \frac{\d}{\d x_i} \sum_{j_1=1}^n \langle f(x)_{j_1}, b_{[j_1]} \rangle \\
    = & ~  - \sum_{j_1=1}^n \langle \frac{\d f(x)_{j_1}}{\d x_i}, b_{[j_1]} \rangle \\
    = & ~ - \sum_{j_1=1}^n \langle f(x)_{j_1} \circ \A_{[j_1],i} - f(x)_{j_1}  \cdot \langle f(x)_{j_1}, \A_{[j_1],i}\rangle, b_{[j_1]} \rangle \\
    = & ~ \sum_{j_1=1}^n  (\langle f(x)_{j_1}, b_{[j_1]} \rangle  \cdot \langle f(x)_{j_1}, \A_{[j_1],i}\rangle - \langle f(x)_{j_1} \circ \A_{[j_1],i}, b_{[j_1]} \rangle) \\
\end{align*}
where the 1st step uses the definition of $L_{\cent}(x)$ (see Definition~\ref{def:L_cent}), the 2nd step is derived from Fact~\ref{fac:derivative_rules}, the 3rd step uses {\bf Part 9} of this lemma, the last step is inner product calculation.

{\bf Proof of Part 15}
\begin{align*}
    \frac {\d L_{\ent} (x)}{\d x_i} = & ~  - \frac{\d}{\d x_i} \sum_{j_1=1}^n \langle f(x)_{j_1}, h(x)_{j_1} \rangle \\
    = & ~ - \sum_{j_1=1}^n (\langle \frac{\d f(x)_{j_1}}{\d x_i},h(x)_{j_1} \rangle + \langle f(x)_{j_1}, \frac{\d h(x)_{j_1}}{\d x_i} \rangle) \\
    = & ~ - \sum_{j_1=1}^n (\langle \frac{\d f(x)_{j_1}}{\d x_i},h(x)_{j_1} \rangle + \langle f(x)_{j_1}, \A_{[j_1],i} - \langle f(x)_{j_1}, \A_{[j_1],i}\rangle \cdot {\bf 1}_n \rangle) \\
    = & ~ - \sum_{j_1=1}^n (\langle f(x)_{j_1} \circ \A_{[j_1],i} - f(x)_{j_1}  \cdot \langle f(x)_{j_1}, \A_{[j_1],i}\rangle , h(x)_{j_1} \rangle ~ + \\
    & ~ \langle f(x)_{j_1}, \A_{[j_1],i} - \langle f(x)_{j_1}, \A_{[j_1],i}\rangle \cdot {\bf 1}_n \rangle) \\
    = & ~ \sum_{j_1=1}^n (\langle f(x)_{j_1}  , h(x)_{j_1} \rangle \cdot \langle f(x)_{j_1}, \A_{[j_1],i}\rangle - \langle f(x)_{j_1} \circ \A_{[j_1],i}, h(x)_{j_1} \rangle ~ + \\
    & ~ \langle f(x)_{j_1}, \A_{[j_1],i}\rangle \cdot \langle f(x)_{j_1}, {\bf 1}_n \rangle - \langle f(x)_{j_1}, \A_{[j_1],i} \rangle)
\end{align*}
where the 1st step is from the definition of $L_{\ent}(x)$ (see Definition~\ref{def:L_ent}), the 2nd step uses the chain rule, the 3rd step is due to {\bf Part 10} of this lemma, the 4th step is due to {\bf Part 9} of this lemma, the last step is inner product calculation and rearrangement.

\end{proof}

\section{Bounds for Basic Functions} \label{sec:upper_bounds}
We state and prove some bounds for auxiliary functions in this section. In Section~\ref{sec:lower_bound:beta} and Section~\ref{sec:lower_bound_A:beta}, we state a lower bound of $\beta$. In Section~\ref{sec:upper_bound:u}, we derive an upper bound for $\|u(x)\|_2$. In Section~\ref{sec:upper_bound:alpha_inverse}, we derive an upper bound for $|\alpha^{-1}(x)|$. In Section~\ref{sec:upper_bound:f}, we state an upper bound for $\|f(x)\|_2$. In Section~\ref{sec:upper_bound:h}, we state an upper bound for $\|h(x)\|_2$. In Section~\ref{sec:upper_bound:c}, we derive an upper bound for $\|c(x)\|_2$. In Section~\ref{sec:upper_bound:alpha}, we derive an upper bound for $\|\alpha(x)\|_2$. 
In Section~\ref{sec:upper_bound:q}, we derive an upper bound for $\|q(x)\|_2$. 

\subsection{Lower bound on \texorpdfstring{$\beta$ of $u(x)$}{}}\label{sec:lower_bound:beta}

\begin{lemma}[\cite{dls23,lsx+23}]\label{lem:lower_bound:beta}
Provided that the subsequent requirement are satisfied
\begin{itemize}
    \item Let $u(x)_{j_1}$ be denoted as Definition~\ref{def:u}
    \item Let $x \in \R^{d^2}$ satisfy $\| x \|_2 \leq R$
    \item The greatest lower bound of $\langle u(x)_{j_1} , {\bf 1}_n \rangle$ is denoted as $\beta$
    \item Let $\max_{j \in [n]}\|\A_{[j],*} \| \leq R$ 
\end{itemize}  
Then we have
\begin{align*}
    \beta \geq \exp(-R^2).
\end{align*}
\end{lemma}

\subsection{Lower bound on \texorpdfstring{$\beta$ of $u(\A)$}{}}\label{sec:lower_bound_A:beta}

\begin{lemma}[\cite{dls23,lsx+23}]\label{lem:lower_bound_A:beta}
Provided that the subsequent requirement are satisfied
\begin{itemize}
    \item Let $\A \in \R^{n^2 \times d^2}$ satisfy $\max_{j_1 \in [n]}\| \A_{[j_1],*} \| \leq R$
    \item Let $x \in \R^{d^2}$ satisfy that $\| x \|_2 \leq R $
    \item We define $u(A)$ as Definition~\ref{def:u}
    \item Let $\beta$ be the greatest lower bound of $\langle u(\A)_{j_1} , {\bf 1}_n \rangle$
\end{itemize}
Then we have
\begin{align*}
    \beta \geq \exp(-R^2).
\end{align*}
\end{lemma}

\subsection{Upper bound on \texorpdfstring{$\| u(x)_{j_1} \|_2$}{}} \label{sec:upper_bound:u}

\begin{lemma}[\cite{dls23,lsx+23}]\label{lem:upper_bound:u}
Provided that the subsequent requirement are satisfied
\begin{itemize}
    \item Let $x \in \R^{d^2}$ satisfy $\| x \|_2 \leq R$
    \item We define $u(x)$ as Definition~\ref{def:u}.
    \item Let $\max_{j_1 \in [n]}\| \A_{[j_1],*} \| \leq R$ 
\end{itemize}
Then we have
\begin{align*}
    \| u(x)_{j_1} \|_2 \leq \sqrt{n} \cdot \exp(R^2).
\end{align*}
\end{lemma}

\subsection{Upper bound on \texorpdfstring{$| \alpha^{-1}(x)_{j_1} |$}{}}\label{sec:upper_bound:alpha_inverse}

\begin{lemma} \label{lem:upper_bound:alpha_inverse}
Provided that the subsequent requirement are satisfied
\begin{itemize}
    \item Let $x \in \R^{d^2}$ satisfy $\| x \|_2 \leq R$
    \item We define $\alpha(x)_{j_1}$ as Definition~\ref{def:alpha}
    \item The greatest lower bound of $\langle u(x)_{j_1} , {\bf 1}_n \rangle$ is denoted as $\beta$.
    \item Let $\max_{j \in [n]}\|\A_{[j],*} \| \leq R$ 
\end{itemize}
Then we have
\begin{align*}
    | \alpha^{-1}(x)_{j_1} | \leq \exp(R^2).
\end{align*}
\end{lemma}
\begin{proof}
\begin{align*}
| \alpha^{-1}(x)_{j_1} | = & ~ \frac{1}{\langle u(x)_{j_1} , {\bf 1}_n \rangle} \\
\leq & ~ \frac{1}{\beta} \\
\leq & ~ \exp(R^2)
\end{align*}
where the first step uses Definition~\ref{def:alpha},
the 2nd step is based on assumptions of Lemma~\ref{lem:upper_bound:alpha_inverse},
the last step is given by Lemma~\ref{lem:lower_bound:beta}.
\end{proof}

\subsection{Upper bound on \texorpdfstring{$\| f(x)_{j_1} \|_2$}{}} \label{sec:upper_bound:f}

\begin{lemma} \label{lem:upper_bound:f}
Provided that the subsequent requirement are satisfied
\begin{itemize}
    \item We define $f(x)$ as Definition~\ref{def:f}
\end{itemize}
It follows that
\begin{align*}
    \| f(x)_{j_1} \|_2 \leq 1.
\end{align*}
\end{lemma}
\begin{proof}
\begin{align*}
    \| f(x)_{j_1} \|_2^2 = & ~ \frac{1}{|\langle u(x)_{j_1}, {\bf 1}_n \rangle|^2} \cdot \| u(x)_{j_1} \|_2^2\\
    \leq & ~ 1
\end{align*}
the reason for the last step is that every entry of $u(x)_{j_1}$ is positive.
\end{proof}

\subsection{Upper bound on \texorpdfstring{$\| h(x)_{j_1} \|_2$}{}} \label{sec:upper_bound:h}
\begin{lemma}
[Lemma F.2 in \cite{ssz23}]\label{lem:upper_bound_h:x}
Provided that the subsequent requirement are satisfied
\begin{itemize}
    \item Let $x \in \R^{d^2}$ satisfy $\| x \|_2 \leq R$
    \item We define $h(x)$ as Definition~\ref{def:h}
    \item The greatest lower bound of $\langle u(x)_{j_1} , {\bf 1}_n \rangle$ is denoted as $\beta$
\end{itemize}
Then, for all $j_1 \in [n]$ we have
\begin{align*}
    \| h(x)_{j_1} \|_2 \leq 2 \sqrt{n} R^2
\end{align*}
\end{lemma}

\subsection{Upper bound on \texorpdfstring{$\| c(x)_{j_1} \|_2$}{}} \label{sec:upper_bound:c}

\begin{lemma} \label{lem:upper_bound:c}
Provided that the subsequent requirement are satisfied
\begin{itemize}
    \item We define $c(x)$ as Definition~\ref{def:c}
    \item Let $\max_{j_1 \in [n]} \| b_{[j_1]} \|_2 \leq 1$
\end{itemize}
Then we have
\begin{align*}
    \| c(x)_{j_1} \|_2 \leq 2.
\end{align*}
\end{lemma}
\begin{proof}
\begin{align*}
    \| c(x)_{j_1} \|_2 = & ~ \| f(x)_{j_1} - b_{[j_1]} \|_2 \\
    \leq & ~ \| f(x)_{j_1} \|_2 + \| b_{[j_1]} \|_2 \\
    \leq & ~ 2
\end{align*}
where the first step uses Definition~\ref{def:c},
the 2nd step is given by triangle inequality,
the last step is due to Lemma~\ref{lem:upper_bound:f} and lemma assumption.
\end{proof}

\subsection{Upper bound on \texorpdfstring{$| \alpha(x)_{j_1} |$}{}} \label{sec:upper_bound:alpha}

\begin{lemma}\label{lem:upper_bound:alpha}
Provided that the subsequent requirement are satisfied
\begin{itemize}
    \item Let $x \in \R^{d^2}$ satisfies $\| x \|_2 \leq R$
    \item We define $\alpha(x)$ as Definition~\ref{def:alpha}
    \item Let $\max_{j_1 \in [n]}\| \A_{[j_1],*} \| \leq R$ 
\end{itemize}
Then, for all $j \in [n]$ it follows that
\begin{align*}
    | \alpha(x)_{j_1} | \leq n \exp(R^2).
\end{align*}
\end{lemma}
\begin{proof}
\begin{align*}
    | \alpha(x)_{j_1} | = & ~ | \langle u(x)_{j_1}, {\bf 1}_n \rangle | \\
    \leq & ~ \sqrt{n} \cdot \| u(x)_{j_1} \|_2 \\
    \leq & ~ \sqrt{n} \cdot \sqrt{n} \cdot \exp(R^2) \\
    = & ~ n \exp(R^2) 
\end{align*}
where the first step is owing to Definition~\ref{def:alpha}, 
the 2nd step follows by Cauchy-Schwartz inequality,
the 3rd step is given by Lemma~\ref{lem:upper_bound:u}.
\end{proof}

\subsection{Upper bound on \texorpdfstring{$\| q(x)_{j_1} \|_2$}{}} \label{sec:upper_bound:q}

\begin{lemma} \label{lem:upper_bound:q}
Provided that the subsequent requirement are satisfied
\begin{itemize}
    \item Let $x \in \R^{d^2}$ satisfies that $\| x \|_2 \leq R$
    \item Let $q(x)$ be defined as Definition~\ref{def:q}
    \item Let $\max_{j_1 \in [n]}\| \A_{[j_1],*} \| \leq R$ 
    \item Let $\max_{j_1 \in [n]} \| b_{[j_1]} \|_2 \leq 1$
\end{itemize}
It follows that
\begin{align*}
    \| q(x)_{j_1} \|_2 \leq 2n\exp(R^2).
\end{align*}
\end{lemma}
\begin{proof}
$\| q(x)_{j_1} \|_2$ is bounded as follows:
\begin{align*}
    \| q(x)_{j_1} \|_2 = & ~ \| u(x)_{j_1} -\alpha(x)_{j_1}b_{[j_1]} \|_2 \\
    \leq & ~ \| u(x)_{j_1} \|_2 + | \alpha(x)_{j_1} | \cdot \| b_{[j_1]} \|_2 \\
    \leq & ~ \| u(x)_{j_1} \|_2 + | \alpha(x)_{j_1} | \\
    \leq & ~ \sqrt{n} \exp(R^2) + | \alpha(x)_{j_1}| \\
    \leq & ~ \sqrt{n} \exp(R^2) + n \exp(R^2) \\
    \leq & ~ 2 n \exp(R^2) 
\end{align*}
where the first step is due to  Definition~\ref{def:q}, 
the 2nd step follows by triangle inequality,
the 3rd step is derived by lemma assumption,
the 4th step follows from Lemma~\ref{lem:upper_bound:u}, the 5th step is due to Lemma~\ref{lem:upper_bound:alpha},
the last step follows from $n \geq 1$.
\end{proof}

\section{Lipschitz of Basic Functions}\label{sec:lipschitz_basic_fcts}

In this section, we state some lemmas of Lipschitz conditions for basic functions. For functions of $\A$, their definitions and gradients are the same as functions of $x$, except that they have different independent variables, i.e., $u(\A) = u(x)$, $\frac{\d L_c(\A)}{\d x_i} = \frac{\d L_c(x)}{\d x_i}$.

In Section~\ref{sec:lipschitz_exp:x}, he Lipschitz for function $u$ with respect to $x$ is computed. 
Section~\ref{sec:lipschitz_alpha:x} presents the Lipschitz for function $\alpha$ with respect to $x$. Section~\ref{sec:lipschitz_alpha_inverse:x} provides the Lipschitz for function $\alpha^{-1}$ with respect to $x$. In Section~\ref{sec:lipschitz_f:x}, we compute the Lipschitz for function $f$ with respect to $x$. In
Section~\ref{sec:lipschitz_h:x}, we state the Lipschitz for function $f$ with respect to $x$. In Section~\ref{sec:lipschitz_c:x}, we compute the Lipschitz for function $c$ with respect to $x$. In Section~\ref{sec:lipschitz_q:x}, we compute the Lipschitz for function $q$ with respect to $x$.

The Lipschitz for function $u$ with respect to $\A$ is computed in Section~\ref{sec:lipschitz_A:u}. 
 In Section~\ref{sec:lipschitz_A:alpha}, we compute the Lipschitz for function $\alpha$ with respect to $\A$. In Section~\ref{sec:lipschitz_A:alpha_inverse}, we compute the Lipschitz for function $\alpha^{-1}$ with respect to $\A$. In Section~\ref{sec:lipschitz_A:f}, we compute the Lipschitz for function $f$ with respect to $\A$. In
Section~\ref{sec:lipschitz_h:A}, we state the Lipschitz for function $h$ with respect to $\A$. In Section~\ref{sec:lipschitz_A:c}, we compute the Lipschitz for function $c$ with respect to $\A$. In Section~\ref{sec:lipschitz_A:q}, we compute the Lipschitz for function $q$ with respect to $\A$.

\subsection{Lipschitz for \texorpdfstring{$u(x)$}{} function} \label{sec:lipschitz_exp:x}
\begin{lemma}\label{lem:lipschitz_exp:x}
Provided that the subsequent requirement are satisfieds
\begin{itemize}
    \item Let $x \in \R^{d^2}, y \in \R^{d^2}$ satisfy $\| x \|_2 \leq R$ and $\| y \|_2 \leq R$
    \item Let $\A \in \R^{n^2 \times d^2}$
    \item We define $u(x)$ as Definition~\ref{def:u}
    \item Let $\max_{j_1 \in [n]} \| \A_{[j_1],*} (y-x) \|_{\infty} < 0.01$
    \item Let $\max_{j_1 \in [n]}\| \A_{[j_1],*} \| \leq R$  
\end{itemize}
Then, for all $j_1 \in [n]$,
\begin{align*}
    \| u(x)_{j_1} - u(y)_{j_1} \|_2 \leq 2 \sqrt{n} R \exp(R^2) \cdot \| x - y \|_2.
\end{align*}
\end{lemma}
\begin{proof}
\begin{align*}
\| \exp( \A_{[j_1],*} x) - \exp( \A_{[j_1],*} y) \|_2 
\leq & ~ \exp(\| \A_{[j_1],*}x \|_2 )  \cdot 2 \| \A_{[j_1],*} (x-y) \|_{\infty} \notag \\
\leq & ~ \sqrt{n} \exp(R^2) \cdot 2 \| \A_{[j_1],*} (x-y) \|_2 \notag \\
\leq & ~ \sqrt{n} \exp(R^2)  \cdot 2 \| \A_{[j_1],*} \| \cdot \| x - y \|_2 \notag\\
\leq & ~ 2 \sqrt{n} R \exp(R^2) \cdot \|x - y\|_2
\end{align*} 
where the 1st step is because of $\max_{j_1 \in [n]} \| ( \A_{[j_1],*} - \B_{[j_1],*} ) x \|_{\infty} < 0.01$ and Fact~\ref{fac:vector_norm},
the 2nd step follows from Lemma~\ref{lem:upper_bound:u},
the 3rd step follows from Fact~\ref{fac:matrix_norm}, 
the last step follows from lemma assumptions.
\end{proof}

\subsection{Lipschitz for \texorpdfstring{$\alpha(x)$}{} function} \label{sec:lipschitz_alpha:x}

\begin{lemma}\label{lem:lipschitz_alpha:x}
Provided that the subsequent requirement are satisfied
\begin{itemize}
    \item Let $x \in \R^{d^2}, y \in \R^{d^2}$ satisfy $\| x \|_2 \leq R$ and $\| y \|_2 \leq R$
    \item Let $\A \in \R^{n^2 \times d^2}$
    \item We define $\alpha(x)$ as Definition~\ref{def:alpha}
    \item Let $\max_{j_1 \in [n]} \| \A_{[j_1],*} (y-x) \|_{\infty} < 0.01$
    \item Let $\max_{j_1 \in [n]}\| \A_{[j_1],*} \| \leq R$ 
\end{itemize}
Then, for all $j_1 \in [n]$, it follows that
\begin{align*}
    | \alpha(x)_{j_1} - \alpha(y)_{j_1} | \leq \sqrt{n} \cdot \| u(x)_{j_1} - u(y)_{j_1} \|_2
\end{align*}

\end{lemma}
\begin{proof}
\begin{align*}
| \alpha(x)_{j_1} - \alpha(y)_{j_1} |
= & ~ | \langle u(x)_{j_1} , {\bf 1}_n \rangle - \langle u(y)_{j_1} , {\bf 1}_n \rangle |  \\
= & ~ | \langle u(x)_{j_1} - u(y)_{j_1}, {\bf 1}_n \rangle | \\
\leq & ~ \| u(x)_{j_1} - u(y)_{j_1} \|_2 \cdot \sqrt{n}
\end{align*}
where the last step follows after Cauchy-Schwarz inequality.
\end{proof}

\subsection{Lipschitz for \texorpdfstring{$\alpha^{-1}(x)$}{} function} \label{sec:lipschitz_alpha_inverse:x}

\begin{lemma}\label{lem:lipschitz_alpha_inverse:x}
Provided that the subsequent requirement are satisfied
\begin{itemize}
    \item Let $x \in \R^{d^2}, y \in \R^{d^2}$ satisfy $\| x \|_2 \leq R$ and $\| y \|_2 \leq R$
    \item We define $\alpha(x)$ as Definition~\ref{def:alpha}
    \item The greatest lower bound of $\langle u(x)_{j_1} , {\bf 1}_n \rangle$ is denoted as $\beta$
\end{itemize}
Then, for all $j_1 \in [n]$ we have
\begin{itemize}
    \item $| \alpha(x)_{j_1}^{-1} - \alpha(y)_{j_1}^{-1} | \leq \beta^{-2} \cdot | \alpha(x)_{j_1} - \alpha(y)_{j_1} |$
\end{itemize}
\end{lemma}
\begin{proof}
The proof is similar to \cite{dls23}, so we  omit the details here.
\end{proof}

\subsection{Lipschitz for \texorpdfstring{$f(x)$}{} function} \label{sec:lipschitz_f:x}

\begin{lemma}\label{lem:lipschitz_f:x}
Provided that the subsequent requirement are satisfied
\begin{itemize}
    \item Let $x \in \R^{d^2}, y \in \R^{d^2}$ satisfy $\| x \|_2 \leq R$ and $\| y \|_2 \leq R$
    \item Let $\A \in \R^{n^2 \times d^2}$
    \item Let $f(x)$ be defined as Definition~\ref{def:f}
    \item Let $\max_{j_1 \in [n]} \| \A_{[j_1],*} (y-x) \|_{\infty} < 0.01$
    \item Let $\max_{j_1 \in [n]}\| \A_{[j_1],*} \| \leq R$
    \item The greatest lower bound of $\langle u(x)_{j_1} , {\bf 1}_n \rangle$ is denoted as $\beta$
\end{itemize}
Then, for all $j_1 \in [n]$ we have
\begin{itemize}
    \item $\| f(x)_{j_1} - f(y)_{j_1} \|_2 \leq 4 \beta^{-2} n^{1.5} R \exp(2R^2) \cdot \|x-y\|_2$
\end{itemize}
\end{lemma}
\begin{proof}
We can show that
\begin{align} \label{eq:diff_f:x}
    & ~ \| f(x)_{j_1} - f(y)_{j_1} \|_2 \nonumber \\
    = & ~ \| \alpha(x)_{j_1}^{-1} u(x)_{j_1} - \alpha(y)_{j_1}^{-1} u(y)_{j_1} \|_2 \nonumber \\
    \leq & ~ \| \alpha(x)_{j_1}^{-1} u(x)_{j_1} - \alpha(y)_{j_1}^{-1} u(x)_{j_1} \|_2 + \| \alpha(y)_{j_1}^{-1} u(x)_{j_1} - \alpha(y)_{j_1}^{-1} u(y)_{j_1} \|_2 \nonumber \\
    = & ~ |\alpha(x)_{j_1}^{-1} - \alpha(y)_{j_1}^{-1} | \cdot \| u(x)_{j_1} \|_2 + | \alpha(y)_{j_1}^{-1} | \cdot \| u(x)_{j_1} -u(y)_{j_1} \|_2
\end{align}
where the first step follow by the definition of $f$ (see Definition~\ref{def:f}), the 2nd and the last step follows from Fact~\ref{fac:matrix_norm}.

For the first term of Eq.~\eqref{eq:diff_f:x}, we have
\begin{align} \label{eq:diff_f_p1:x}
& ~ |\alpha(x)_{j_1}^{-1} - \alpha(y)_{j_1}^{-1} | \cdot \| u(x)_{j_1} \|_2 \nonumber \\
\leq & ~ \beta^{-2} \cdot | \alpha(x)_{j_1} - \alpha(y)_{j_1} | \cdot \| u(x)_{j_1} \|_2 \nonumber\\
\leq & ~ \beta^{-2} \cdot | \alpha(x)_{j_1} - \alpha(y)_{j_1} | \cdot \sqrt{n} \cdot \exp(R^2) \nonumber\\
\leq & ~ \beta^{-2} \cdot \| u(x)_{j_1} - u(y)_{j_1} \|_2 \cdot \sqrt{n} \cdot \sqrt{n} \cdot \exp(R^2) \nonumber\\
\leq & ~ \beta^{-2} \cdot 2 \sqrt{n} R \exp(R^2) \cdot \| x - y \|_2 \cdot \sqrt{n} \cdot \sqrt{n} \cdot \exp(R^2) \nonumber \\
\leq & ~ 2 \beta^{-2}  n^{1.5} R \exp(2R^2) \cdot \|x-y\|_2
\end{align}
where the first step holds because of Lemma~\ref{lem:lipschitz_alpha_inverse:x}, the 2nd step follows by Lemma~\ref{lem:upper_bound:u}, the 3rd step follows from Lemma~\ref{lem:lipschitz_alpha:x}, the 4th step is due to Lemma~\ref{lem:lipschitz_exp:x}, the last step is a rearrangement of step 4.

For the second term of the equation, we have
\begin{align} \label{eq:diff_f_p2:x}
| \alpha(y)_{j_1}^{-1} | \cdot \| u(x)_{j_1} -u(y)_{j_1} \|_2 \leq & ~ \beta^{-1} \cdot \| u(x)_{j_1} -u(y)_{j_1} \|_2 \nonumber \\
\leq & ~ \beta^{-1} \cdot 2 \sqrt{n} R \exp(R^2) \cdot \| x - y \|_2 \nonumber \\
\leq & ~ 2 \beta^{-1} n^{0.5} R \exp(R^2) \cdot \| x - y \|_2
\end{align}
where the first step is given by the assumption in Lemma~\ref{lem:lipschitz_alpha_inverse:x}, the second step follows from Lemma~\ref{lem:lipschitz_exp:x}.

Putting Eq.~\eqref{eq:diff_f_p1:x} and Eq.~\eqref{eq:diff_f_p2:x} into Eq.~\eqref{eq:diff_f:x}, we have
\begin{align*}
    \| f(x)_{j_1} - f(y)_{j_1} \|_2 \leq & ~ 2 \beta^{-2}  n^{1.5} R \exp(2R^2) \cdot \|x-y\|_2 + 2 \beta^{-1} n^{0.5} R \exp(R^2) \cdot \| x - y \|_2 \\
    \leq & ~ 4 \beta^{-2} n^{1.5} R \exp(2R^2) \cdot \|x-y\|_2 
\end{align*}
where the 2nd step follows by Lemma~\ref{lem:lower_bound:beta}
\end{proof}

\subsection{Lipschitz for \texorpdfstring{$h(x)$}{} function} \label{sec:lipschitz_h:x}

\begin{lemma}
[Lemma G.2 in \cite{ssz23}]\label{lem:lipschitz_h:x}
Provided that the subsequent requirement are satisfied
\begin{itemize}
    \item Let $x \in \R^{d^2}, y \in \R^{d^2}$ satisfy $\| x \|_2 \leq R$ and $\| y \|_2 \leq R$
    \item We define $h(x)$ as Definition~\ref{def:h}
    \item Let $\A \in \R^{n^2 \times d^2}$
    \item Let $\max_{j_1 \in [n]}\| \A_{[j_1],*} \| \leq R$
    \item $\| {\bf 1}_n - \frac{f(x)_{j_1}}{f(y)_{j_1}} \|_\infty \leq 0.1$
\end{itemize}
Then, for all $j_1 \in [n]$ we have
\begin{align*}
    \| h(x)_{j_1} - h(y)_{j_1} \|_2 \leq \| f(x)_{j_1} - f(y)_{j_1} \|_2
\end{align*}
\end{lemma}

\subsection{Lipschitz for \texorpdfstring{$c(x)$}{} function} \label{sec:lipschitz_c:x}

\begin{lemma}\label{lem:lipschitz_c:x}
Provided that the subsequent requirement are satisfied
\begin{itemize}
    \item Let $x \in \R^{d^2}, y \in \R^{d^2}$ satisfy $\| x \|_2 \leq R$ and $\| y \|_2 \leq R$
    \item Let $\A \in \R^{n^2 \times d^2}$
    \item Let $c(x)$ be defined as Definition~\ref{def:c}
    \item Let $\max_{j_1 \in [n]} \| \A_{[j_1],*} (y-x) \|_{\infty} < 0.01$
    \item Let $\max_{j_1 \in [n]}\| \A_{[j_1],*} \| \leq R$
    \item The greatest lower bound of $\langle u(x)_{j_1} , {\bf 1}_n \rangle$ is denoted as $\beta$
\end{itemize}
Then, for all $j_1 \in [n]$ we have
\begin{itemize}
    \item $\| c(x)_{j_1} - c(y)_{j_1} \|_2 \leq 4 \beta^{-2} n^{1.5} R \exp(2R^2) \cdot \| x - y \|_2$
\end{itemize}
\end{lemma}
\begin{proof}
It follows from the definition of $c$ (see Definition~\ref{def:c}) and we use Lemma~\ref{lem:lipschitz_f:x} directly.
\end{proof}

\subsection{Lipschitz for \texorpdfstring{$q(x)$}{} function} \label{sec:lipschitz_q:x}

\begin{lemma} \label{lem:Lipschitz_q:x}
Provided that the subsequent requirement are satisfied
\begin{itemize}
    \item Let $x \in \R^{d^2}, y \in \R^{d^2}$ satisfy $\| x \|_2 \leq R$ and $\| y \|_2 \leq R$
    \item Let $\A \in \R^{n^2 \times d^2}$
    \item Let $q(x)$ be defined as Definition~\ref{def:q}
    \item Let $\max_{j_1 \in [n]} \| \A_{[j_1],*} (y-x) \|_{\infty} < 0.01$
    \item Let $\max_{j_1 \in [n]}\| \A_{[j_1],*} \| \leq R$
    \item Let $\max_{j_1 \in [n]} \| b_{[j_1]} \|_2 \leq 1$
\end{itemize}
Then we have
\begin{itemize}
    \item $\| q(x)_{j_1} - q(y)_{j_1} \|_2 \leq 4 n R \exp(R^2) \| x - y \|_2$
\end{itemize}
\end{lemma}
\begin{proof}
We have
\begin{align*}
\| q(x)_{j_1} - q(y)_{j_1} \|_2 
= & ~  \| u(x)_{j_1} - \alpha(x)_{j_1} b_{[j_1]} - ( u(y)_{j_1} - \alpha(y)_{j_1} b_{[j_1]}) \|_2 \\
\leq & ~  \| u(x)_{j_1} - u(y)_{j_1} \|_2 + \| \alpha(x)_{j_1} b_{[j_1]} - \alpha(y)_{j_1} b_{[j_1]} \|_2 \\
\leq & ~ \| u(x)_{j_1} - u(y)_{j_1} \|_2 + | \alpha(x)_{j_1} - \alpha(y)_{j_1} | \cdot \| b_{[j_1]} \|_2 \\
\leq & ~ \| u(x)_{j_1} - u(y)_{j_1} \|_2 + | \alpha(x)_{j_1} - \alpha(y)_{j_1} | \\
\leq & ~ \| u(x)_{j_1} - u(y)_{j_1} \|_2 + \sqrt{n} \| u(x)_{j_1} - u(y)_{j_1} \|_2 \\
\leq & ~ 2 \sqrt{n} \cdot \| u(x)_{j_1} - u(y)_{j_1} \|_2  \\
\leq & ~ 2 \sqrt{n} \cdot 2 \sqrt{n} R \exp(R^2) \| x - y \|_2 \\
= & ~ 4 n R \exp(R^2) \| x - y \|_2
\end{align*}
where the first step is from the definition of $q(x)$ (see Definition~\ref{def:q}), the 2nd step is because of triangle inequality, the 3rd step follows by property of norm, the 4th step follows from lemma assumption, the 5th step is due to Lemma~\ref{lem:lipschitz_alpha:x}, the 6th step follows from $n \geq 1$, the 7th step follows from Lemma~\ref{lem:lipschitz_exp:x}, and the last step is a rearrangement of step 6.
\end{proof}

\subsection{Lipschitz for \texorpdfstring{$u(\A)$}{} function}\label{sec:lipschitz_A:u}

\begin{lemma}\label{lem:lipschitz_exp:A}
Provided that the subsequent requirement are satisfieds
\begin{itemize}
    \item Let $\A, \B \in \R^{n^2 \times d^2}$ satisfy $\max_{j_1 \in [n]}\| \A_{[j_1],*} \| \leq R$, $\max_{j_1 \in [n]} \| \B_{[j_1],*} \| \leq R$ \item Let $\max_{j_1 \in [n]} \| ( \A_{[j_1],*} - \B_{[j_1],*} ) x \|_{\infty} < 0.01$
    \item Let $x \in \R^{d^2}$ satisfy that $\| x \|_2 \leq R $
    \item We define $u(A)$ as Definition~\ref{def:u}
\end{itemize}
Then, for all $j_1 \in [n]$ we have
\begin{align*}
    \| u(\A)_{j_1} - u(\B)_{j_1} \|_2 \leq 2 \sqrt{n} R \exp(R^2) \cdot \| \A_{[j_1],*} - \B_{[j_1],*} \|.
\end{align*}
\end{lemma}
\begin{proof}
We have
\begin{align*}
\| \exp( \A_{[j_1],*} x) - \exp( \B_{[j_1],*} x) \|_2 
\leq & ~ \| \exp( \A_{[j_1],*} x) \|_2 \cdot 2 \| ( \A_{[j_1],*} - \B_{[j_1],*}) x \|_{\infty} \notag \\
\leq & ~ \sqrt{n} \exp(R^2) \cdot 2 \| ( \A_{[j_1],*} - \B_{[j_1],*} ) x \|_2 \notag \\
\leq & ~ \sqrt{n} \exp(R^2)  \cdot 2 \| x \| \cdot \| \A_{[j_1],*} - \B_{[j_1],*} \| \notag\\
\leq & ~ 2 \sqrt{n} R \exp(R^2) \cdot \| \A_{[j_1],*} -\B_{[j_1],*} \|
\end{align*} 
where the 1st step is due to $\max_{j_1 \in [n]} \| ( \A_{[j_1],*} - \B_{[j_1],*} ) x \|_{\infty} < 0.01$ and Fact~\ref{fac:vector_norm}, 
the 2nd step follows by Lemma~\ref{lem:upper_bound:u}, 
the 3rd step follows after Fact~\ref{fac:matrix_norm}, 
the last step follows from lemma assumptions.
\end{proof}

\subsection{Lipschitz for \texorpdfstring{$\alpha(\A)$}{} function}\label{sec:lipschitz_A:alpha}

\begin{lemma}\label{lem:lipschitz_alpha:A}
Provided that the subsequent requirement are satisfied
\begin{itemize}
    \item We define $\alpha(\A)$ as Definition~\ref{def:alpha}.
    \item We define $u(A)$ as Definition~\ref{def:u}
\end{itemize}
Then, for all $j_1 \in [n]$ we have
\begin{itemize}
    \item $| \alpha(\A)_{j_1} - \alpha(\B)_{j_1} | \leq \sqrt{n} \cdot \| u(\A)_{j_1} - u(\B)_{j_1} \|_2$
\end{itemize}
\end{lemma}
\begin{proof}
We have
\begin{align*}
| \alpha( \A )_{j_1} - \alpha( \B )_{j_1} |
= & ~ | \langle u( \A )_{j_1} , {\bf 1}_n \rangle - \langle u( \B )_{j_1} , {\bf 1}_n \rangle |  \\
= & ~ | \langle u( \A )_{j_1} - u( \B )_{j_1}, {\bf 1}_n \rangle | \\
\leq & ~ \| u( \A )_{j_1} - u( \B )_{j_1} \|_2 \cdot \sqrt{n}
\end{align*}
where the last step is because of Cauchy-Schwarz inequality.
\end{proof}

\subsection{Lipschitz for \texorpdfstring{$\alpha^{-1}(\A)$}{} function}\label{sec:lipschitz_A:alpha_inverse}

\begin{lemma}\label{lem:lipschitz_alpha_inverse:A}
Provided that the subsequent requirement are satisfied
\begin{itemize}
    \item Let $\A, \B \in \R^{n^2 \times d^2}$ satisfy $\max_{j_1 \in [n]}\| \A_{[j_1],*} \| \leq R$, $\max_{j_1 \in [n]} \| \B_{[j_1],*} \| \leq R$ 
    \item Let $x \in \R^{d^2}$ satisfy that $\| x \|_2 \leq R $
    \item We define $\alpha(\A)$ as Definition~\ref{def:alpha}
    \item Let $\max_{j_1 \in [n]} \| b_{[j_1]} \|_2 \leq 1$
    \item Let $\beta$ be the greatest lower bound of $\langle u(\A)_{j_1} , {\bf 1}_n \rangle$
\end{itemize}
for all $j_1 \in [n]$, it follows that
\begin{itemize}
    \item $| \alpha( \A )_{j_1}^{-1} - \alpha( \B )_{j_1}^{-1} | \leq \beta^{-2} \cdot | \alpha( \A )_{j_1} - \alpha( \B )_{j_1} |$
\end{itemize}
\end{lemma}
\begin{proof}
The proof shares similarities with that of \cite{dls23}, so we omit the details here.
\end{proof}

\subsection{Lipschitz for \texorpdfstring{$f(\A)$}{} function}\label{sec:lipschitz_A:f}

\begin{lemma}\label{lem:lipschitz_f:A}
Provided that the subsequent requirement are satisfied
\begin{itemize}
    \item Let $\A, \B \in \R^{n^2 \times d^2}$ satisfy $\max_{j_1 \in [n]}\| \A_{[j_1],*} \| \leq R$, $\max_{j_1 \in [n]} \| \B_{[j_1],*} \| \leq R$ \item Let $\max_{j_1 \in [n]} \| ( \A_{[j_1],*} - \B_{[j_1],*} ) x \|_{\infty} < 0.01$
    \item Let $x \in \R^{d^2}$ satisfy that $\| x \|_2 \leq R $
    \item We define $f(\A)$ as Definition~\ref{def:f}
    \item Let $\max_{j_1 \in [n]} \| b_{[j_1]} \|_2 \leq 1$
    \item Let $\beta$ be the greatest lower bound of $\langle u(\A)_{j_1} , {\bf 1}_n \rangle$
\end{itemize}
Then, for all $j_1 \in [n]$ we have
\begin{itemize}
    \item $\| f(\A)_{j_1} - f(\B )_{j_1} \|_2 \leq 4 \beta^{-2} n^{1.5} R \exp(2R^2) \cdot \| \A_{[j_1],*} - \B_{[j_1],*} \|$
\end{itemize}
\end{lemma}
\begin{proof}
We can show that
\begin{align}\label{eq:diff_f_A_f_B}
    & ~ \| f( \A )_{j_1} - f( \B )_{j_1} \|_2 \notag \\
    = & ~ \| \alpha( \A )_{j_1}^{-1} u( \A )_{j_1} - \alpha( \B )_{j_1}^{-1} u( \B )_{j_1} \|_2 \notag \\
    \leq & ~ \| \alpha( \A )_{j_1}^{-1} u( \A )_{j_1} - \alpha( \B )_{j_1}^{-1} u(\A )_{j_1} \|_2 + \| \alpha(\B )_{j_1}^{-1} u(\A)_{j_1} - \alpha( \B )_{j_1}^{-1} u( \B )_{j_1} \|_2 \notag \\
    \leq & ~ |\alpha( \A )_{j_1}^{-1} - \alpha( \B )_{j_1}^{-1} | \cdot \| u( \A )_{j_1} \|_2 + | \alpha(\B )_{j_1}^{-1} | \cdot \| u(\A )_{j_1} -u(\B )_{j_1} \|_2
\end{align}
where the first step uses Definition~\ref{def:f}, the 2nd and the last step follows by Fact~\ref{fac:matrix_norm}.

For the first term of the Eq.~\eqref{eq:diff_f_A_f_B}, it follows that
\begin{align} \label{eq:diff_f_p1:A}
& ~ |\alpha(\A)_{j_1}^{-1} - \alpha( \B )_{j_1}^{-1} | \cdot \| u(\A)_{j_1} \|_2 \nonumber \\
\leq & ~ \beta^{-2} \cdot | \alpha(\A )_{j_1} - \alpha( \B )_{j_1} | \cdot \| u( \A )_{j_1} \|_2 \nonumber \\
\leq & ~ \beta^{-2} \cdot | \alpha( \A )_{j_1} - \alpha( \B )_{j_1} | \cdot \sqrt{n} \cdot \exp(R^2) \nonumber \\
\leq & ~ \beta^{-2} \cdot \| u( \A )_{j_1} - u( \B )_{j_1} \|_2 \cdot \sqrt{n} \cdot \sqrt{n} \cdot \exp(R^2) \nonumber \\
\leq & ~ \beta^{-2} \cdot 2 \sqrt{n} R \exp(R^2) \cdot \| \A_{[j_1],*} -\B_{[j_1],*} \| \cdot \sqrt{n} \cdot \sqrt{n} \cdot \exp(R^2) \nonumber \\
\leq & ~ 2 \beta^{-2}  n^{1.5} R \exp(2R^2) \cdot \| \A_{[j_1],*} -\B_{[j_1],*} \|
\end{align}
where the first step is derived from Lemma~\ref{lem:lipschitz_alpha_inverse:A}, the 2nd step follows by Lemma~\ref{lem:upper_bound:u}, the 3rd step is owing to Lemma~\ref{lem:lipschitz_alpha:A}, the 4th step is because of Lemma~\ref{lem:lipschitz_exp:A}, the last step is a rearrangement of step 4.

For the second term of the Eq.~\eqref{eq:diff_f_A_f_B}, we have
\begin{align} \label{eq:diff_f_p2:A}
| \alpha(\B )_{j_1}^{-1} | \cdot \| u( \A )_{j_1} -u( \B )_{j_1} \|_2 \leq & ~ \beta^{-1} \cdot \| u(\A  )_{j_1} -u( \B )_{j_1} \|_2 \nonumber \\
\leq & ~ \beta^{-1} \cdot 2 \sqrt{n} R \exp(R^2) \cdot \| \A_{[j_1],*} - \B_{[j_1],*} \| \nonumber \\
\leq & ~ 2 \beta^{-1} n^{0.5} R \exp(R^2) \cdot \| \A_{[j_1],*} - \B_{[j_1],*} \|
\end{align}
where the first step holds because of the assumption in Lemma~\ref{lem:lipschitz_alpha_inverse:A}, the second step follows from Lemma~\ref{lem:lipschitz_exp:A}.

Putting Eq.~\eqref{eq:diff_f_p1:A} and Eq.~\eqref{eq:diff_f_p2:A} into Eq.~\eqref{eq:diff_f_A_f_B}, we have
\begin{align*}
    \| f(\A)_{j_1} - f(\B )_{j_1} \|_2 \leq & ~ 2 \beta^{-2}  n^{1.5} R \exp(2R^2) \cdot \| \A_{[j_1],*} -\B_{[j_1],*} \| + 2 \beta^{-1} n^{0.5} R \exp(R^2) \cdot \| \A_{[j_1],*} - \B_{[j_1],*} \| \\
    \leq & ~ 4 \beta^{-2} n^{1.5} R \exp(2R^2) \cdot \| \A_{[j_1],*} - \B_{[j_1],*} \|
\end{align*}
where the 2nd step is because of Lemma~\ref{lem:lower_bound_A:beta}
 
\end{proof}

\subsection{Lipschitz for \texorpdfstring{$h(\A)$}{} function} \label{sec:lipschitz_h:A}

\begin{lemma}
[Lemma G.2 in \cite{ssz23}]\label{lem:lipschitz_h:A}
Provided that the subsequent requirement are satisfied
\begin{itemize}
    \item Let $\A, \B \in \R^{n^2 \times d^2}$ satisfy $\max_{j_1 \in [n]}\| \A_{[j_1],*} \| \leq R$, $\max_{j_1 \in [n]} \| \B_{[j_1],*} \| \leq R$ 
    \item Let $x \in \R^{d^2}$ satisfy that $\| x \|_2 \leq R $
    \item We define $h(x)$ as Definition~\ref{def:h}
    \item $\| {\bf 1}_n - \frac{f(\A)_{j_1}}{f(\B)_{j_1}} \|_\infty \leq 0.1$
\end{itemize}
Then, for all $j_1 \in [n]$ we have
\begin{align*}
    \| h(\A)_{j_1} - h(\B)_{j_1} \|_2 \leq \| f(\A)_{j_1} - f(\B)_{j_1} \|_2
\end{align*}
\end{lemma}

\subsection{Lipschitz for \texorpdfstring{$c(\A)$}{} function}\label{sec:lipschitz_A:c}

\begin{lemma}\label{lem:lipschitz_c:A}
Provided that the subsequent requirement are satisfied
\begin{itemize}
    \item Let $\A, \B \in \R^{n^2 \times d^2}$ satisfy $\max_{j_1 \in [n]}\| \A_{[j_1],*} \| \leq R$, $\max_{j_1 \in [n]} \| \B_{[j_1],*} \| \leq R$ \item Let $\max_{j_1 \in [n]} \| ( \A_{[j_1],*} - \B_{[j_1],*} ) x \|_{\infty} < 0.01$
    \item Let $x \in \R^{d^2}$ satisfy that $\| x \|_2 \leq R $
    \item Let $c(\A)$ be defined as Definition~\ref{def:c}
    \item Let $\max_{j_1 \in [n]} \| b_{[j_1]} \|_2 \leq 1$
    \item Let $\beta$ be the greatest lower bound of $\langle u(\A)_{j_1} , {\bf 1}_n \rangle$
\end{itemize}
Then, for all $j_1 \in [n]$ we have
\begin{itemize}
    \item $\| c( \A )_{j_1} - c( \B )_{j_1} \|_2 \leq 4 \beta^{-2} n^{1.5} R \exp(2R^2) \cdot \| \A_{[j_1],*} - \B_{[j_1],*} \|$
\end{itemize}
\end{lemma}
\begin{proof}
It follows from using Lemma~\ref{lem:lipschitz_f:A} directly.
\end{proof}

\subsection{Lipschitz for \texorpdfstring{$q(\A)$}{} function}\label{sec:lipschitz_A:q}

\begin{lemma} \label{lem:lipschitz_q:A}
Provided that the subsequent requirement are satisfied
\begin{itemize}
    \item Let $\A, \B \in \R^{n^2 \times d^2}$ satisfy $\max_{j_1 \in [n]}\| \A_{[j_1],*} \| \leq R$, $\max_{j_1 \in [n]} \| \B_{[j_1],*} \| \leq R$ \item Let $\max_{j_1 \in [n]} \| ( \A_{[j_1],*} - \B_{[j_1],*} ) x \|_{\infty} < 0.01$
    \item Let $x \in \R^{d^2}$ satisfy that $\| x \|_2 \leq R $
    \item Let $q(\A)$ be defined as Definition~\ref{def:q}
    \item Let $\max_{j_1 \in [n]} \| b_{[j_1]} \|_2 \leq 1$
\end{itemize}
Then, for all $j_1 \in [n]$ we have
\begin{itemize}
    \item $\| q(\A )_{j_1} - q(\B)_{j_1} \|_2 \leq 4 n R \exp(R^2) \cdot \| \A_{[j_1],*} - \B_{[j_1],*} \|$
\end{itemize}
\end{lemma}
\begin{proof}
We have
\begin{align*}
\| q( \A )_{j_1} - q( \B )_{j_1} \|_2 
= & ~ \| u( \A )_{j_1} - \alpha( \A )_{j_1} b_{[j_1]} - (u( \B )_{j_1} - \alpha( \B )_{j_1} b_{[j_1]}) \|_2 \\
\leq & ~  \| u( \A )_{j_1} - u( \B )_{j_1} \|_2 + \| \alpha( \A )_{j_1} b_{[j_1]} - \alpha( \B )_{j_1} b_{[j_1]} \|_2 \\
\leq & ~ \| u( \A )_{j_1} - u( \B )_{j_1} \|_2 + | \alpha(\A )_{j_1} - \alpha(\B)_{j_1} | \cdot \| b_{[j_1]} \|_2 \\
\leq & ~ \| u(\A )_{j_1} - u(\B)_{j_1} \|_2 + | \alpha(\A)_{j_1} - \alpha(\B)_{j_1} | \\
\leq & ~ \| u(\A)_{j_1} - u(\B)_{j_1} \|_2 + \sqrt{n} \| u(\A )_{j_1} - u( \B )_{j_1} \|_2 \\
\leq & ~ 2 \sqrt{n} \cdot \| u( \A )_{j_1} - u( \B )_{j_1} \|_2  \\
\leq & ~ 2 \sqrt{n} \cdot 2 \sqrt{n} R \exp(R^2) \cdot  \| \A_{[j_1],*} - \B_{[j_1],*} \| \\
\leq & ~ 4 n R \exp(R^2) \cdot \| \A_{[j_1],*} - \B_{[j_1],*} \|
\end{align*}
where the 1st step follows by Definition~\ref{def:q}, the 2nd step is due to the triangle inequality, the 3rd step is derived from property of norm, the 4th step uses lemma assumption, the 5th step follows from Lemma~\ref{lem:lipschitz_alpha:A}, the 6th step holds because $n \geq 1$, the 7th step is given by Lemma~\ref{lem:lipschitz_exp:A}, and the last step is a rearrangement of step 6.
\end{proof}

\section{Lipschitz for Softmax Loss Function (Normalized Version)} \label{sec:Lipschitz_L_c}

In this section, we discuss the Lipschitz conditions for function $L_c$ and $\nabla L_c$.

\subsection{Lipschitz for \texorpdfstring{$L_c(x)$}{} function} \label{sec:lipschitz_Lc:x}

\begin{lemma}
Provided that the subsequent requirement are satisfied
\begin{itemize}
    \item Let $x \in \R^{d^2}, y \in \R^{d^2}$ satisfy $\| x \|_2 \leq R$ and $\| y \|_2 \leq R$
    \item Let $\A \in \R^{n^2 \times d^2}$
    \item Let $L_c(x)$ be defined as Definition~\ref{def:L_c}
    \item Let $\max_{j_1 \in [n]} \| \A_{[j_1],*} (y-x) \|_{\infty} < 0.01$
    \item Let $\max_{j_1 \in [n]}\| \A_{[j_1],*} \| \leq R$
    \item Let $\max_{j_1 \in [n]} \| b_{[j_1]} \|_2 \leq 1$
    \item The greatest lower bound of $\langle u(x)_{j_1} , {\bf 1}_n \rangle$ is denoted as $\beta$
    \item $R > 4$ 
\end{itemize}
then, we have
\begin{align*}
    | L_c(x) - L_c(y) | \leq n^{2.5} \exp(5 R^2) \cdot \| x - y \|_2
\end{align*}
\end{lemma}
\begin{proof}
We can show
\begin{align*}
    & ~ | L_c(x) - L_c(y) | \\
    = & ~ \frac{1}{2} \cdot |\sum_{j_1 = 1}^n (\| c(x)_{j_1} \|_2^2 - \| c(y)_{j_1} \|_2^2) | \\ 
    \leq & ~ \frac{1}{2} \cdot \sum_{j_1 = 1}^n | \| c(x)_{j_1} \|_2^2 - \| c(y)_{j_1} \|_2^2 | \\
    \leq & ~ \frac{1}{2} \cdot \sum_{j_1 = 1}^n | \langle c(x)_{j_1}, c(x)_{j_1} \rangle - \langle c(x)_{j_1}, c(y)_{j_1} \rangle + \langle c(x)_{j_1}, c(y)_{j_1} \rangle - \langle c(y)_{j_1}, c(y)_{j_1} \rangle | \\
    \leq & ~ \frac{1}{2} \cdot \sum_{j_1 = 1}^n | \langle c(x)_{j_1}, c(x)_{j_1}  -  c(y)_{j_1} \rangle| + |\langle c(x)_{j_1} -  c(y)_{j_1}, c(y)_{j_1} \rangle | \\
    \leq & ~ \frac{1}{2} \cdot \sum_{j_1=1}^n (\| c(x)_{j_1} \|_2 + \| c(y)_{j_1} \|_2) \cdot \|c(x)_{j_1} - c(y)_{j_1} \|_2 \\
    \leq & ~ \frac{1}{2} \cdot \sum_{j_1=1}^n 4 \|c(x)_{j_1} - c(y)_{j_1} \|_2 \\
    \leq & ~ 2n \cdot 4\beta^{-2} n^{1.5}  R \exp(2 R^2) \cdot \| x - y \|_2 \\
    \leq & ~ 2R n^{2.5} \exp(4 R^2) \cdot \| x - y \|_2 \\
    \leq & ~ n^{2.5} \exp(5 R^2) \cdot \| x - y \|_2
\end{align*}
where the 1st step is from Definition~\ref{def:L_c},
the 2nd step is because of triangle inequality,
the 3rd step is derived from definition of inner product,
the 4th step is from triangle inequality,
the 5th step uses Cauchy-Schwartz inequality,
the 6th step holds, according to Lemma~\ref{lem:upper_bound:c}, the 7th step follows by Lemma~\ref{lem:lipschitz_c:x}, 
the 8th step is due to Lemma~\ref{lem:lower_bound:beta},
the last step is because $ R > 4$. 
\end{proof}

\subsection{Lipschitz for \texorpdfstring{$L_c(\A)$}{} function} \label{sec:lipschitz_L_c:A}

\begin{lemma}
Provided that the subsequent requirement are satisfied
\begin{itemize}
    \item Let $\A, \B \in \R^{n^2 \times d^2}$ satisfy $\max_{j_1 \in [n]}\| \A_{[j_1],*} \| \leq R$, $\max_{j_1 \in [n]} \| \B_{[j_1],*} \| \leq R$ \item Let $\max_{j_1 \in [n]} \| ( \A_{[j_1],*} - \B_{[j_1],*} ) x \|_{\infty} < 0.01$
    \item Let $x \in \R^{d^2}$ satisfy that $\| x \|_2 \leq R $
    \item Let $L_c(\A)$ be defined as Definition~\ref{def:L_c}
    \item Let $\max_{j_1 \in [n]} \| b_{[j_1]} \|_2 \leq 1$
    \item The greatest lower bound of $\langle u(\A)_{j_1} , {\bf 1}_n \rangle$ is denoted as $\beta$ 
    \item $R > 4$ 
    \item Let $\| \A - \B \|_{\infty,2} = \max_{j_1 \in [n]} \| \A_{[j_1],*} - \B_{[j_1],*} \| $
\end{itemize}
then, we have
\begin{align*}
    | L_c(\A) - L_c(\B) | \leq n^{2.5} \exp(5 R^2) \cdot \| \A - \B \|_{\infty,2}
\end{align*}
\end{lemma}
\begin{proof}
We can show
\begin{align*}
    & ~ | L_c(\A) - L_c(\B) | \\
    = & ~ \frac{1}{2} \cdot |\sum_{j_1 = 1}^n (\| c(\A)_{j_1} \|_2^2 - \| c(\B)_{j_1} \|_2^2) | \\ 
    \leq & ~ \frac{1}{2} \cdot \sum_{j_1 = 1}^n | \| c(\A)_{j_1} \|_2^2 - \| c(\B)_{j_1} \|_2^2 | \\
    \leq & ~ \frac{1}{2} \cdot \sum_{j_1 = 1}^n | \langle c(\A)_{j_1}, c(\A)_{j_1} \rangle - \langle c(\A)_{j_1}, c(\B)_{j_1} \rangle + \langle c(\A)_{j_1}, c(\B)_{j_1} \rangle - \langle c(\A)_{j_1}, c(\B)_{j_1} \rangle | \\
    \leq & ~ \frac{1}{2} \cdot \sum_{j_1 = 1}^n | \langle c(\A)_{j_1}, c(\A)_{j_1}  -  c(\B)_{j_1} \rangle| + |\langle c(\A)_{j_1} -  c(\B)_{j_1}, c(\B)_{j_1} \rangle | \\
    \leq & ~ \frac{1}{2} \cdot \sum_{j_1=1}^n (\| c(\A)_{j_1} \|_2 + \| c(\B)_{j_1} \|_2) \cdot \|c(\A)_{j_1} - c(\B)_{j_1} \|_2 \\
    \leq & ~ \frac{1}{2} \cdot \sum_{j_1=1}^n 4 \cdot \|c(\A)_{j_1} - c(\B)_{j_1} \|_2 \\
    \leq & ~ 2n \cdot 4\beta^{-2} n^{1.5}  R \exp(2 R^2) \cdot \| \A - \B \|_{\infty,2} \\
    \leq & ~ 8 n^{2.5} R \exp(4 R^2) \cdot \| \A - \B \|_{\infty,2} \\
    \leq & ~ n^{2.5} \exp(5 R^2) \cdot \| \A - \B \|_{\infty,2}
\end{align*}
where the first step is from the definition of $L_c(\A)$ (see Definition~\ref{def:L_c}),
the 2nd step is due to triangle inequality,
the 3rd step follows from definition of inner product,
the 4th step follows by triangle inequality,
the 5th step is because of Cauchy-Schwartz inequality,
the 6th step follows by Lemma~\ref{lem:upper_bound:c},
the 7th step is due to Lemma~\ref{lem:lipschitz_c:A}, 
the 8th step is because $R\geq 4$,
the 9th step is following after Lemma~\ref{lem:lower_bound_A:beta}. 
\end{proof}

\subsection{Lipschitz for \texorpdfstring{$\nabla L_c(x)$}{} function} \label{sec:lipschitz_grad_Lc:x}

\begin{lemma}\label{lem:lipschitz_grad_L_c:x}
Provided that the subsequent requirement are satisfied
\begin{itemize}
    \item Let $x \in \R^{d^2}, y \in \R^{d^2}$ satisfy $\| x \|_2 \leq R$ and $\| y \|_2 \leq R$
    \item Let $\A \in \R^{n^2 \times d^2}$
    \item Let $L_c(x)$ be defined as Definition~\ref{def:L_c}
    \item Let $\max_{j_1 \in [n]} \| \A_{[j_1],*} (y-x) \|_{\infty} < 0.01$
    \item Let $\max_{j_1 \in [n]}\| \A_{[j_1],*} \| \leq R$
    \item Let $\max_{j_1 \in [n]} \| b_{[j_1]} \|_2 \leq 1$
    \item The greatest lower bound of $\langle u(x)_{j_1} , {\bf 1}_n \rangle$ is denoted as $\beta$
    \item $R > 4$ 
\end{itemize}
then, we have
\begin{align*}
    \| \nabla L_c(x) - \nabla L_c(y) \|_2 \leq d n^2 \exp(5 R^2) \cdot \| x - y \|_2
\end{align*}
\end{lemma}
\begin{proof}
We can show
\begin{align}\label{eq:diff_grad_c}
    & ~ \| \nabla L_c(x) - \nabla L_c(y) \|_2^2 \notag\\
    = & ~ \sum_{i=1}^{d^2} | \frac{\d L_c(x)}{\d x_i} - \frac{\d L_c(y)}{\d y_i} |^2  \notag \\
    = & ~ \sum_{i=1}^{d^2} |\sum_{j_1=1}^n (\langle c(x)_{j_1}, f(x)_{j_1} \circ \A_{[j_1],i} \rangle - \langle c(x)_{j_1}, f(x)_{j_1} \rangle \cdot \langle f(x)_{j_1}, \A_{[j_1],i} \rangle) ~ -  \notag \\
    & ~ \sum_{j_1=1}^n (\langle c(y)_{j_1}, f(y)_{j_1} \circ \A_{[j_1],i} \rangle - \langle c(y)_{j_1}, f(y)_{j_1} \rangle \cdot \langle f(y)_{j_1}, \A_{[j_1],i} \rangle)|^2  \notag \\
    \leq & ~ \sum_{i=1}^{d^2} \sum_{j_1=1}^n ( ~ |\langle c(x)_{j_1}, f(x)_{j_1} \circ \A_{[j_1],i} \rangle - \langle c(y)_{j_1}, f(y)_{j_1} \circ \A_{[j_1],i} \rangle| ~ +  \notag \\
    & ~ |\langle c(x)_{j_1}, f(x)_{j_1} \rangle \cdot \langle f(x)_{j_1}, \A_{[j_1],i} \rangle - \langle c(y)_{j_1}, f(y)_{j_1} \rangle \cdot \langle f(y)_{j_1}, \A_{[j_1],i} \rangle| ~ )^2 
\end{align} 
where the first step follows from the definition of gradient,
the 2nd step follows by {\bf Part 11} of Lemma~\ref{lem:basic_derivatives},
the last step follows by triangle inequality.

For the first part of Eq.~\eqref{eq:diff_grad_c}, because of triangle inequality, we have
\begin{align} \label{eq:diff_c_f_A:x}
    & |\langle c(x)_{j_1}, f(x)_{j_1} \circ \A_{[j_1],i} \rangle - \langle c(y)_{j_1}, f(y)_{j_1} \circ \A_{[j_1],i} \rangle| \nonumber \\
    \leq & ~ |\langle c(x)_{j_1}, f(x)_{j_1} \circ \A_{[j_1],i} \rangle - \langle c(x)_{j_1}, f(y)_{j_1} \circ \A_{[j_1],i} \rangle| ~ + \nonumber \\
    & ~ |\langle c(x)_{j_1}, f(y)_{j_1} \circ \A_{[j_1],i} \rangle - \langle c(y)_{j_1}, f(y)_{j_1} \circ \A_{[j_1],i} \rangle| \notag \\
    := & ~ C_1 + C_2
\end{align}

For the first item ($C_1$) of Eq.~\eqref{eq:diff_c_f_A:x} we have,
\begin{align} \label{eq:diff_c_f_A_first_part:x}
   C_1 = & |\langle c(x)_{j_1}, f(x)_{j_1} \circ \A_{[j_1],i} \rangle - \langle c(x)_{j_1}, f(y)_{j_1} \circ \A_{[j_1],i} \rangle| \nonumber \\
    = & ~ |\langle c(x)_{j_1}, f(x)_{j_1} \circ \A_{[j_1],i} - f(y)_{j_1} \circ \A_{[j_1],i} \rangle| \nonumber \\ 
    = & ~ |\langle c(x)_{j_1}, (f(x)_{j_1} - f(y)_{j_1}) \circ \A_{[j_1],i} \rangle| \nonumber \\ 
    \leq & ~ \| c(x)_{j_1} \|_2 \cdot \| (f(x)_{j_1} - f(y)_{j_1}) \circ \A_{[j_1],i} \|_2 \nonumber \\
    \leq & ~ 2 \| (f(x)_{j_1} - f(y)_{j_1}) \circ \A_{[j_1],i} \|_2 \nonumber \\
    \leq & ~ 2 \| f(x)_{j_1} - f(y)_{j_1} \|_2 \cdot \| \A_{[j_1,i]} \|_\infty \nonumber \\
    \leq & ~ 2 \| f(x)_{j_1} - f(y)_{j_1} \|_2 \cdot \| \A_{[j_1,i]} \|_2 \nonumber \\
    \leq & ~ 2R \cdot \| f(x)_{j_1} - f(y)_{j_1} \|_2 \nonumber \\
    \leq & ~ 8R^2 \beta^{-2} n^{1.5} \exp(2 R^2) \cdot \| x - y \|_2
\end{align}
where the 1st step is inner product calculation, the 2nd is Hadamard product calculation, the 3rd step is due to Cauchy-Schwartz inequality, the 4th step follows by Lemma~\ref{lem:upper_bound:c}, the 5th step and the 6th step follow from Fact~\ref{fac:vector_norm}, the 7th step holds because of lemma assumptions, the last step is given by Lemma~\ref{lem:lipschitz_f:x}.

Next, we can upper bound the second term ($C_2$),
\begin{align} \label{eq:diff_c_f_A_part_two:x}
  C_2 =  & ~ |\langle c(x)_{j_1}, f(y)_{j_1} \circ \A_{[j_1],i} \rangle - \langle c(y)_{j_1}, f(y)_{j_1} \circ \A_{[j_1],i} \rangle| \nonumber \\
    = & ~ |\langle c(x)_{j_1} - c(y)_{j_1}, f(y)_{j_1} \circ \A_{[j_1],i} \rangle| \nonumber \\
     \leq & ~ \| c(x)_{j_1} - c(y)_{j_1} \|_2 \cdot \| f(y)_{j_1} \circ \A_{[j_1],i} \|_2 \nonumber \\
     \leq & ~ \| c(x)_{j_1} - c(y)_{j_1} \|_2 \cdot \| f(y)_{j_1} \|_2 \cdot \| \A_{[j_1],i} \|_\infty \nonumber \\
     \leq & ~ \| c(x)_{j_1} - c(y)_{j_1} \|_2 \cdot \| f(y)_{j_1} \|_2 \cdot \| \A_{[j_1],i} \|_2 \nonumber \\
     \leq & ~ 4 \beta^{-2} n^{1.5} R \exp(2R^2) \cdot \| x - y \|_2 \cdot \| f(y)_{j_1} \|_2 \cdot \| \A_{[j_1],i} \|_2 \nonumber \\
     \leq & ~ 4\beta^{-2} n^{1.5} R \exp(2R^2) \cdot \| x - y \|_2 \cdot \| \A_{[j_1],i} \|_2 \nonumber \\
     \leq & ~ 4\beta^{-2} R^2 n^{1.5} \exp(2R^2) \cdot \| x - y \|_2
\end{align}
where the first step is inner product calculation,
the 2nd step follows by Cauchy-Schwartz inequality,
the 3rd step and the 4th step are due to Fact~\ref{fac:vector_norm},
the 5th step is given by Lemma~\ref{lem:lipschitz_c:x},
the 6th step is due to Lemma~\ref{lem:upper_bound:f},
the last step follows by $\max_{j_1 \in [n]}\| \A_{[j_1],*} \| \leq R$.

Putting Eq.~\eqref{eq:diff_c_f_A_first_part:x} and Eq.~\eqref{eq:diff_c_f_A_part_two:x} into Eq.~\eqref{eq:diff_c_f_A:x} we have,
\begin{align} \label{eq:diff_c_f_A_whole}
    & |\langle c(x)_{j_1}, f(x)_{j_1} \circ \A_{[j_1],i} \rangle - \langle c(y)_{j_1}, f(y)_{j_1} \circ \A_{[j_1],i} \rangle| \notag \\
    \leq & ~ C_1 + C_2 \notag \\
    \leq & ~ 12\beta^{-2} R^2 n^{1.5} \exp(2R^2) \cdot \| x - y \|_2
\end{align}

For the second item of Eq~\eqref{eq:diff_grad_c}, due to triangle inequality we have
\begin{align} \label{eq:diff_c_f_f_A:x}
    &|\langle c(x)_{j_1}, f(x)_{j_1} \rangle \cdot \langle f(x)_{j_1}, \A_{[j_1],i} \rangle - \langle c(y)_{j_1}, f(y)_{j_1} \rangle \cdot \langle f(y)_{j_1}, \A_{[j_1],i} \rangle| \nonumber \\
    \leq & ~ |\langle c(x)_{j_1}, f(x)_{j_1} \rangle \cdot \langle f(x)_{j_1}, \A_{[j_1],i} \rangle - \langle c(x)_{j_1}, f(x)_{j_1} \rangle \cdot \langle f(y)_{j_1}, \A_{[j_1],i} \rangle| ~ + \nonumber\\
    & ~ |\langle c(x)_{j_1}, f(x)_{j_1} \rangle \cdot \langle f(y)_{j_1}, \A_{[j_1],i} \rangle - \langle c(x)_{j_1}, f(y)_{j_1} \rangle \cdot \langle f(y)_{j_1}, \A_{[j_1],i} \rangle| ~ +\nonumber\\
    & ~ |\langle c(x)_{j_1}, f(y)_{j_1} \rangle \cdot \langle f(y)_{j_1}, \A_{[j_1],i} \rangle - \langle c(y)_{j_1}, f(y)_{j_1} \rangle \cdot \langle f(y)_{j_1}, \A_{[j_1],i} \rangle| \notag \\
    := & ~ C_4 + C_5 + C_6
\end{align}

For the first term ($C_4$) of Eq~\eqref{eq:diff_c_f_f_A:x} we have,
\begin{align} \label{eq:diff_c_f_f_A_p1:x}
    C_4 = & |\langle c(x)_{j_1}, f(x)_{j_1} \rangle \cdot \langle f(x)_{j_1}, \A_{[j_1],i} \rangle - \langle c(x)_{j_1}, f(x)_{j_1} \rangle \cdot \langle f(y)_{j_1}, \A_{[j_1],i} \rangle|\nonumber\\
    = & ~ |\langle c(x)_{j_1}, f(x)_{j_1} \rangle | \cdot |\langle f(x)_{j_1} - f(y)_{j_1}, \A_{[j_1],i} \rangle| \nonumber \\
    \leq & ~ \| c(x)_{j_1} \|_2 \cdot \| f(x)_{j_1} \|_2 \cdot \| f(x)_{j_1} - f(y)_{j_1} \|_2 \cdot \| \A_{[j_1],i} \|_2 \nonumber \\
    \leq & ~ 2 \| f(x)_{j_1} \|_2 \cdot \| f(x)_{j_1} - f(y)_{j_1} \|_2 \cdot \| \A_{[j_1],i} \|_2 \nonumber \\
    \leq & ~ 2 \| f(x)_{j_1} - f(y)_{j_1} \|_2 \cdot \| \A_{[j_1],i} \|_2 \nonumber \\
    \leq & ~ 8 R \beta^{-2} n^{1.5} \exp(2 R^2) \cdot \| x - y \|_2 \cdot \| \A_{[j_1],i} \|_2 \nonumber \\
    \leq & ~ 8 R^2 \beta^{-2} n^{1.5} \exp(2 R^2) \cdot \| x - y \|_2
\end{align}
where the 1st step is inner product calculation, the 2nd step is from Cauchy-Schwartz inequality, the 3rd step follows by Lemma~\ref{lem:upper_bound:c}, the 4th step is given by Lemma~\ref{lem:upper_bound:f}, the 5th step is due to Lemma~\ref{lem:lipschitz_f:x}, the 6th step holds because $\max_{j_1 \in [n]}\| \A_{[j_1],*} \| \leq R$.

For the secomd term ($C_5$) of Eq~\eqref{eq:diff_c_f_f_A:x} we have,
\begin{align} \label{eq:diff_c_f_f_A_p2:x}
    C_5 = & ~ |\langle c(x)_{j_1}, f(x)_{j_1} \rangle \cdot \langle f(y)_{j_1}, \A_{[j_1],i} \rangle - \langle c(x)_{j_1}, f(y)_{j_1} \rangle \cdot \langle f(y)_{j_1}, \A_{[j_1],i} \rangle|\nonumber\\
    = & ~ |\langle c(x)_{j_1}, f(x)_{j_1} - f(y)_{j_1} \rangle | \cdot |\langle f(y)_{j_1}, \A_{[j_1],i} \rangle| \nonumber \\
    \leq & ~ \| c(x)_{j_1} \|_2 \cdot \| f(x)_{j_1} - f(y)_{j_1} \|_2 \cdot \| f(y)_{j_1} \|_2 \cdot \| \A_{[j_1],i} \|_2 \nonumber \\
    \leq & ~ 2 \| f(y)_{j_1} \|_2 \cdot \| f(x)_{j_1} - f(y)_{j_1} \|_2 \cdot \| \A_{[j_1],i} \|_2 \nonumber \\
    \leq & ~ 2 \| f(x)_{j_1} - f(y)_{j_1} \|_2 \cdot \| \A_{[j_1],i} \|_2 \nonumber \\
    \leq & ~ 8 R \beta^{-2} n^{1.5} \exp(2 R^2) \cdot \| x - y \|_2 \cdot \| \A_{[j_1],i} \|_2 \nonumber \\
    \leq & ~ 8 R^2 \beta^{-2} n^{1.5} \exp(2 R^2) \cdot \| x - y \|_2
\end{align}
where the 1st step is inner product calculation, the 2nd step is from Cauchy-Schwartz inequality, the 3rd step follows by Lemma~\ref{lem:upper_bound:c}, the 4th step is given by Lemma~\ref{lem:upper_bound:f}, the 5th step is due to Lemma~\ref{lem:lipschitz_f:x}, the 6th step holds because $\max_{j_1 \in [n]}\| \A_{[j_1],*} \| \leq R$.

For the third term ($C_6$) of Eq~\eqref{eq:diff_c_f_f_A:x} we have,
\begin{align} \label{eq:diff_c_f_f_A_p3:x}
    C_6 = & ~ |\langle c(x)_{j_1}, f(y)_{j_1} \rangle \cdot \langle f(y)_{j_1}, \A_{[j_1],i} \rangle - \langle c(y)_{j_1}, f(y)_{j_1} \rangle \cdot \langle f(y)_{j_1}, \A_{[j_1],i} \rangle|\nonumber\\
    = & ~ |\langle c(x)_{j_1} - c(y)_{j_1}, f(y)_{j_1} \rangle | \cdot |\langle f(y)_{j_1}, \A_{[j_1],i} \rangle| \nonumber \\
    \leq & ~ \| c(x)_{j_1} - c(y)_{j_1} \|_2 \cdot \| f(y)_{j_1} \|_2^2 \cdot \| \A_{[j_1],i} \|_2 \nonumber \\
    \leq & ~ \| c(x)_{j_1} - c(y)_{j_1} \|_2 \cdot \| \A_{[j_1],i} \|_2 \nonumber \\
    \leq & ~ 4 R \beta^{-2} n^{1.5} \exp(2 R^2) \cdot \| x - y \|_2 \cdot \| \A_{[j_1],i} \|_2 \nonumber \\
    \leq & ~ 4 R^2 \beta^{-2} n^{1.5} \exp(2 R^2) \cdot \| x - y \|_2
\end{align}
where the 1st step is inner product calculation, the 2nd step follows from Cauchy-Schwartz inequality, the 3rd  step is due to Lemma~\ref{lem:upper_bound:f}, the 4th step is due to Lemma~\ref{lem:lipschitz_c:x}, the 5th step holds because $\max_{j_1 \in [n]}\| \A_{[j_1],*} \| \leq R$.

Putting Eq.~\eqref{eq:diff_c_f_f_A_p1:x}, Eq.~\eqref{eq:diff_c_f_f_A_p2:x}, and Eq.~\eqref{eq:diff_c_f_f_A_p3:x} into Eq.~\eqref{eq:diff_c_f_f_A:x}, we have
\begin{align} \label{eq:diff_c_f_f_A_whole:x}
    &|\langle c(x)_{j_1}, f(x)_{j_1} \rangle \cdot \langle f(x)_{j_1}, \A_{[j_1],i} \rangle - \langle c(y)_{j_1}, f(y)_{j_1} \rangle \cdot \langle f(y)_{j_1}, \A_{[j_1],i} \rangle| \nonumber \\
    \leq & ~ C_4 + C_5 + C_6 \notag \\
    \leq & ~ 20 R^2 \beta^{-2} n^{1.5} \exp(2 R^2) \cdot \| x - y \|_2
\end{align}

Putting Eq.~\eqref{eq:diff_c_f_A_whole} and Eq.~\eqref{eq:diff_c_f_f_A_whole:x} into Eq.~\eqref{eq:diff_grad_c}, we have
\begin{align*}
     & \| \nabla L_c(x) - \nabla L_c(y) \|_2^2 \\
     \leq & \sum_{i=1}^{d^2} \sum_{j_1=1}^n (32 R^2 \beta^{-2} n^{1.5} \exp(2 R^2) \cdot \| x - y \|_2)^2 \\
     \leq & \sum_{i=1}^{d^2} \sum_{j_1=1}^n (32 R^2  n^{1.5} \exp(4 R^2) \cdot \| x - y \|_2)^2 \\
     \leq & \sum_{i=1}^{d^2} \sum_{j_1=1}^n (n^{1.5} \exp(2 R^2) \cdot \| x - y \|_2)^2 \\
     = & ~ nd^2 (n^{1.5} \exp(5 R^2) \cdot \| x - y \|_2)^2
\end{align*}
where the second step is from Lemma~\ref{lem:lower_bound:beta}, the 3rd step is given by $R>4$.

Therefore, we have
\begin{align*}
     \| \nabla L_c(x) - \nabla L_c(y) \|_2 \leq d n^2 \exp(5 R^2) \cdot \| x - y \|_2
\end{align*}
The proof is complete.
\end{proof}

\subsection{Lipschitz for \texorpdfstring{$\nabla L_c(\A)$}{} function} \label{sec:lipschitz_grad_L_c:A}

\begin{lemma} \label{lem:lipschitz_grad_L_c:A}
Provided that the subsequent requirement are satisfied
\begin{itemize}
    \item Let $\A, \B \in \R^{n^2 \times d^2}$ satisfy $\max_{j_1 \in [n]}\| \A_{[j_1],*} \| \leq R$, $\max_{j_1 \in [n]} \| \B_{[j_1],*} \| \leq R$ \item Let $\max_{j_1 \in [n]} \| ( \A_{[j_1],*} - \B_{[j_1],*} ) x \|_{\infty} < 0.01$
    \item Let $x \in \R^{d^2}$ satisfy that $\| x \|_2 \leq R $
    \item Let $L_c(\A)$ be defined as Definition~\ref{def:L_c}
    \item Let $\max_{j_1 \in [n]} \| b_{[j_1]} \|_2 \leq 1$
    \item 
    The greatest lower bound of $\langle u(\A)_{j_1} , {\bf 1}_n \rangle$ is denoted as $\beta$
    \item $R > 4$ 
    \item Let $\| \A - \B \|_{\infty,2} = \max_{j_1 \in [n]} \| \A_{[j_1],*} - \B_{[j_1],*} \| $ 
\end{itemize}
then, we have
\begin{align*}
    \| \nabla L_c(\A) - \nabla L_c(\B) \|_2 \leq d n^2 \exp(5 R^2) \cdot \| \A - \B \|_{\infty,2}
\end{align*}
\end{lemma}
\begin{proof}
We can show
\begin{align}\label{eq:diff_grad_c:A}
    & ~ \| \nabla L_c(\A) - \nabla L_c(\B) \|_2^2 \notag\\
    = & ~ \sum_{i=1}^{d^2} | \frac{\d L_c(\A)}{\d x_i} - \frac{\d L_c(\B)}{\d x_i} |^2  \notag \\
    = & ~ \sum_{i=1}^{d^2} |\sum_{j_1=1}^n (\langle c(\A)_{j_1}, f(\A)_{j_1} \circ \A_{[j_1],i} \rangle - \langle c(\A)_{j_1}, f(\A)_{j_1} \rangle \cdot \langle f(\A)_{j_1}, \A_{[j_1],i} \rangle) ~ -  \notag \\
    & ~ \sum_{j_1=1}^n (\langle c(\B)_{j_1}, f(\B)_{j_1} \circ \B_{[j_1],i} \rangle - \langle c(\B)_{j_1}, f(\B)_{j_1} \rangle \cdot \langle f(\B)_{j_1}, \B_{[j_1],i} \rangle)|^2  \notag \\
    \leq & ~ \sum_{i=1}^{d^2} \sum_{j_1=1}^n ( ~ |\langle c(\A)_{j_1}, f(\A)_{j_1} \circ \A_{[j_1],i} \rangle - \langle c(\B)_{j_1}, f(\B)_{j_1} \circ \B_{[j_1],i} \rangle| ~ +  \notag \\
    & ~ |\langle c(\A)_{j_1}, f(\A)_{j_1} \rangle \cdot \langle f(\A)_{j_1}, \A_{[j_1],i} \rangle - \langle c(\B)_{j_1}, f(\B)_{j_1} \rangle \cdot \langle f(\B)_{j_1}, \B_{[j_1],i} \rangle| ~ )^2 
\end{align} 
where the first step is because of the definition of gradient,
the 2nd step is due to {\bf Part 11} of Lemma~\ref{lem:basic_derivatives},
the last step is based on triangle inequality.

For the first item of Eq.~\eqref{eq:diff_grad_c:A}, according to triangle inequality, we have
\begin{align} \label{eq:diff_c_f_A:A}
    & |\langle c(\A)_{j_1}, f(\A)_{j_1} \circ \A_{[j_1],i} \rangle - \langle c(\B)_{j_1}, f(\B)_{j_1} \circ \B_{[j_1],i} \rangle| \nonumber \\
    \leq & ~ |\langle c(\A)_{j_1}, f(\A)_{j_1} \circ \A_{[j_1],i} \rangle - \langle c(\A)_{j_1}, f(\A)_{j_1} \circ \B_{[j_1],i} \rangle| ~ + \nonumber \\
    & ~ |\langle c(\A)_{j_1}, f(\A)_{j_1} \circ \B_{[j_1],i} \rangle - \langle c(\A)_{j_1}, f(\B)_{j_1} \circ \B_{[j_1],i} \rangle| ~ + \nonumber \\
    & ~ |\langle c(\A)_{j_1}, f(\B)_{j_1} \circ \B_{[j_1],i} \rangle - \langle c(\A)_{j_1}, f(\B)_{j_1} \circ \B_{[j_1],i} \rangle | \notag \\
    := & ~ C_1 + C_2 + C_3
\end{align}

For the first item ($C_1$) of Eq.~\ref{eq:diff_c_f_A:A}, we have
\begin{align} \label{eq:diff_C_f_A_p1:A}
    C_1 = & ~ |\langle c(\A)_{j_1}, f(\A)_{j_1} \circ \A_{[j_1],i} \rangle - \langle c(\A)_{j_1}, f(\A)_{j_1} \circ \B_{[j_1],i} \rangle| \nonumber \\
    = & ~ |\langle c(\A)_{j_1}, f(\A)_{j_1} \circ \A_{[j_1],i} - f(\A)_{j_1} \circ \B_{[j_1],i} \rangle| \nonumber \\
    = & ~ |\langle c(\A)_{j_1}, f(\A)_{j_1} \circ (\A_{[j_1],i} - \B_{[j_1],i}) \rangle| \nonumber \\
    \leq & ~ \| c(\A)_{j_1} \|_2 \cdot \| f(\A)_{j_1} \circ ( \A_{[j_1],i} - \B_{[j_1],i} )\|_2 \nonumber \\
    \leq & ~ \| c(\A)_{j_1} \|_2 \cdot \| f(\A)_{j_1} \|_2 \cdot \| \A_{[j_1],i} - \B_{[j_1],i} \|_\infty \nonumber \\
    \leq & ~ \| c(\A)_{j_1} \|_2 \cdot \| f(\A)_{j_1} \|_2 \cdot \| \A_{[j_1],i} - \B_{[j_1],i} \|_2 \nonumber \\
    \leq & ~ 2 \| f(\A)_{j_1} \|_2 \cdot \| \A_{[j_1],i} - \B_{[j_1],i} \|_2 \nonumber \\
    \leq & ~ 2 \| \A_{[j_1],i} - \B_{[j_1],i} \|_2 \nonumber \\
    \leq & ~ 2 \| \A_{[j_1],*} - \B_{[j_1],*} \|
\end{align}
where the 1st is the calculation of inner product, the 2nd step is the calculation of Hadamard product, the 3rd step is because of Cauchy-Schwartz inequality, the 4th step and the 5th step are given by Fact~\ref{fac:vector_norm}, the 6th step is derived from Lemma~\ref{lem:upper_bound:c}, the 7th step is derived from Lemma~\ref{lem:upper_bound:f}, the last step is due to the definition of matrix norm.

For the second item ($C_2$) of Eq.~\eqref{eq:diff_c_f_A:A}, we have
\begin{align} \label{eq:diff_C_f_A_p2:A}
    C_2 = & ~ |\langle c(\A)_{j_1}, f(\A)_{j_1} \circ \B_{[j_1],i} \rangle - \langle c(\A)_{j_1}, f(\B)_{j_1} \circ \B_{[j_1],i} \rangle| \nonumber \\
    = & ~ |\langle c(\A)_{j_1}, f(\A)_{j_1} \circ \B_{[j_1],i} - f(\B)_{j_1} \circ \B_{[j_1],i} \rangle| \nonumber \\
    = & ~ |\langle c(\A)_{j_1}, (f(\A)_{j_1} -f(\B)_{j_1}) \circ \B_{[j_1],i} \rangle| \nonumber \\
    \leq & ~ \| c(\A)_{j_1} \|_2 \cdot \| (f(\A)_{j_1} - f(\B)_{j_1}) \circ \B_{[j_1],i} \|_2 \nonumber \\
    \leq & ~ \| c(\A)_{j_1} \|_2 \cdot \| f(\A)_{j_1} - f(\B)_{j_1} \|_\infty \cdot \| \B_{[j_1],i} \|_2 \nonumber \\
    \leq & ~ \| c(\A)_{j_1} \|_2 \cdot \| f(\A)_{j_1} -f(\B)_{j_1} \|_2 \cdot \| \B_{[j_1],i} \|_2 \nonumber \\
    \leq & ~ 2 \| f(\A)_{j_1} - f(\B)_{j_1} \|_2 \cdot \| \B_{[j_1],i} \|_2 \nonumber \\
    \leq & ~ 2R \cdot \| f(\A)_{j_1} - f(\B)_{j_1} \|_2 \nonumber \\
    \leq & ~ 8 \beta^{-2} n^{1.5} R^2 \exp(2R^2) \cdot \| \A_{[j_1],*} - \B_{[j_1],*} \|
\end{align}
where the 1st is the calculation of inner product, the 2nd step is the calculation of Hadamard product, the 3rd step is because of Cauchy-Schwartz inequality, the 4th step and the 5th step are given by Fact~\ref{fac:vector_norm}, the 6th step is derived from Lemma~\ref{lem:upper_bound:c}, the 7th step holds because $\max_{j_1 \in [n]} \| \B_{[j_1],*} \| \leq R$, the 8th step is derived from Lemma~\ref{lem:lipschitz_f:A}.

For the third item ($C_3$) of Eq.~\eqref{eq:diff_c_f_A:A}, we have
\begin{align} \label{eq:diff_C_f_A_p3:A}
    C_3 = & ~ |\langle c(\A)_{j_1}, f(\B)_{j_1} \circ \B_{[j_1],i} \rangle - \langle c(\B)_{j_1}, f(\B)_{j_1} \circ \B_{[j_1],i} \rangle| \nonumber \\
    = & ~ |\langle c(\A)_{j_1} - c(\B)_{j_1}, f(\B)_{j_1} \circ \B_{[j_1],i} \rangle| \nonumber \\
    \leq & ~ \| c(\A)_{j_1} - c(\B)_{j_1} \|_2 \cdot \| f(\B)_{j_1} \circ \B_{[j_1],i} \|_2 \nonumber \\
    \leq & ~ \| c(\A)_{j_1} - c(\B)_{j_1} \|_2 \cdot \| f(\B)_{j_1} \|_\infty \cdot \| \B_{[j_1],i} \|_2 \nonumber \\
    \leq & ~ \| c(\A)_{j_1} - c(\B)_{j_1} \|_2 \cdot \| f(\B)_{j_1}) \|_2 \cdot \| \B_{[j_1],i} \|_2 \nonumber \\
    \leq & ~ \| c(\A)_{j_1} - c(\B)_{j_1} \|_2 \cdot \| \B_{[j_1],i} \|_2 \nonumber \\
    \leq & ~ R \cdot \| c(\A)_{j_1} - c(\B)_{j_1} \|_2 \nonumber \\
    \leq & ~ 4 \beta^{-2} n^{1.5} R^2 \exp(2R^2) \cdot \| \A_{[j_1],*} - \B_{[j_1],*} \|
\end{align}
where the 1st is the calculation of inner product, the 2nd step is because of Cauchy-Schwartz inequality, the 3rd step and the 4th step are given by Fact~\ref{fac:vector_norm}, the 5th step is derived from Lemma~\ref{lem:upper_bound:f}, the 6th step holds since $\max_{j_1 \in [n]} \| \B_{[j_1],*} \| \leq R$, the 7th step is derived from Lemma~\ref{lem:lipschitz_c:A}.

Putting Eq.~\eqref{eq:diff_C_f_A_p1:A}, Eq.~\eqref{eq:diff_C_f_A_p2:A}, and Eq.~\eqref{eq:diff_C_f_A_p3:A} into Eq.~\eqref{eq:diff_c_f_A:A}, we have
\begin{align} \label{eq:diff_c_f_A_whole:A}
    & |\langle c(\A)_{j_1}, f(\A)_{j_1} \circ \A_{[j_1],i} \rangle - \langle c(\B)_{j_1}, f(\B)_{j_1} \circ \B_{[j_1],i} \rangle| \nonumber \\
    \leq & ~ C_1 + C_2 + C_3 \notag \\
    \leq & ~ (2 + 12 \beta^{-2} n^{1.5} R^2 \exp(2R^2)) \cdot \| \A_{[j_1],*} - \B_{[j_1],*} \|
\end{align}

For the second item of Eq~\eqref{eq:diff_grad_c:A}, because of the triangle inequality, we have
\begin{align} \label{eq:diff_c_f_f_A:A}
    &|\langle c(\A)_{j_1}, f(\A)_{j_1} \rangle \cdot \langle f(\A)_{j_1}, \A_{[j_1],i} \rangle - \langle c(\B)_{j_1}, f(\B)_{j_1} \rangle \cdot \langle f(\B)_{j_1}, \B_{[j_1],i} \rangle| \nonumber\\
    \leq & ~ |\langle c(\A)_{j_1}, f(\A)_{j_1} \rangle \cdot \langle f(\A)_{j_1}, \A_{[j_1],i} \rangle - \langle c(\A)_{j_1}, f(\A)_{j_1} \rangle \cdot \langle f(\A)_{j_1}, \B_{[j_1],i} \rangle| ~ + \nonumber \\
    & ~ |\langle c(\A)_{j_1}, f(\A)_{j_1} \rangle \cdot \langle f(\A)_{j_1}, \B_{[j_1],i} \rangle - \langle c(\A)_{j_1}, f(\A)_{j_1} \rangle \cdot \langle f(\B)_{j_1}, \B_{[j_1],i} \rangle| ~ + \nonumber \\
    & ~ |\langle c(\A)_{j_1}, f(\A)_{j_1} \rangle \cdot \langle f(\B)_{j_1}, \B_{[j_1],i} \rangle - \langle c(\A)_{j_1}, f(\B)_{j_1} \rangle \cdot \langle f(\B)_{j_1}, \B_{[j_1],i} \rangle| ~ + \nonumber \\
    & ~ |\langle c(\A)_{j_1}, f(\B)_{j_1} \rangle \cdot \langle f(\B)_{j_1}, \B_{[j_1],i} \rangle - \langle c(\B)_{j_1}, f(\B)_{j_1} \rangle \cdot \langle f(\B)_{j_1}, \B_{[j_1],i} \rangle| \notag \\
    := & ~ C_4 + C_5 + C_6 + C_7
\end{align}

For the first item ($C_4$) of Eq~\eqref{eq:diff_c_f_f_A:A}, we have
\begin{align} \label{eq:c_f_f_A_p1:A}
    C_4 = & ~ |\langle c(\A)_{j_1}, f(\A)_{j_1} \rangle \cdot \langle f(\A)_{j_1}, \A_{[j_1],i} \rangle - \langle c(\A)_{j_1}, f(\A)_{j_1} \rangle \cdot \langle f(\A)_{j_1}, \B_{[j_1],i} \rangle| \nonumber \\
    = & ~ |\langle c(\A)_{j_1}, f(\A)_{j_1} \rangle| \cdot |\langle f(\A)_{j_1}, \A_{[j_1],i} - \B_{[j_1],i} \rangle| \nonumber \\
    \leq & ~ \| c(\A)_{j_1} \|_2 \cdot \| f(\A)_{j_1} \|_2^2 \cdot \| \A_{[j_1],i} - \B_{[j_1],i} \|_2 \nonumber \\
    \leq & ~ 2 \| f(\A)_{j_1} \|_2^2 \cdot \| \A_{[j_1],i} - \B_{[j_1],i} \|_2 \nonumber \\
    \leq & ~ 2 \| \A_{[j_1],i} - \B_{[j_1],i} \|_2 \nonumber \\
    \leq & ~ 2 \| \A_{[j_1],*} - \B_{[j_1],*} \|
\end{align}
where the 1st step is the calculation of inner product, the 2nd step follows by Cauchy-Schwartz inequality, the 3rd step is due to Lemma~\ref{lem:upper_bound:c}, the 4th step is because of Lemma~\ref{lem:upper_bound:f}, the 5th step is due to the definition of matrix norm.

For the second item ($C_5$) of Eq~\eqref{eq:diff_c_f_f_A:A}, we have
\begin{align} \label{eq:c_f_f_A_p2:A}
    C_5 = & ~ |\langle c(\A)_{j_1}, f(\A)_{j_1} \rangle \cdot \langle f(\A)_{j_1}, \B_{[j_1],i} \rangle - \langle c(\A)_{j_1}, f(\A)_{j_1} \rangle \cdot \langle f(\B)_{j_1}, \B_{[j_1],i} \rangle| \nonumber \\
    = & ~ |\langle c(\A)_{j_1}, f(\A)_{j_1} \rangle| \cdot |\langle f(\A)_{j_1} - f(\B)_{j_1}, \B_{[j_1],i} \rangle| \nonumber \\
    \leq & ~ \| c(\A)_{j_1} \|_2 \cdot \| f(\A)_{j_1} \|_2 \cdot \| f(\A)_{j_1} - f(\B)_{j_1} \|_2 \cdot \| \B_{[j_1],i} \|_2 \nonumber \\
    \leq & ~ 2 \| f(\A)_{j_1} \|_2 \cdot \| f(\A)_{j_1} - f(\B)_{j_1} \|_2 \cdot \| \B_{[j_1],i} \|_2 \nonumber \\
    \leq & ~ 2 \| f(\A)_{j_1} - f(\B)_{j_1} \|_2 \cdot \| \B_{[j_1],i} \|_2 \nonumber \\
    \leq & ~ 2R \cdot  \| f(\A)_{j_1} - f(\B)_{j_1} \|_2 \nonumber \\
    \leq & ~ 8 R^2 \beta^{-2} \exp(2 R^2) \cdot \| \A_{[j_1],*} - \B_{[j_1],*} \|
\end{align}
where the 1st step is the calculation of inner product, the 2nd step follows by Cauchy-Schwartz inequality, the 3rd step is due to Lemma~\ref{lem:upper_bound:c}, the 4th step is because of Lemma~\ref{lem:upper_bound:f}, the 5th step follows from $\max_{j_1 \in [n]} \| \B_{[j_1],*} \| \leq R$, the last step is due to Lemma~\ref{lem:lipschitz_f:A}.

For the third item ($C_6$) of Eq~\eqref{eq:diff_c_f_f_A:A}, we have
\begin{align} \label{eq:c_f_f_A_p3:A}
    C_6 = & ~ |\langle c(\A)_{j_1}, f(\A)_{j_1} \rangle \cdot \langle f(\B)_{j_1}, \B_{[j_1],i} \rangle - \langle c(\A)_{j_1}, f(\B)_{j_1} \rangle \cdot \langle f(\B)_{j_1}, \B_{[j_1],i} \rangle| \nonumber \\
    = & ~ |\langle c(\A)_{j_1}, f(\A)_{j_1} - f(\B)_{j_1} \rangle| \cdot | \langle f(\B)_{j_1}, \B_{[j_1],i} \rangle| \nonumber \\
    \leq & ~ \| c(\A)_{j_1} \|_2 \cdot \| f(\A)_{j_1} - f(\B)_{j_1} \|_2 \cdot \| f(\B)_{j_1} \|_2 \cdot \| \B_{[j_1],i} \|_2 \nonumber \\
    \leq & ~ 2 \| f(\A)_{j_1} - f(\B)_{j_1} \|_2 \cdot \| f(\B)_{j_1} \|_2 \cdot \| \B_{[j_1],i} \|_2 \nonumber \\
    \leq & ~ 2 \| f(\A)_{j_1} - f(\B)_{j_1} \|_2 \cdot \| \B_{[j_1],i} \|_2 \nonumber \\
    \leq & ~ 2R \cdot  \| f(\A)_{j_1} - f(\B)_{j_1} \|_2 \nonumber \\
    \leq & ~ 8 R^2 \beta{-2} exp(2 R^2) \cdot \| \A_{[j_1],*} - \B_{[j_1],*} \|
\end{align}
where the 1st step is the calculation of inner product, the 2nd step follows by Cauchy-Schwartz inequality, the 3rd step is due to Lemma~\ref{lem:upper_bound:c}, the 4th step is because of Lemma~\ref{lem:upper_bound:f}, the 5th step follows from $\max_{j_1 \in [n]} \| \B_{[j_1],*} \| \leq R$, the last step is due to Lemma~\ref{lem:lipschitz_f:A}.

For the fourth item ($C_7$) of Eq~\eqref{eq:diff_c_f_f_A:A}, we have
\begin{align} \label{eq:c_f_f_A_p4:A}
    C_7 = & ~ |\langle c(\A)_{j_1}, f(\B)_{j_1} \rangle \cdot \langle f(\B)_{j_1}, \B_{[j_1],i} \rangle - \langle c(\B)_{j_1}, f(\B)_{j_1} \rangle \cdot \langle f(\B)_{j_1}, \B_{[j_1],i} \rangle| \nonumber \\
    = & ~ |\langle c(\A)_{j_1} - c(\B)_{j_1},  f(\A)_{j_1}\rangle| \cdot | \langle f(\B)_{j_1}, \B_{[j_1],i} \rangle| \nonumber \\
    \leq & ~ \| c(\A)_{j_1} - c(\B)_{j_1} \|_2 \cdot \| f(\A)_{j_1} \|_2 \cdot \| f(\B)_{j_1} \|_2 \cdot \| \B_{[j_1],i} \|_2 \nonumber \\
    \leq & ~ \| c(\A)_{j_1} - c(\B)_{j_1} \|_2 \cdot \| \B_{[j_1],i} \|_2 \nonumber \\
    \leq & ~ R \cdot \| c(\A)_{j_1} - c(\B)_{j_1} \|_2 \nonumber \\
    \leq & ~ 4 R^2 \beta{-2} exp(2 R^2) \cdot \| \A_{[j_1],*} - \B_{[j_1],*} \|
\end{align}
where the 1st step is the calculation of inner product, the 2nd step follows by Cauchy-Schwartz inequality, the 3rd step is due to Lemma~\ref{lem:upper_bound:f}, the 4th step follows from $\max_{j_1 \in [n]} \| \B_{[j_1],*} \| \leq R$, the last step is due to Lemma~\ref{lem:lipschitz_c:A}.

Putting Eq.~\eqref{eq:c_f_f_A_p1:A}, Eq.~\eqref{eq:c_f_f_A_p2:A}, 
Eq.~\eqref{eq:c_f_f_A_p3:A}, and Eq.~\eqref{eq:c_f_f_A_p4:A} into Eq.~\eqref{eq:diff_c_f_f_A:A}, we have
\begin{align} \label{eq:diff_c_f_f_A_whole:A}
    & |\langle c(\A)_{j_1}, f(\A)_{j_1} \rangle \cdot \langle f(\A)_{j_1}, \A_{[j_1],i} \rangle - \langle c(\A)_{j_1}, f(\A)_{j_1} \rangle \cdot \langle f(\A)_{j_1}, \B_{[j_1],i} \rangle| \nonumber \\
    \leq & ~ C_4 + C_5 + C_6 + C_7 \notag \\
    \leq & ~ (2 + 20 \beta^{-2} n^{1.5} R^2 \exp(2R^2)) \cdot \| \A_{[j_1],*} - \B_{[j_1],*} \|
\end{align}

Putting Eq.~\eqref{eq:diff_c_f_A_whole:A} and Eq.~\eqref{eq:diff_c_f_f_A_whole:A} into Eq.~\eqref{eq:diff_grad_c:A}, we have
\begin{align*}
     \| \nabla L_c(\A) - \nabla L_c(\B) \|_2^2 \leq & \sum_{i=1}^{d^2} \sum_{j_1=1}^n ( (4 + 32 \beta^{-2} n^{1.5} R^2 \exp(2 R^2)) \cdot \| \A_{[j_1],*} - \B_{[j_1],*} \| )^2 \\
     \leq & \sum_{i=1}^{d^2} \sum_{j_1=1}^n ( (4 + 32 n^{1.5} R^2 \exp(4 R^2)) \cdot \| \A_{[j_1],*} - \B_{[j_1],*} \| )^2 \\
     \leq & \sum_{i=1}^{d^2} \sum_{j_1=1}^n ( n^{1.5} \exp(5 R^2)) \cdot \| \A_{[j_1],*} - \B_{[j_1],*} \| )^2 \\
     \leq & ~ nd^2 (n^{1.5} \exp(5 R^2) \cdot \| \A - \B \|_{\infty,2} )^2
\end{align*}
where the second step holds since Lemma~\ref{lem:lower_bound_A:beta}, the 3rd step holds because $R > 4$, the last step is given by the definition of $\| \A - \B \|_{\infty,2}$.

Therefore, we have
\begin{align*}
     \| \nabla L_c(\A) - \nabla L_c(\B) \|_2 \leq d n^2 \exp(5 R^2) \cdot \| \A - \B \|_{\infty,2}
\end{align*}
The proof is complete.
\end{proof}

\section{Lipschitz for Softmax Loss Function (Rescaled Version)} \label{sec:Lipschitz_L_q}

In this section, we discuss the Lipschitz condition for function $L_q$ and $\nabla L_q$.

\subsection{Lipschitz for \texorpdfstring{$L_q(x)$}{} function} \label{sec:lipschitz_Lq:x}

\begin{lemma}
Provided that the subsequent requirement are satisfied
\begin{itemize}
    \item Let $x \in \R^{d^2}, y \in \R^{d^2}$ satisfy $\| x \|_2 \leq R$ and $\| y \|_2 \leq R$
    \item Let $\A \in \R^{n^2 \times d^2}$
    \item We define $L_q(x)$ as Definition~\ref{def:L_q}
    \item Let $\max_{j_1 \in [n]} \| \A_{[j_1],*} (y-x) \|_{\infty} < 0.01$
    \item Let $\max_{j_1 \in [n]}\| \A_{[j_1],*} \| \leq R$
    \item Let $\max_{j_1 \in [n]} \| b_{[j_1]} \|_2 \leq 1$
    \item $R > 4$ 
\end{itemize}
then, we have
\begin{align*}
    | L_q(x) - L_q(y) | \leq  n^3 \cdot \exp(3 R^2) \cdot \| x - y \|_2
\end{align*}
\end{lemma}

\begin{proof}

We can upper bound $| L_q(x) - L_q(y) |$ as follows:
\begin{align*}
    | L_q(x) - L_q(y) | = & ~ \frac{1}{2} \cdot |\sum_{j_1 = 1}^n (\| q(x)_{j_1} \|_2^2 - \| q(y)_{j_1} \|_2^2) | \\ 
    \leq & ~ \frac{1}{2} \cdot \sum_{j_1 = 1}^n | \| q(x)_{j_1} \|_2^2 - \| q(y)_{j_1} \|_2^2 | \\
    \leq & ~ \frac{1}{2} \cdot \sum_{j_1 = 1}^n | \langle q(x)_{j_1}, q(x)_{j_1} \rangle - \langle q(x)_{j_1}, q(y)_{j_1} \rangle + \langle q(x)_{j_1}, q(y)_{j_1} \rangle - \langle q(y)_{j_1}, q(y)_{j_1} \rangle | \\
    \leq & ~ \frac{1}{2} \cdot \sum_{j_1 = 1}^n | \langle q(x)_{j_1}, q(x)_{j_1}  -  q(y)_{j_1} \rangle| + |\langle q(x)_{j_1} -  q(y)_{j_1}, q(y)_{j_1} \rangle | \\
    \leq & ~ \frac{1}{2} \cdot \sum_{j_1=1}^n (\| q(x)_{j_1} \|_2 + \| q(y)_{j_1} \|_2) \cdot \|q(x)_{j_1} - q(y)_{j_1} \|_2 \\
    \leq & ~ \frac{1}{2} \cdot \sum_{j_1=1}^n 4n \exp(R^2) \cdot \|q(x)_{j_1} - q(y)_{j_1} \|_2 \\
    \leq & ~ n \cdot 2n \exp(R^2) \cdot 4nR \exp(R^2) \cdot \| x - y \|_2 \\
    = & ~ 8 n^3 R \exp( 2 R^2) \cdot \| x - y \|_2 \\
    \leq & ~ n^3 \exp(3 R^2) \cdot \| x - y \|_2
\end{align*}
where the first step holds because of the definition of $L_q(x)$ (see Definition~\ref{def:L_q}),
the 2nd step follows by triangle inequality,
the 3rd step follows from definition of inner product,
the 4th step is given by triangle inequality,
the 5th step is from Cauchy-Schwartz inequality,
the 6th step follows from Lemma~\ref{lem:upper_bound:q}, the 7th step is due to Lemma~\ref{lem:Lipschitz_q:x}, 
and the last step is due to $R\geq 4$.
\end{proof}

\subsection{Lipschitz for \texorpdfstring{$L_q(\A)$}{} function} \label{sec:lipschitz_L_q:A}

\begin{lemma}
Provided that the subsequent requirement are satisfied
\begin{itemize}
     \item Let $\A, \B \in \R^{n^2 \times d^2}$ satisfy $\max_{j_1 \in [n]}\| \A_{[j_1],*} \| \leq R$, $\max_{j_1 \in [n]} \| \B_{[j_1],*} \| \leq R$ \item Let $\max_{j_1 \in [n]} \| ( \A_{[j_1],*} - \B_{[j_1],*} ) x \|_{\infty} < 0.01$
    \item Let $x \in \R^{d^2}$ satisfy that $\| x \|_2 \leq R $
    \item We define $L_q(\A)$ as Definition~\ref{def:L_q}
    \item Let $\max_{j_1 \in [n]} \| b_{[j_1]} \|_2 \leq 1$
    \item $R > 4$
    \item Let $\| \A - \B \|_{\infty,2} = \max_{j_1 \in [n]} \| \A_{[j_1],*} - \B_{[j_1],*} \| $
\end{itemize}
it follows that
\begin{align*}
    | L_q(\A) - L_q(\B) | \leq  n^3 \cdot \exp(3 R^2) \cdot \| \A - \B \|_{\infty,2}
\end{align*}
\end{lemma}

\begin{proof}

We can show:
\begin{align*}
    | L_q(\A) - L_q(\B) | = & ~ \frac{1}{2} \cdot |\sum_{j_1 = 1}^n (\| q(\A)_{j_1} \|_2^2 - \| q(\B)_{j_1} \|_2^2) | \\ 
    \leq & ~ \frac{1}{2} \cdot \sum_{j_1 = 1}^n | \| q(\A)_{j_1} \|_2^2 - \| q(\B)_{j_1} \|_2^2 | \\
    \leq & ~ \frac{1}{2} \cdot \sum_{j_1 = 1}^n | \langle q(\A)_{j_1}, q(\A)_{j_1} \rangle - \langle q(\A)_{j_1}, q(\B)_{j_1} \rangle + \langle q(\A)_{j_1}, q(\B)_{j_1} \rangle - \langle q(\A)_{j_1}, q(\B)_{j_1} \rangle | \\
    \leq & ~ \frac{1}{2} \cdot \sum_{j_1 = 1}^n | \langle q(\A)_{j_1}, q(\A)_{j_1}  -  q(\B)_{j_1} \rangle| + |\langle q(\A)_{j_1} -  q(\B)_{j_1}, q(\B)_{j_1} \rangle | \\
    \leq & ~ \frac{1}{2} \cdot \sum_{j_1=1}^n (\| q(\A)_{j_1} \|_2 + \| q(\B)_{j_1} \|_2) \cdot \|q(\A)_{j_1} - q(\B)_{j_1} \|_2 \\
    \leq & ~ \frac{1}{2} \cdot \sum_{j_1=1}^n 2n \exp(R^2) \cdot \|q(\A)_{j_1} - q(\B)_{j_1} \|_2 \\
    \leq & ~ 4 n^3 R \exp( 2 R^2) \cdot \| \A - \B \|_{\infty,2} \\
    \leq & ~ n^3 \exp(3 R^2) \cdot \| \A - \B \|_{\infty,2}
\end{align*}
where the first step follows by the definition of $L_q(\A)$ (see Definition~\ref{def:L_q}),
the 2nd step is due to triangle inequality,
the 3rd step follows by definition of inner product,
the 4th step is from triangle inequality,
the 5th step follows from Cauchy-Schwartz inequality,
the 6th step follows from Lemma~\ref{lem:upper_bound:q}, the 7th step is because of Lemma~\ref{lem:lipschitz_q:A}, 
and the 8th step holds since $R\geq 4$.
\end{proof}

\subsection{Lipschitz for \texorpdfstring{$\nabla L_q(x)$}{} function} \label{sec:lipschitz_grad_Lq:x}

\begin{lemma}\label{lem:lipschitz_grad_L_q:x}
Provided that the subsequent requirement are satisfied
\begin{itemize}
    \item Let $x \in \R^{d^2}, y \in \R^{d^2}$ satisfy $\| x \|_2 \leq R$ and $\| y \|_2 \leq R$
    \item Let $\A \in \R^{n^2 \times d^2}$
    \item We define $L_q(x)$ as Definition~\ref{def:L_q}
    \item Let $\max_{j_1 \in [n]} \| \A_{[j_1],*} (y-x) \|_{\infty} < 0.01$
    \item Let $\max_{j_1 \in [n]}\| \A_{[j_1],*} \| \leq R$
    \item Let $\max_{j_1 \in [n]} \| b_{[j_1]} \|_2 \leq 1$
    \item $R > 4$ 
\end{itemize}
then, we have
\begin{align*}
    |\nabla L_q(x) - \nabla L_q(y) | \leq dn^2 \exp(3 R^2) \cdot \| x - y \|_2
\end{align*}
\end{lemma}
\begin{proof}
We can show
\begin{align}\label{eq:diff_grad_q}
    & ~ \| \nabla L_q(x) - \nabla L_q(y) \|_2^2 \notag\\
    = & ~ \sum_{i=1}^{d^2} | \frac{\d L_q(x)}{\d x_i} - \frac{\d L_q(y)}{\d y_i} |^2  \notag \\
    = & ~ \sum_{i=1}^{d^2} |\sum_{j_1=1}^n (\langle q(x)_{j_1}, u(x)_{j_1} \circ \A_{[j_1],i} \rangle - \langle q(x)_{j_1}, b_{[j_1]} \rangle \cdot \langle u(x)_{j_1}, \A_{[j_1],i} \rangle) ~ -  \notag \\
    & ~ \sum_{j_1=1}^n (\langle q(y)_{j_1}, u(y)_{j_1} \circ \A_{[j_1],i} \rangle - \langle q(y)_{j_1}, b_{[j_1]} \rangle \cdot \langle u(y)_{j_1}, \A_{[j_1],i} \rangle)|^2  \notag \\
    \leq & ~ \sum_{i=1}^{d^2} \sum_{j_1=1}^n ( ~ |\langle q(x)_{j_1}, u(x)_{j_1} \circ \A_{[j_1],i} \rangle - \langle q(y)_{j_1}, u(y)_{j_1} \circ \A_{[j_1],i} \rangle| ~ +  \notag \\
    & ~ |\langle q(x)_{j_1}, b_{[j_1]} \rangle \cdot \langle u(x)_{j_1}, \A_{[j_1],i} \rangle - \langle q(y)_{j_1}, b_{[j_1]} \rangle \cdot \langle u(y)_{j_1}, \A_{[j_1],i} \rangle| ~ )^2 
\end{align} 
where the first step follows from the definition of gradient,
the 2nd step follows by {\bf Part 11} of Lemma~\ref{lem:basic_derivatives},
the last step follows by triangle inequality.

For the first term of Eq.~\eqref{eq:diff_grad_q}, according to triangle inequality, we have
\begin{align} \label{eq:diff_q_u_A:x}
    & |\langle q(x)_{j_1}, u(x)_{j_1} \circ \A_{[j_1],i} \rangle - \langle q(y)_{j_1}, u(y)_{j_1} \circ \A_{[j_1],i} \rangle| \nonumber \\
    \leq & ~ |\langle q(x)_{j_1}, u(x)_{j_1} \circ \A_{[j_1],i} \rangle - \langle q(x)_{j_1}, u(y)_{j_1} \circ \A_{[j_1],i} \rangle| ~ + \nonumber \\
    & ~ |\langle q(x)_{j_1}, u(y)_{j_1} \circ \A_{[j_1],i} \rangle - \langle q(y)_{j_1}, u(y)_{j_1} \circ \A_{[j_1],i} \rangle| \notag \\
    := & ~ C_1 + C_2
\end{align}

For the first term $C_1$ of Eq.~\eqref{eq:diff_q_u_A:x}, we have
\begin{align} \label{eq:diff_q_u_A_p1:x}
    C_1 = & ~ |\langle q(x)_{j_1}, u(x)_{j_1} \circ \A_{[j_1],i} \rangle - \langle q(x)_{j_1}, u(y)_{j_1} \circ \A_{[j_1],i} \rangle| \nonumber \\
    = & ~ |\langle q(x)_{j_1}, u(x)_{j_1} \circ \A_{[j_1],i} - u(y)_{j_1} \circ \A_{[j_1],i} \rangle|\nonumber \\
    = & ~ |\langle q(x)_{j_1}, (u(x)_{j_1} - u(y)_{j_1}) \circ \A_{[j_1],i} \rangle| \nonumber \\
    \leq & ~ \| q(x)_{j_1} \|_2 \cdot \| (u(x)_{j_1} - u(y)_{j_1}) \circ \A_{[j_1],i} \|_2 \nonumber \\
    \leq & ~ \| q(x)_{j_1} \|_2 \cdot \| u(x)_{j_1} - u(y)_{j_1} \|_2 \cdot \| \A_{[j_1],i} \|_\infty \nonumber \\
    \leq & ~ \| q(x)_{j_1} \|_2 \cdot \| u(x)_{j_1} - u(y)_{j_1} \|_2 \cdot \| \A_{[j_1],i} \|_2 \nonumber \\
    \leq & ~ 2n \exp(R^2) \cdot \| u(x)_{j_1} - u(y)_{j_1} \|_2 \cdot \| \A_{[j_1],i} \|_2 \nonumber \\
    \leq & ~ 2nR \exp(R^2) \cdot \| u(x)_{j_1} - u(y)_{j_1} \|_2 \nonumber \\
    \leq & ~ 4n^{1.5} R \exp(2R^2) \cdot \| x - y \|_2
\end{align}
where the 1st step is inner product calculation, the 2nd step is Hadamard product calculation, the 3rd step is given by Cauchy-Schwartz inequality, the 4th step and the 5th step are due to Fact~\ref{fac:vector_norm}, the 6th step is derived from Lemma~\ref{lem:upper_bound:q}, the 7th step is according to $\max_{j_1 \in [n]}\| \A_{[j_1],*} \| \leq R$, the last step is following from Lemma~\ref{lem:lipschitz_exp:x}.

For the second term ($C_2$) of Eq.~\eqref{eq:diff_q_u_A:x}, we have
\begin{align} \label{eq:diff_q_u_A_p2:x}
    C_2 = & ~ |\langle q(x)_{j_1}, u(y)_{j_1} \circ \A_{[j_1],i} \rangle - \langle q(y)_{j_1}, u(y)_{j_1} \circ \A_{[j_1],i} \rangle| \nonumber \\
    = & ~ |\langle q(x)_{j_1} - q(y)_{j_1}, u(y)_{j_1} \circ \A_{[j_1],i} \rangle| \nonumber \\
    \leq & ~ \| q(x)_{j_1} - q(y)_{j_1} \|_2 \cdot \| u(y)_{j_1} \circ \A_{[j_1],i} \|_2 \nonumber \\
    \leq & ~ \| q(x)_{j_1} - q(y){j_1} \|_2 \cdot \| u(y)_{j_1} \|_2 \cdot \| \A_{[j_1],i} \|_\infty \nonumber \\
    \leq & ~ \| q(x)_{j_1} - q(y){j_1} \|_2 \cdot \| u(y)_{j_1} \|_2 \cdot \| \A_{[j_1],i} \|_2 \nonumber \\
    \leq & ~ 4nR \exp(R^2) \cdot \| x - y \|_2 \cdot \| u(y)_{j_1} \|_2 \cdot \| \A_{[j_1],i} \|_2 \nonumber \\
    \leq & ~ 4nR^2 \exp(2 R^2) \cdot \| x - y \|_2 \cdot \| u(y)_{j_1} \|_2 \nonumber \\
    \leq & ~ 4n^{1.5} R^2 \exp(2 R^2) \cdot \| x - y \|_2
\end{align}
where the first step is inner product calculation, the 2nd is due to Cauchy-Schwartz inequality, the 3rd and the 4th step holds because of Fact~\ref{fac:vector_norm}, the 5th step is derived after Lemma~\ref{lem:Lipschitz_q:x}, the 6th step follows by $\max_{j_1 \in [n]}\| \A_{[j_1],*} \| \leq R$, the 7th step follows from Lemma~\ref{lem:upper_bound:u}.

Putting Eq.~\eqref{eq:diff_q_u_A_p1:x} and Eq.~\eqref{eq:diff_q_u_A_p2:x} into Eq.~\eqref{eq:diff_q_u_A:x} we have,
\begin{align} \label{eq:diff_q_u_A_whole:x}
    & |\langle q(x)_{j_1}, u(x)_{j_1} \circ \A_{[j_1],i} \rangle - \langle q(y)_{j_1}, u(y)_{j_1} \circ \A_{[j_1],i} \rangle| \nonumber \\
    \leq & ~ C_1 + C_2 \notag \\
    \leq & ~ 8 R^2 n^{1.5} \exp(2R^2) \cdot \| x - y \|_2
\end{align}

For the second term of Eq~\eqref{eq:diff_grad_q}, by triangle inequality, we derive
\begin{align} \label{eq:diff_q_b_u_A:x}
    &|\langle q(x)_{j_1}, b_{[j_1]} \rangle \cdot \langle u(x)_{j_1}, \A_{[j_1],i} \rangle - \langle q(y)_{j_1}, b_{[j_1]} \rangle \cdot \langle u(y)_{j_1}, \A_{[j_1],i} \rangle| \nonumber \\
    \leq & ~ |\langle q(x)_{j_1}, b_{[j_1]} \rangle \cdot \langle u(x)_{j_1}, \A_{[j_1],i} \rangle - \langle q(x)_{j_1}, b_{[j_1]} \rangle \cdot \langle u(y)_{j_1}, \A_{[j_1],i} \rangle| ~ + \nonumber \\
    & ~ |\langle q(x)_{j_1}, b_{[j_1]} \rangle \cdot \langle u(y)_{j_1}, \A_{[j_1],i} \rangle - \langle q(y)_{j_1}, b_{[j_1]} \rangle \cdot \langle u(y)_{j_1}, \A_{[j_1],i} \rangle| \notag \\
    := & ~ C_3 + C_4
\end{align}

For the first term ($C_3$) of Eq.~\eqref{eq:diff_q_b_u_A:x}, we have
\begin{align} \label{eq:diff_q_b_u_A_p1:x}
    C_3 = & ~ |\langle q(x)_{j_1}, b_{[j_1]} \rangle \cdot \langle u(x)_{j_1}, \A_{[j_1],i} \rangle - \langle q(x)_{j_1}, b_{[j_1]} \rangle \cdot \langle u(y)_{j_1}, \A_{[j_1],i} \rangle| \nonumber \\
    = & ~ |\langle q(x)_{j_1}, b_{[j_1]} \rangle | \cdot |\langle u(x)_{j_1} - u(y)_{j_1}, \A_{[j_1],i} \rangle| \nonumber \\
    \leq & ~ \| q(x)_{j_1} \|_2 \cdot \| b_{[j_1]} \|_2 \cdot \| u(x)_{j_1} - u(y)_{j_1} \|_2 \cdot \| \A_{[j_1],i} \|_2 \nonumber \\
    \leq & ~ \| q(x)_{j_1} \|_2 \cdot \| u(x)_{j_1} - u(y)_{j_1} \|_2 \cdot \| \A_{[j_1],i} \|_2 \nonumber \\
    \leq & ~ R \cdot \| q(x)_{j_1} \|_2 \cdot \| u(x)_{j_1} - u(y)_{j_1} \|_2 \nonumber \\
    \leq & ~ 2nR \exp(R^2) \| u(x)_{j_1} - u(y)_{j_1} \|_2 \nonumber \\
    \leq & ~ 4n^{1.5} R^2 \exp(2 R^2) \| x - y \|_2
\end{align}
where the fisrt step is inner product calculation, the 2nd step is beacause of Cauchy-Schwartz inequality, the 3rd step is given by $\max_{j_1 \in [n]} \| b_{[j_1]} \|_2 \leq 1$. the 4th step follows by $\max_{j_1 \in [n]}\| \A_{[j_1],*} \| \leq R$, the 5th step is due to Lemma~\ref{lem:upper_bound:q}, the 6th step is from Lemma~\ref{lem:lipschitz_exp:x}.

For the second term($C_4$) of Eq.~\eqref{eq:diff_q_b_u_A:x}, we have
\begin{align} \label{eq:diff_q_b_u_A_p2:x}
    C_4 = & ~ |\langle q(x)_{j_1}, b_{[j_1]} \rangle \cdot \langle u(y)_{j_1}, \A_{[j_1],i} \rangle - \langle q(y)_{j_1}, b_{[j_1]} \rangle \cdot \langle u(y)_{j_1}, \A_{[j_1],i} \rangle| \nonumber \\
    = & ~ |\langle q(x)_{j_1} - q(y)_{j_1}, b_{[j_1]} \rangle | \cdot |\langle u(y)_{j_1}, \A_{[j_1],i} \rangle| \nonumber \\
    \leq & ~ \| q(x)_{j_1} - q(y)_{j_1} \|_2 \cdot \| b_{[j_1]} \|_2 \cdot \| u(y)_{j_1} \|_2 \cdot \| \A_{[j_1],i} \|_2 \nonumber \\
    \leq & ~ \| q(x)_{j_1} - q(y)_{j_1} \|_2 \cdot \| u(y)_{j_1} \|_2 \cdot \| \A_{[j_1],i} \|_2 \nonumber \\
    \leq & ~ R \cdot \| q(x)_{j_1} - q(y)_{j_1} \|_2 \cdot \| u(y)_{j_1} \|_2 \nonumber \\
    \leq & ~ 4nR^2 \exp(R^2) \cdot \| x - y \|_2 \cdot \| u(y)_{j_1} \|_2 \nonumber \\
    \leq & ~ 4n^{1.5} R^2 \exp(2 R^2) \| x - y \|_2
\end{align}
where the fisrt step is inner product calculation, the 2nd step is beacause of Cauchy-Schwartz inequality, the 3rd step is given by $\max_{j_1 \in [n]} \| b_{[j_1]} \|_2 \leq 1$. the 4th step follows by $\max_{j_1 \in [n]}\| \A_{[j_1],*} \| \leq R$, the 5th step is due to Lemma~\ref{lem:Lipschitz_q:x}, the 6th step is from Lemma~\ref{lem:upper_bound:u}.

Putting Eq.~\eqref{eq:diff_q_b_u_A_p1:x} and Eq.~\eqref{eq:diff_q_b_u_A_p2:x} into Eq.~\eqref{eq:diff_q_b_u_A:x} we have,
\begin{align} \label{eq:diff_q_b_u_A_whole:x}
    & |\langle q(x)_{j_1}, u(x)_{j_1} \circ \A_{[j_1],i} \rangle - \langle q(y)_{j_1}, u(y)_{j_1} \circ \A_{[j_1],i} \rangle| \nonumber \\
    \leq & ~ C_3 + C_4 \notag \\
    \leq & ~ 8 R^2 n^{1.5} \exp(2R^2) \cdot \| x - y \|_2
\end{align}

Putting Eq.~\eqref{eq:diff_q_u_A_whole:x} and Eq.~\eqref{eq:diff_q_b_u_A_whole:x} into Eq.~\eqref{eq:diff_grad_q} we have,
\begin{align*}
     & \| \nabla L_q(x) - \nabla L_q(y) \|_2^2 \\
     \leq & \sum_{i=1}^{d^2} \sum_{j_1=1}^n (16 R^2 n^{1.5} \exp(2 R^2) \cdot \| x - y \|_2)^2 \\
     \leq & \sum_{i=1}^{d^2} \sum_{j_1=1}^n (n^{1.5} \exp(3 R^2) \cdot \| x - y \|_2)^2 \\
     = & ~ nd^2 \cdot (n^{1.5} \exp(3 R^2) \cdot \| x - y \|_2)^2
\end{align*}
where the second step is because of $ R>4 $.

Therefore, we have
\begin{align*}
     \| \nabla L_q(x) - \nabla L_q(y) \|_2 \leq d n^2 \exp(3 R^2) \cdot \| x - y \|_2
\end{align*}
The proof is complete.
\end{proof}

\subsection{Lipschitz for \texorpdfstring{$\nabla L_q(\A)$}{} function} \label{sec:lipschitz_grad_L_q:A}

\begin{lemma} \label{lem:lipschitz_grad_L_q:A}
Provided that the subsequent requirement are satisfied
\begin{itemize}
    \item Let $\A, \B \in \R^{n^2 \times d^2}$ satisfy $\max_{j_1 \in [n]}\| \A_{[j_1],*} \| \leq R$, $\max_{j_1 \in [n]} \| \B_{[j_1],*} \| \leq R$ 
    \item Let $\max_{j_1 \in [n]} \| ( \A_{[j_1],*} - \B_{[j_1],*} ) x \|_{\infty} < 0.01$
    \item Let $x \in \R^{d^2}$ satisfy that $\| x \|_2 \leq R $
    \item We define $L_q(\A)$ as Definition~\ref{def:L_q}
    \item Let $\max_{j_1 \in [n]} \| b_{[j_1]} \|_2 \leq 1$
    \item $R > 4$ 
    \item Let $\| \A - \B \|_{\infty,2} = \max_{j_1 \in [n]} \| \A_{[j_1],*} - \B_{[j_1],*} \| $ 
\end{itemize}
then, we have
\begin{align*}
    \| \nabla L_q(\A) - \nabla L_q(\B) \|_2 \leq d n^2 \exp(3 R^2) \cdot \| \A - \B \|_{\infty,2}
\end{align*}
\end{lemma}
\begin{proof}
We can show
\begin{align}\label{eq:diff_grad_q:A}
    & ~ \| \nabla L_q(\A) - \nabla L_q(\A) \|_2^2 \notag\\
    = & ~ \sum_{i=1}^{d^2} | \frac{\d L_q(\A)}{\d x_i} - \frac{\d L_q(\B)}{\d x_i} |^2  \notag \\
    = & ~ \sum_{i=1}^{d^2} |\sum_{j_1=1}^n (\langle q(\A)_{j_1}, u(\A)_{j_1} \circ \A_{[j_1],i} \rangle - \langle q(\A)_{j_1}, b_{[j_1]} \rangle \cdot \langle u(\A)_{j_1}, \A_{[j_1],i} \rangle) ~ -  \notag \\
    & ~ \sum_{j_1=1}^n (\langle q(\B)_{j_1}, u(\B)_{j_1} \circ \B_{[j_1],i} \rangle - \langle q(\B)_{j_1}, b_{[j_1]} \rangle \cdot \langle u(\B)_{j_1}, \B_{[j_1],i} \rangle)|^2  \notag \\
    \leq & ~ \sum_{i=1}^{d^2} \sum_{j_1=1}^n ( ~ |\langle q(\A)_{j_1}, u(\A)_{j_1} \circ \A_{[j_1],i} \rangle - \langle q(\B)_{j_1}, u(\B)_{j_1} \circ \B_{[j_1],i} \rangle| ~ +  \notag \\
    & ~ |\langle q(\A)_{j_1}, b_{[j_1]} \rangle \cdot \langle u(\A)_{j_1}, \A_{[j_1],i} \rangle - \langle q(\B)_{j_1}, b_{[j_1]} \rangle \cdot \langle u(\B)_{j_1}, \B_{[j_1],i} \rangle| ~ )^2 
\end{align} 
where the first step is due to the definition of gradient,
the 2nd step is based on {\bf Part 11} of Lemma~\ref{lem:basic_derivatives},
the last step is because of triangle inequality.

For the first term of Eq.~\eqref{eq:diff_grad_q:A}, by the triangle inequality, we have
\begin{align} \label{eq:q_u_A:A}
    & |\langle q(\A)_{j_1}, u(\A)_{j_1} \circ \A_{[j_1],i} \rangle - \langle q(\B)_{j_1}, u(\B)_{j_1} \circ \B_{[j_1],i} \rangle| \nonumber \\
    \leq & ~ |\langle q(\A)_{j_1}, u(\A)_{j_1} \circ \A_{[j_1],i} \rangle - \langle q(\A)_{j_1}, u(\A)_{j_1} \circ \B_{[j_1],i} \rangle| ~ + \nonumber \\
    & ~ |\langle q(\A)_{j_1}, u(\A)_{j_1} \circ \B_{[j_1],i} \rangle - \langle q(\A)_{j_1}, u(\B)_{j_1} \circ \B_{[j_1],i} \rangle| ~ +\nonumber \\
    & ~ |\langle q(\A)_{j_1}, u(\B)_{j_1} \circ \B_{[j_1],i} \rangle - \langle q(\B)_{j_1}, u(\B)_{j_1} \circ \B_{[j_1],i} \rangle| \notag \\
    := & ~ C_1 + C_2 + C_3
\end{align}

For the first item ($C_1$) of Eq.~\eqref{eq:q_u_A:A}, we have
\begin{align} \label{eq:q_u_A_p1:A}
    C_1 = & ~ |\langle q(\A)_{j_1}, u(\A)_{j_1} \circ \A_{[j_1],i} \rangle - \langle q(\A)_{j_1}, u(\A)_{j_1} \circ \B_{[j_1],i} \rangle| \nonumber \\
    = & ~ |\langle q(\A)_{j_1}, u(\A)_{j_1} \circ \A_{[j_1],i} - u(\A)_{j_1} \circ \B_{[j_1],i} \rangle| \nonumber \\
    = & ~ |\langle q(\A)_{j_1}, u(\A)_{j_1} \circ (\A_{[j_1],i} - \B_{[j_1],i}) \rangle| \nonumber \\
    \leq & ~ \| q(\A)_{j_1} \|_2 \cdot \| u(\A)_{j_1} \circ (\A_{[j_1],i} - \B_{[j_1],i}) \|_2 \nonumber \\
    \leq & ~ \| q(\A)_{j_1} \|_2 \cdot \| u(\A)_{j_1} \|_\infty \cdot \| \A_{[j_1],i} - \B_{[j_1],i} \|_2 \nonumber \\
    \leq & ~ \| q(\A)_{j_1} \|_2 \cdot \| u(\A)_{j_1} \|_2 \cdot \| \A_{[j_1],i} - \B_{[j_1],i} \|_2 \nonumber \\
    \leq & ~ 2n \exp(R^2) \cdot \| u(\A)_{j_1} \|_2 \cdot \| \A_{[j_1],i} - \B_{[j_1],i} \|_2 \nonumber \\
    \leq & ~ 2n \exp(R^2) \cdot \sqrt{n} \exp(R^2) \cdot \| \A_{[j_1],i} - \B_{[j_1],i} \|_2 \nonumber \\
    \leq & ~ 2n^{1.5} \exp(2R^2) \cdot \| \A_{[j_1],*} - \B_{[j_1],*} \|
\end{align}
where the 1st step is inner product calculation, the 2nd step is calculation of Hadamard product, the 3rd step follows by Cauchy-Schwartz inequality, the 4th step and the 5th step are due to Fact~\ref{fac:vector_norm}, the 6th step follows from Lemma~\ref{lem:upper_bound:q}, the 7th step is derived from Lemma~\ref{lem:upper_bound:u}, the last step follows after the definition of matrix norm.

For the second term ($C_2$) of Eq.~\eqref{eq:q_u_A:A}, we have
\begin{align} \label{eq:q_u_A_p2:A}
    C_2 = & ~ |\langle q(\A)_{j_1}, u(\A)_{j_1} \circ \B_{[j_1],i} \rangle - \langle q(\A)_{j_1}, u(\B)_{j_1} \circ \B_{[j_1],i} \rangle| \nonumber \\
    = & ~ |\langle q(\A)_{j_1}, u(\A)_{j_1} \circ \B_{[j_1],i} - u(\B)_{j_1} \circ \B_{[j_1],i} \rangle| \nonumber \\
    = & ~ |\langle q(\A)_{j_1}, (u(\A)_{j_1} - u(\B)_{j_1} \circ \B_{[j_1],i} \rangle| \nonumber \\
    \leq & ~ \| q(\A)_{j_1} \|_2 \cdot \| (u(\A)_{j_1} - u(\B)_{j_1}) \circ \B_{[j_1],i} \|_2 \nonumber \\
    \leq & ~ \| q(\A)_{j_1} \|_2 \cdot \| u(\A)_{j_1} - u(\B)_{j_1} \|_\infty \cdot \| \B_{[j_1],i} \|_2 \nonumber \\
    \leq & ~ \| q(\A)_{j_1} \|_2 \cdot \| u(\A)_{j_1} - u(\B)_{j_1} \|_2 \cdot \| \B_{[j_1],i} \|_2 \nonumber \\
    \leq & ~ 2n \exp(R^2) \cdot \| u(\A)_{j_1} - u(\B)_{j_1} \|_2 \cdot \| \B_{[j_1],i} \|_2 \nonumber \\
    \leq & ~ 2n \exp(R^2) R \cdot \| u(\A)_{j_1} - u(\B)_{j_1} \|_2 \nonumber \\
    \leq & ~ 4n^{1.5} R^2 \exp(2R^2) \cdot \| \A_{[j_1],*} - \B_{[j_1],*} \|
\end{align}
where the 1st step is inner product calculation, the 2nd step is calculation of Hadamard product, the 3rd step follows by Cauchy-Schwartz inequality, the 4th step and the 5th step are due to Fact~\ref{fac:vector_norm}, the 6th step follows from Lemma~\ref{lem:upper_bound:q}, the 7th step is derived because $\max_{j_1 \in [n]} \| \B_{[j_1],*} \| \leq R$ , the last step follows after Lemma~\ref{lem:lipschitz_exp:A}.

For the third item ($C_3$) of Eq.~\eqref{eq:q_u_A:A}, we have
\begin{align} \label{eq:q_u_A_p3:A}
    C_3 = & ~ |\langle q(\A)_{j_1}, u(\B)_{j_1} \circ \B_{[j_1],i} \rangle - \langle q(\B)_{j_1}, u(\B)_{j_1} \circ \B_{[j_1],i} \rangle| \nonumber \\
    = & ~ |\langle q(\A)_{j_1} - q(\B)_{j_1}, u(\B)_{j_1} \circ \B_{[j_1],i} \rangle| \nonumber \\
    \leq & ~ \| q(\A)_{j_1} - q(\B)_{j_1} \|_2 \cdot \| u(\A)_{j_1} \circ \B_{[j_1],i} \|_2 \nonumber \\
    \leq & ~ \| q(\A)_{j_1} - q(\B)_{j_1} \|_2 \cdot \| u(\A)_{j_1} \|_\infty \cdot \| \B_{[j_1],i} \|_2 \nonumber \\
    \leq & ~ \| q(\A)_{j_1} - q(\B)_{j_1} \|_2 \cdot \| u(\A)_{j_1} \|_2 \cdot \| \B_{[j_1],i} \|_2 \nonumber \\
    \leq & ~ \sqrt{n} \exp(R^2) \cdot \| q(\A)_{j_1} - q(\B)_{j_1} \|_2 \cdot \| \B_{[j_1],i} \|_2 \nonumber \\
    \leq & ~ \sqrt{n} \exp(R^2) R \cdot \| q(\A)_{j_1} - q(\B)_{j_1} \|_2 \nonumber \\
    \leq & ~ 4n^{1.5} R^2 \exp(2R^2) \cdot \| \A_{[j_1],*} - \B_{[j_1],*} \|
\end{align}
where the 1st step is inner product calculation, the 2nd step follows by Cauchy-Schwartz inequality, the 3rd step and the 4th step are due to Fact~\ref{fac:vector_norm}, the 5th step follows from Lemma~\ref{lem:upper_bound:u}, the 6th step is derived because $\max_{j_1 \in [n]} \| \B_{[j_1],*} \| \leq R$ , the last step follows after Lemma~\ref{lem:lipschitz_q:A}.

Putting Eq.~\eqref{eq:q_u_A_p1:A}, Eq.~\eqref{eq:q_u_A_p2:A}, and
Eq.~\eqref{eq:q_u_A_p3:A} into Eq.~\eqref{eq:q_u_A:A}, we have
\begin{align} \label{eq:q_u_A_whole:A}
    & |\langle q(\A)_{j_1}, u(\A)_{j_1} \circ \A_{[j_1],i} \rangle - \langle q(\B)_{j_1}, u(\B)_{j_1} \circ \B_{[j_1],i} \rangle| \nonumber \\
    \leq & ~ C_1 + C_2 + C_3 \notag \\
    \leq & ~ (2 + 8R^2) \cdot n^{1.5} \exp(2R^2) \cdot \| \A_{[j_1],*} - \B_{[j_1],*} \|
\end{align}

For the second term of Eq~\eqref{eq:diff_grad_q:A}, due to the triangle inequality, we have
\begin{align} \label{eq:q_b_u_A:A}
    & |\langle q(\A)_{j_1}, b_{[j_1]} \rangle \cdot \langle u(\A)_{j_1}, \A_{[j_1],i} \rangle - \langle q(\B)_{j_1}, b_{[j_1]} \rangle \cdot \langle u(\B)_{j_1}, \B_{[j_1],i} \rangle| \nonumber \\
    \leq & ~ |\langle q(\A)_{j_1}, b_{[j_1]} \rangle \cdot \langle u(\A)_{j_1}, \A_{[j_1],i} \rangle - \langle q(\A)_{j_1}, b_{[j_1]} \rangle \cdot \langle u(\A)_{j_1}, \B_{[j_1],i} \rangle| ~ + \nonumber \\
    & ~ |\langle q(\A)_{j_1}, b_{[j_1]} \rangle \cdot \langle u(\A)_{j_1}, \B_{[j_1],i} \rangle - \langle q(\A)_{j_1}, b_{[j_1]} \rangle \cdot \langle u(\B)_{j_1}, \B_{[j_1],i} \rangle| ~ + \nonumber \\
    & ~ |\langle q(\A)_{j_1}, b_{[j_1]} \rangle \cdot \langle u(\B)_{j_1}, \B_{[j_1],i} \rangle - \langle q(\B)_{j_1}, b_{[j_1]} \rangle \cdot \langle u(\B)_{j_1}, \B_{[j_1],i} \rangle| \notag \\
    := & ~ C_4 + C_5 + C_6
\end{align}

For the first term ($C_4$) of Eq.~\eqref{eq:q_b_u_A:A}, we have
\begin{align} \label{eq:q_b_u_A_p1:A}
    C_4 = & ~ |\langle q(\A)_{j_1}, b_{[j_1]} \rangle \cdot \langle u(\A)_{j_1}, \A_{[j_1],i} \rangle - \langle q(\A)_{j_1}, b_{[j_1]} \rangle \cdot \langle u(\A)_{j_1}, \B_{[j_1],i} \rangle| \nonumber \\
    = & ~ |\langle q(\A)_{j_1}, b_{[j_1]} \rangle | \cdot |\langle u(\A)_{j_1}, \A_{[j_1],i} - \B_{[j_1],i} \rangle| \nonumber \\
    \leq & ~ \| q(\A)_{j_1} \|_2 \cdot \| b_{[j_1]} \|_2 \cdot \| u(\A)_{j_1} \|_2 \cdot \| \A_{[j_1],i} - \B_{[j_1],i} \|_2 \nonumber \\
    \leq & ~ 2n \exp(R^2) \cdot \| b_{[j_1]} \|_2 \cdot \| u(\A)_{j_1} \|_2 \cdot \| \A_{[j_1],i} - \B_{[j_1],i} \|_2 \nonumber \\
    \leq & ~ 2n \exp(R^2) \cdot \| u(\A)_{j_1} \|_2 \| \A_{[j_1],i} - \B_{[j_1],i} \|_2 \nonumber \\
    \leq & ~ 2n \exp(R^2) \cdot \sqrt{n} \exp(R^2) \cdot \| \A_{[j_1],i} - \B_{[j_1],i} \|_2 \nonumber \\
    \leq & ~ 2 n^{1.5} \exp(2 R^2) \cdot \| \A_{[j_1],*} - \B_{[j_1],*} \|
\end{align}
where the first step is inner product calculation, the 2nd step is given by Cauchy-Schwartz inequality, the 3rd step is derived by Lemma~\ref{lem:upper_bound:q}, the 4th step is given by $\max_{j_1 \in [n]} \| b_{[j_1]} \|_2 \leq 1$, the 5th step follows by Lemma~\ref{lem:upper_bound:u}, the 6th step follows from the definition of matrix norm.

For the second term ($C_5$) of Eq.~\eqref{eq:q_b_u_A:A}, we have
\begin{align} \label{eq:q_b_u_A_p2:A}
    C_5 = & ~ |\langle q(\A)_{j_1}, b_{[j_1]} \rangle \cdot \langle u(\A)_{j_1}, \B_{[j_1],i} \rangle - \langle q(\A)_{j_1}, b_{[j_1]} \rangle \cdot \langle u(\B)_{j_1}, \B_{[j_1],i} \rangle| \nonumber \\
    = & ~ |\langle q(\A)_{j_1}, b_{[j_1]} \rangle | \cdot |\langle u(\A)_{j_1} - u(\B)_{j_1}, \B_{[j_1],i} \rangle| \nonumber \\
    \leq & ~ \| q(\A)_{j_1} \|_2 \cdot \| b_{[j_1]} \|_2 \cdot \| u(\A)_{j_1} - u(\B)_{j_1} \|_2 \cdot \| \B_{[j_1],i} \|_2 \nonumber \\
    \leq & ~ 2n \exp(R^2) \cdot \| b_{[j_1]} \|_2 \cdot \| u(\A)_{j_1} - u(\B)_{j_1} \|_2 \cdot \| \B_{[j_1],i} \|_2 \nonumber \\
    \leq & ~ 2n \exp(R^2) \cdot \| u(\A)_{j_1} - u(\B)_{j_1} \|_2 \| \B_{[j_1],i} \|_2 \nonumber \\
    \leq & ~ 2n R \exp(R^2) \cdot \| u(\A)_{j_1} - u(\B)_{j_1} \|_2 \nonumber \\
    \leq & ~ 4 R^2 n^{1.5} \exp(2 R^2) \cdot \| \A_{[j_1],*} - \B_{[j_1],*} \|
\end{align}
where the first step is inner product calculation, the 2nd step is given by Cauchy-Schwartz inequality, the 3rd step is derived by Lemma~\ref{lem:upper_bound:q}, the 4th step is given by $\max_{j_1 \in [n]} \| b_{[j_1]} \|_2 \leq 1 $, the 5th step follows by $\max_{j_1 \in [n]} \| \B_{[j_1],*} \| \leq R$ , the 6th step follows from Lemma~\ref{lem:lipschitz_exp:A}.

For the third term ($C_6$) of Eq.~\eqref{eq:q_b_u_A:A}, we have
\begin{align} \label{eq:q_b_u_A_p3:A}
    C_6 = & ~ |\langle q(\A)_{j_1}, b_{[j_1]} \rangle \cdot \langle u(\B)_{j_1}, \B_{[j_1],i} \rangle - \langle q(\B)_{j_1}, b_{[j_1]} \rangle \cdot \langle u(\B)_{j_1}, \B_{[j_1],i} \rangle| \nonumber \\
    = & ~ |\langle q(\A)_{j_1} - q(\B)_{j_1}, b_{[j_1]} \rangle | \cdot |\langle u(\B)_{j_1}, \B_{[j_1],i} \rangle| \nonumber \\
    \leq & ~ \| q(\A)_{j_1} - q(\B)_{j_1} \|_2 \cdot \| b_{[j_1]} \|_2 \cdot \| u(\B)_{j_1} \|_2 \cdot \| \B_{[j_1],i} \|_2 \nonumber \\
    \leq & ~ 4nR \exp(R^2) \cdot \| \A_{[j_1],*} - \B_{[j_1],*} \| \cdot \| b_{[j_1]} \|_2 \cdot \| u(\B)_{j_1} \|_2 \cdot \| \B_{[j_1],i} \|_2 \nonumber \\
    \leq & ~ 4nR \exp(R^2) \cdot \| \A_{[j_1],*} - \B_{[j_1],*} \| \cdot \| u(\B)_{j_1} \|_2 \cdot \| \B_{[j_1],i} \|_2 \nonumber \\
    \leq & ~ 4nR^2 \exp(R^2) \cdot \| \A_{[j_1],*} - \B_{[j_1],*} \| \cdot \| u(\B)_{j_1} \|_2 \nonumber \\
    \leq & ~ 4 R^2 n^{1.5} \exp(2 R^2) \cdot \| \A_{[j_1],*} - \B_{[j_1],*} \|
\end{align}
where the first step is inner product calculation, the 2nd step is given by Cauchy-Schwartz inequality, the 3rd step is derived by Lemma~\ref{lem:lipschitz_q:A}, the 4th step is given by $\max_{j_1 \in [n]} \| b_{[j_1]} \|_2 \leq 1 $, the 5th step follows by $\max_{j_1 \in [n]} \| \B_{[j_1],*} \| \leq R$ , the 6th step follows from Lemma~\ref{lem:upper_bound:u}.

Putting Eq.~\eqref{eq:q_b_u_A_p1:A}, Eq.~\eqref{eq:q_b_u_A_p2:A}, and
Eq.~\eqref{eq:q_b_u_A_p2:A} into Eq.~\eqref{eq:q_b_u_A:A}, we have
\begin{align} \label{eq:q_b_u_A_whole:A}
     & |\langle q(\A)_{j_1}, b_{[j_1]} \rangle \cdot \langle u(\A)_{j_1}, \A_{[j_1],i} \rangle - \langle q(\B)_{j_1}, b_{[j_1]} \rangle \cdot \langle u(\B)_{j_1}, \B_{[j_1],i} \rangle| \nonumber \\
     \leq & ~ C_4 + C_5 + C_6 \notag \\
     \leq & ~ (2 + 8R^2) \cdot n^{1.5} \exp(2R^2) \cdot \| \A_{[j_1],*} - \B_{[j_1],*} \|
\end{align}

Putting Eq.~\eqref{eq:q_u_A_whole:A} and Eq.~\eqref{eq:q_b_u_A_whole:A} into Eq.~\eqref{eq:diff_grad_q:A}, we have
\begin{align*}
     \| \nabla L_q(\A) - \nabla L_q(\B) \|_2^2 \leq & \sum_{i=1}^{d^2} \sum_{j_1=1}^n ((4 + 16R^2) \cdot n^{1.5} \exp(2R^2) \cdot \| \A_{[j_1],*} - \B_{[j_1],*} \| )^2 \\
     \leq & ~ \sum_{i=1}^{d^2} \sum_{j_1=1}^n (n^{1.5} \exp(3 R^2) \cdot \| \A_{[j_1],*} - \B_{[j_1],*} \| )^2 \\
     \leq & ~ nd^2 (n^{1.5} \exp(3 R^2) \cdot \| \A - \B \|_{\infty,2})^2
\end{align*}
where the second step holds since $R > 4$, the last step is given by the definition of $\| \A - \B \|_{\infty,2}$.

Therefore, we have
\begin{align*}
     \| \nabla L_q(\A) - \nabla L_q(\B) \|_2 \leq d n^2 \exp(3 R^2) \cdot \| \A - \B \|_{\infty,2}
\end{align*}
The proof is complete.

\end{proof}

\section{Lipschitz for Sparse Loss Function} \label{sec:lipschitz_L_sparse}

In this Section, we discuss Lipschitz conditions for function $L_{\sparse}$ and $\nabla L_{\sparse}$.

\subsection{Lipschitz for \texorpdfstring{$L_{\sparse}(x)$}{} Function}\label{sec:lipschitz_L_sparse:x}

\begin{lemma}\label{lem:lipschitz_L_sparse:x}
Provided that the subsequent requirement are satisfied
\begin{itemize}
    \item Let $x \in \R^{d^2}, y \in \R^{d^2}$ satisfy $\| x \|_2 \leq R$ and $\| y \|_2 \leq R$
    \item Let $\A \in \R^{n^2 \times d^2}$
    \item Let $L_{\sparse}(x)$ be defined as Definition~\ref{def:L_sparse}
    \item Let $\max_{j_1 \in [n]} \| \A_{[j_1],*} (y-x) \|_{\infty} < 0.01$
    \item Let $\max_{j_1 \in [n]}\| \A_{[j_1],*} \| \leq R$ 
\end{itemize}
Then, for all $j_1 \in [n]$ we have
\begin{align*}
    | L_{\sparse}(x) - L_{\sparse}(y) | \leq 2n^2 R \exp(R^2) \cdot \| x - y \|_2
\end{align*}
\begin{proof}
\begin{align*}
    | L_{\sparse}(x) - L_{\sparse}(y) | 
    \leq & ~ | \sum_{j_1=1}^n \alpha(x)_{j_1} - \sum_{j_1=1}^n \alpha(y)_{j_1} | \\
    \leq & ~ \sum_{j_1=1}^n | \alpha(x)_{j_1} - \alpha(y)_{j_1} | \\
    \leq & ~ \sum_{j_1=1}^n \sqrt{n} \cdot \| u(x)_{j_1} - u(y)_{j_1} \|_2 \\
    \leq & ~ 2n^2 R \exp(R^2) \cdot \| x - y \|_2
\end{align*}
where the 1st step is due to definition of $L_{\sparse}(x)$ (see Definition~\ref{def:L_sparse}), the 2nd step follows by triangle inequality, the 3rd step is from Lemma~\ref{lem:lipschitz_alpha:x}, the last step is because of Lemma~\ref{lem:lipschitz_exp:x}.
\end{proof}
\end{lemma}

\subsection{Lipschitz for \texorpdfstring{$L_{\sparse}(\A)$}{} Function}\label{sec:lipschitz_L_sparse:A}

\begin{lemma}\label{lem:lipschitz_L_sparse:A}
Provided that the subsequent requirement are satisfied
\begin{itemize}
    \item Let $\A, \B \in \R^{n^2 \times d^2}$ satisfy $\max_{j_1 \in [n]}\| \A_{[j_1],*} \| \leq R$, $\max_{j_1 \in [n]} \| \B_{[j_1],*} \| \leq R$ \item Let $\max_{j_1 \in [n]} \| ( \A_{[j_1],*} - \B_{[j_1],*} ) x \|_{\infty} < 0.01$
    \item Let $x \in \R^{d^2}$ satisfy that $\| x \|_2 \leq R $
    \item Let $L_{\sparse}(\A)$ be defined as Definition~\ref{def:L_sparse}
    \item Let $\| \A - \B \|_{\infty,2} = \max_{j_1 \in [n]} \| \A_{[j_1],*} - \B_{[j_1],*} \| $ 
\end{itemize}
Then, for all $j_1 \in [n]$ we have
\begin{align*}
    | L_{\sparse}(\A) - L_{\sparse}(\B) | \leq 2n^2 R \exp(R^2) \cdot \| \A - \B \|_{\infty,2}
\end{align*}
\begin{proof}
\begin{align*}
    | L_{\sparse}(\A) - L_{\sparse}(\B) | 
    \leq & ~ | \sum_{j_1=1}^n \alpha(\A)_{j_1} - \sum_{j_1=1}^n \alpha(\B)_{j_1} | \\
    \leq & ~ \sum_{j_1=1}^n | \alpha(\A)_{j_1} - \alpha(\B)_{j_1} | \\
    \leq & ~ \sum_{j_1=1}^n \sqrt{n} \cdot \| u(\A)_{j_1} - u(\B)_{j_1} \|_2 \\
    \leq & ~ 2n^2 R \exp(R^2) \cdot \| \A - \B \|_{\infty,2}
\end{align*}
where the 1st step is due to definition of $L_{\sparse}(x)$ (see Definition~\ref{def:L_sparse}), the 2nd step follows by triangle inequality, the 3rd step is from Lemma~\ref{lem:lipschitz_alpha:A}, the last step is because of Lemma~\ref{lem:lipschitz_exp:A}.
\end{proof}
\end{lemma}

\subsection{Lipschitz for \texorpdfstring{$\nabla L_{\sparse}(x)$}{} Function}\label{sec:lipschitz_grad_L_sparse:x}

\begin{lemma}\label{lem:lipschitz_grad_L_sparse:x}
Provided that the subsequent requirement are satisfied
\begin{itemize}
    \item Let $x \in \R^{d^2}, y \in \R^{d^2}$ satisfy $\| x \|_2 \leq R$ and $\| y \|_2 \leq R$
    \item Let $\A \in \R^{n^2 \times d^2}$
    \item Let $L_{\sparse}(x)$ be defined as Definition~\ref{def:L_sparse}
    \item Let $\max_{j_1 \in [n]} \| \A_{[j_1],*} (y-x) \|_{\infty} < 0.01$
    \item Let $\max_{j_1 \in [n]}\| \A_{[j_1],*} \| \leq R$ 
\end{itemize}
Then, for all $j_1 \in [n]$, we have
\begin{align*}
    \| \nabla L_{\sparse}(x) - \nabla L_{\sparse}(y) \|_2 \leq 2ndR \exp(R^2) \cdot \| x - y \|_2
\end{align*}
\begin{proof}
\begin{align*}
    \| \nabla L_{\sparse}(x) - \nabla L_{\sparse}(y) \|_2^2 = & ~ \sum_{i=1}^{d^2} | \sum_{j_1=1}^n \langle u(x)_{j_1}, \A_{[j_1],i} \rangle - \sum_{j_1=1}^n \langle u(y)_{j_1}, \A_{[j_1],i} \rangle |^2 \\
    \leq & ~ \sum_{i=1}^{d^2} \sum_{j_1=1}^n | \langle u(x)_{j_1}, \A_{[j_1],i} \rangle - \langle u(y)_{j_1}, \A_{[j_1],i} \rangle |^2 \\
    = & ~ \sum_{i=1}^{d^2} \sum_{j_1=1}^n | \langle u(x)_{j_1} - u(y)_{j_1}, \A_{[j_1],i} \rangle |^2 \\
    \leq & ~ \sum_{i=1}^{d^2} \sum_{j_1=1}^n (\|  u(x)_{j_1} - u(y)_{j_1} \|_2 \cdot \| \A_{[j_1],i} \|_2)^2 \\
    \leq & ~ \sum_{i=1}^{d^2} \sum_{j_1=1}^n (2 \sqrt{n} R \exp(R^2) \cdot \| x - y \|_2 \cdot \| \A_{[j_1],i} \|_2)^2 \\
    \leq & ~ \sum_{i=1}^{d^2} \sum_{j_1=1}^n (2 \sqrt{n} R^2 \exp(R^2) \cdot \| x - y \|_2)^2 \\
    = & ~ d^2n \cdot (2 \sqrt{n} R^2 \exp(R^2) \cdot \| x - y \|_2)^2
\end{align*}
where the 1st step is due to {\bf Part 13} of Lemma~\ref{lem:basic_derivatives}, the 2nd step is from triangle inequality, the 3rd step is inner product calculation, the 4th step uses Cauchy-Schwartz inequality, the 5th step is due to Lemma~\ref{lem:lipschitz_exp:x}, the 6th holds because $\max_{j_1 \in [n]} \| \A_{[j_1],*} \| \leq R$.

Therefore, we have
\begin{align*}
    \| \nabla L_{\sparse}(x) - \nabla L_{\sparse}(y) \|_2 \leq 2ndR \exp(R^2) \cdot \| x - y \|_2
\end{align*}
\end{proof}

\end{lemma}

\subsection{Lipschitz for \texorpdfstring{$\nabla L_{\sparse}(\A)$}{} Function}\label{sec:lipschitz_grad_L_sparse:A}

\begin{lemma}\label{lem:lipschitz_grad_L_sparse:A}
Provided that the subsequent requirement are satisfied
\begin{itemize}
    \item Let $\A, \B \in \R^{n^2 \times d^2}$ satisfy $\max_{j_1 \in [n]}\| \A_{[j_1],*} \| \leq R$, $\max_{j_1 \in [n]} \| \B_{[j_1],*} \| \leq R$ 
    \item Let $\max_{j_1 \in [n]} \| ( \A_{[j_1],*} - \B_{[j_1],*} ) x \|_{\infty} < 0.01$
    \item Let $x \in \R^{d^2}$ satisfy that $\| x \|_2 \leq R $
    \item Let $L_{\sparse}(\A)$ be defined as Definition~\ref{def:L_sparse}
    \item Let $\| \A - \B \|_{\infty,2} = \max_{j_1 \in [n]} \| \A_{[j_1],*} - \B_{[j_1],*} \|$
    \item $R > 4$
\end{itemize}
Then, for all $j_1 \in [n]$ we have
\begin{align*}
\| \nabla L_{\sparse}(\A) - \nabla L_{\sparse}(\B) \|_2 \leq dn \exp(2 R^2) \cdot \| \A - \B \|_{\infty,2}
\end{align*}
\begin{proof}
\begin{align} \label{eq:grad_L_sparse}
    \| \nabla L_{\sparse}(\A) - \nabla L_{\sparse}(\B) \|_2^2 = & ~ \sum_{i=1}^{d^2} | \sum_{j_1=1}^n \langle u(\A)_{j_1}, \A_{[j_1],i} \rangle - \sum_{j_1=1}^n \langle u(\B)_{j_1}, \B_{[j_1],i} \rangle |^2 \notag \\
    \leq & ~ \sum_{i=1}^{d^2} \sum_{j_1=1}^n (|  \langle u(\A)_{j_1}, \A_{[j_1],i} \rangle - \langle u(\A)_{j_1}, \B_{[j_1],i} \rangle | ~ + \notag \\
    & ~ |\langle u(\A)_{j_1}, \B_{[j_1],i} \rangle - \langle u(\B)_{j_1}, \B_{[j_1],i} \rangle|)^2 \notag \\
    := & ~ \sum_{i=1}^{d^2} \sum_{j_1=1}^n (C_1 + C_2)^2 \\
\end{align}
where the 1st step is due to {\bf Part 13} of Lemma~\ref{lem:basic_derivatives}, the 2nd step is from triangle inequality.

For the first item ($C_1$) of Eq.~\eqref{eq:grad_L_sparse} , we have
\begin{align} \label{eq:grad_L_sparse_p1}
    C_1 = & ~ | \langle u(\A)_{j_1}, \A_{[j_1],i} \rangle - \langle u(\A)_{j_1}, \B_{[j_1],i} \rangle | \notag \\
    = & ~ | \langle u(\A)_{j_1}, \A_{[j_1],i} - \B_{[j_1],i} \rangle | \notag \\
    \leq & ~ \| u(\A)_{j_1} \|_2 \cdot \| \A_{[j_1],i} - \B_{[j_1],i} \|_2 \notag \\
    \leq & ~ \sqrt{n} \exp(R^2) \cdot \| \A_{[j_1],i} - \B_{[j_1],i} \|_2 \notag \\
    \leq & ~ \sqrt{n} \exp(R^2) \cdot \| \A_{[j_1],*} - \B_{[j_1],*} \|
\end{align}
where the 1st step is calculation of inner product, the 2nd step is due to Cauchy-Schwartz inequality, the 3rd step follows from Lemma~\ref{lem:upper_bound:u}, the 4th step holds because of definition of matrix norm.

For the second item ($C_2$) of Eq.~\eqref{eq:grad_L_sparse} , we have
\begin{align} \label{eq:grad_L_sparse_p2}
    C_2 = & ~ | \langle u(\A)_{j_1}, \B_{[j_1],i} \rangle - \langle u(\B)_{j_1}, \B_{[j_1],i} \rangle | \notag \\
    = & ~ | \langle u(\A)_{j_1} - u(\B)_{j_1} , \B_{[j_1],i} \rangle | \notag \\
    \leq & ~ \| u(\A)_{j_1} - u(\B)_{j_1} \|_2 \cdot \|  \B_{[j_1],i} \|_2 \notag \\
    \leq & ~ R \cdot \| u(\A)_{j_1} - u(\B)_{j_1} \|_2 \notag \\
    \leq & ~ 2 \sqrt{n} R^2 \exp(R^2) \cdot \| \A_{[j_1],*} - \B_{[j_1],*} \|
\end{align}
where the 1st step is calculation of inner product, the 2nd step is due to Cauchy-Schwartz inequality, the 3rd step follows from $\max_{j_1 \in [n]} \| \B_{[j_1],*} \| \leq R$, the 4th step holds because of Lemma~\ref{lem:lipschitz_exp:A}.

Putting Eq.~\eqref{eq:grad_L_sparse_p1} and Eq.~\eqref{eq:grad_L_sparse_p2} into Eq.~\eqref{eq:grad_L_sparse}, since $R > 4$, we have
\begin{align*}
    \| \nabla L_{\sparse}(\A) - \nabla L_{\sparse}(\B) \|_2^2 \leq & ~ \sum_{i=1}^{d^2} \sum_{j_1=1}^n (C_1 + C_2)^2 \\
    \leq & ~  d^2 n^2 \exp(2 R^2)^2 \cdot \| \A - \B \|_{\infty,2}^2
\end{align*}

Therefore, we have
\begin{align*}
    \| \nabla L_{\sparse}(\A) - \nabla L_{\sparse}(\B) \|_2 \leq dn \exp(2 R^2) \cdot \| \A - \B \|_{\infty,2}
\end{align*}
\end{proof}
\end{lemma}

\section{Lipschitz for Cross Entropy Loss Function} \label{sec:lipschitz_L_cent}

In this Section, we discuss Lipschitz conditions for function $L_{\cent}$ and $\nabla L_{\cent}$.

\subsection{Lipschitz for \texorpdfstring{$L_{\cent}(x)$}{} Function}\label{sec:lipschitz_L_cent:x}

\begin{lemma}\label{lem:lipschitz_L_cent:x}
Provided that the subsequent requirement are satisfied
\begin{itemize}
    \item Let $x \in \R^{d^2}, y \in \R^{d^2}$ satisfy $\| x \|_2 \leq R$ and $\| y \|_2 \leq R$
    \item Let $\A \in \R^{n^2 \times d^2}$
    \item Let $L_{\cent}(x)$ be defined as Definition~\ref{def:L_sparse}
    \item Let $\max_{j_1 \in [n]} \| \A_{[j_1],*} (y-x) \|_{\infty} < 0.01$
    \item Let $\max_{j_1 \in [n]}\| \A_{[j_1],*} \| \leq R$ 
    \item The greatest lower bound of $\langle u(x)_{j_1}, {\bf 1}_n \rangle$ is denoted as $\beta$
    \item $\max_{j_1 \in [n]} \| b_{[j_1]} \|_2 \leq 1$
\end{itemize}
Then, for all $j_1 \in [n]$ we have
\begin{align*}
    | L_{\cent}(x) - L_{\cent}(y) | \leq 4 n^{2.5} R \beta^{-2} \exp(2 R^2) \cdot \| x - y \|_2
\end{align*}
\begin{proof}
\begin{align*}
    | L_{\cent}(x) - L_{\cent}(y) | = & ~ |\sum_{j_1=1}^n \langle f(x)_{j_1}, b_{[j_1]} \rangle - \sum_{j_1=1}^n \langle f(y)_{j_1}, b_{[j_1]} \rangle| \\
    \leq & ~ \sum_{j_1=1}^n | \langle f(x)_{j_1}, b_{[j_1]} \rangle - \langle f(y)_{j_1}, b_{[j_1]} \rangle | \\
    \leq & ~ \sum_{j_1=1}^n | \langle f(x)_{j_1} -  f(y)_{j_1}, b_{[j_1]} \rangle | \\
    \leq & ~ \sum_{j_1=1}^n \| f(x)_{j_1} - f(y)_{j_1} \|_2 \cdot \| b_{[j_1]} \|_2 \\
    \leq & ~ \sum_{j_1=1}^n \| f(x)_{j_1} - f(y)_{j_1} \|_2 \\
    \leq & ~ 4 n^{2.5} R \beta^{-2} \exp(2 R^2) \cdot \| x - y \|_2
\end{align*}
where the 1st step is due to the definition of $L_{\cent}(x)$ (see Definition~\ref{def:L_cent}), the 2nd is because of the triangle inequality, the 3rd step is inner product calculation, the 4th step holds since $\max_{j_1 \in [n]} \| b_{[j_1]} \|_2 \leq 1$, the 5th step is given by Lemma~\ref{lem:lipschitz_f:x}.
 
\end{proof}
\end{lemma}

\subsection{Lipschitz for \texorpdfstring{$L_{\cent}(\A)$}{} Function}\label{sec:lipschitz_L_cent:A}
\begin{lemma}\label{lem:lipschitz_L_cent:A}
Provided that the subsequent requirement are satisfied
\begin{itemize}
    \item Let $\A, \B \in \R^{n^2 \times d^2}$ satisfy $\max_{j_1 \in [n]}\| \A_{[j_1],*} \| \leq R$, $\max_{j_1 \in [n]} \| \B_{[j_1],*} \| \leq R$ \item Let $\max_{j_1 \in [n]} \| ( \A_{[j_1],*} - \B_{[j_1],*} ) x \|_{\infty} < 0.01$
    \item Let $x \in \R^{d^2}$ satisfy that $\| x \|_2 \leq R $
    \item Let $L_{\sparse}(\A)$ be defined as Definition~\ref{def:L_sparse}
    \item The greatest lower bound of $\langle u(\A)_{j_1}, {\bf 1}_n \rangle$ is denoted as $\beta$
    \item $\max_{j_1 \in [n]} \| b_{[j_1]} \|_2 \leq 1$
    \item Let $\| \A - \B \|_{\infty,2} = \max_{j_1 \in [n]}$
\end{itemize}
Then, for all $j_1 \in [n]$ we have
\begin{align*}
    | L_{\cent}(\A) - L_{\cent}(\B) | \leq 4 n^{2.5} R \beta^{-2} \exp(2 R^2) \cdot \| \A - \B \|_{\infty,2}
\end{align*}
\begin{proof}
\begin{align*}
    | L_{\cent}(\A) - L_{\cent}(\B) | = & ~ |\sum_{j_1=1}^n \langle f(\A)_{j_1}, b_{[j_1]} \rangle - \sum_{j_1=1}^n \langle f(\B)_{j_1}, b_{[j_1]} \rangle| \\
    \leq & ~ \sum_{j_1=1}^n | \langle f(\A)_{j_1}, b_{[j_1]} \rangle - \langle f(\B)_{j_1}, b_{[j_1]} \rangle | \\
    \leq & ~ \sum_{j_1=1}^n | \langle f(\A)_{j_1} -  f(\B)_{j_1}, b_{[j_1]} \rangle | \\
    \leq & ~ \sum_{j_1=1}^n \| f(\A)_{j_1} - f(\B)_{j_1} \|_2 \cdot \| b_{[j_1]} \|_2 \\
    \leq & ~ \sum_{j_1=1}^n \| f(\A)_{j_1} - f(\B)_{j_1} \|_2 \\
    \leq & ~ 4 n^{2.5} R \beta^{-2} \exp(2 R^2) \cdot \| \A - \B \|_{\infty,2}
\end{align*}
where the 1st step is due to the definition of $L_{\cent}(x)$ (see Definition~\ref{def:L_cent}), the 2nd is because of the triangle inequality, the 3rd step is inner product calculation, the 4th step holds since $\max_{j_1 \in [n]} \| b_{[j_1]} \|_2 \leq 1$, the 5th step is given by Lemma~\ref{lem:lipschitz_f:A}.
 
\end{proof}
\end{lemma}

\subsection{Lipschitz for \texorpdfstring{$\nabla L_{\cent}(x)$}{} Function}\label{sec:lipschitz_grad_L_cent:x}

\begin{lemma}\label{lem:lipschitz_grad_L_cent:x}
Provided that the subsequent requirement are satisfied
\begin{itemize}
    \item Let $x \in \R^{d^2}, y \in \R^{d^2}$ satisfy $\| x \|_2 \leq R$ and $\| y \|_2 \leq R$
    \item Let $\A \in \R^{n^2 \times d^2}$
    \item Let $L_{\cent}(x)$ be defined as Definition~\ref{def:L_sparse}
    \item Let $\max_{j_1 \in [n]} \| \A_{[j_1],*} (y-x) \|_{\infty} < 0.01$
    \item Let $\max_{j_1 \in [n]}\| \A_{[j_1],*} \| \leq R$ 
    \item The greatest lower bound of $\langle u(x)_{j_1}, {\bf 1}_n \rangle$ is denoted as $\beta$
    \item $\max_{j_1 \in [n]} \| b_{[j_1]} \|_2 \leq 1$
    \item $R > 4$
\end{itemize}
Then, for all $j_1 \in [n]$ we have
\begin{align*}
    \| \nabla L_{\cent}(x) - \nabla L_{\cent}(y) \|_2
    \leq d n^2 \exp(5 R^2) \cdot \| x - y \|_2
\end{align*}
\begin{proof}
\begin{align} \label{eq:grad_L_cent}
    & \| \nabla L_{\cent}(x) - \nabla L_{\cent}(y) \|_2^2 \notag \\
    = & ~ \sum_{i=1}^{d^2} | \sum_{j_1=1}^n (\langle f(x)_{j_1}, b_{[j_1]} \rangle  \cdot \langle f(x)_{j_1}, \A_{[j_1],i}\rangle - \langle f(x)_{j_1} \circ \A_{[j_1],i}, b_{[j_1]} \rangle)\notag \\
    & ~ - \sum_{j_1=1}^n (\langle f(y)_{j_1}, b_{[j_1]} \rangle  \cdot \langle f(y)_{j_1}, \A_{[j_1],i}\rangle - \langle f(y)_{j_1} \circ \A_{[j_1],i}, b_{[j_1]} \rangle) |^2 \notag \\
    \leq & ~ \sum_{i=1}^{d^2} \sum_{j_1=1}^n (| \langle f(x)_{j_1}, b_{[j_1]} \rangle  \cdot \langle f(x)_{j_1}, \A_{[j_1],i}\rangle - \langle f(y)_{j_1}, b_{[j_1]} \rangle  \cdot \langle f(y)_{j_1}, \A_{[j_1],i}\rangle | \notag \\
    & ~ + | \langle f(x)_{j_1} \circ \A_{[j_1],i}, b_{[j_1]} \rangle - \langle f(y)_{j_1} \circ \A_{[j_1],i}, b_{[j_1]} \rangle |)^2 \notag \\
\end{align}

For the first term of Eq.~\eqref{eq:grad_L_cent}, according to triangle inequality:
\begin{align} \label{eq:diff_f_b_f_A}
    & | \langle f(x)_{j_1}, b_{[j_1]} \rangle  \cdot \langle f(x)_{j_1}, \A_{[j_1],i}\rangle - \langle f(y)_{j_1}, b_{[j_1]} \rangle  \cdot \langle f(y)_{j_1}, \A_{[j_1],i}\rangle | \notag \\
    \leq & ~ | \langle f(x)_{j_1}, b_{[j_1]} \rangle  \cdot \langle f(x)_{j_1}, \A_{[j_1],i}\rangle - \langle f(x)_{j_1}, b_{[j_1]} \rangle  \cdot \langle f(y)_{j_1}, \A_{[j_1],i}\rangle | ~ + \notag \\
    & ~ | \langle f(x)_{j_1}, b_{[j_1]} \rangle  \cdot \langle f(y)_{j_1}, \A_{[j_1],i}\rangle - \langle f(y)_{j_1}, b_{[j_1]} \rangle  \cdot \langle f(y)_{j_1}, \A_{[j_1],i}\rangle | \notag \\
    := & ~ C_1 + C_2
\end{align}

For the first item ($C_1$) of Eq.~\eqref{eq:diff_f_b_f_A}, we have
\begin{align} \label{eq:diff_f_b_f_A_p1}
    C_1 = & ~ | \langle f(x)_{j_1}, b_{[j_1]} \rangle  \cdot \langle f(x)_{j_1}, \A_{[j_1],i}\rangle - \langle f(x)_{j_1}, b_{[j_1]} \rangle  \cdot \langle f(y)_{j_1}, \A_{[j_1],i}\rangle | \notag \\
    = & ~ | \langle f(x)_{j_1}, b_{[j_1]} \rangle  \cdot \langle f(x)_{j_1} - f(y)_{j_1}, \A_{[j_1],i}\rangle | \notag \\
    \leq & ~ \| f(x)_{j_1} \|_2 \cdot \| \A_{[j_1],i} \|_2 \cdot \| f(x)_{j_1} - f(y)_{j_1} \|_2 \cdot \| b_{[j_1]} \|_2 \notag \\
    \leq & ~ \| \A_{[j_1],i} \|_2 \cdot \| f(x)_{j_1} - f(y)_{j_1} \|_2 \cdot \| b_{[j_1]} \|_2 \notag \\
    \leq & ~ \| \A_{[j_1],i} \|_2 \cdot \| f(x)_{j_1} - f(y)_{j_1} \|_2 \notag \\
    \leq & ~ R \cdot \| f(x)_{j_1} - f(y)_{j_1} \|_2 \cdot \| b_{[j_1]} \|_2 \notag \\
    \leq & ~ 4n^{1.5} R^2 \beta^{-2} \exp(2 R^2) \cdot \| x - y \|_2
\end{align}
where the 1st step is inner product calculation, the 2nd step uses Cauchy-Schwartz inequality, the 3rd step uses Lemma~\ref{lem:upper_bound:f}, the 4th step is due to $\max_{j_1 \in [n]} \| \A_{[j_1],*} \| \leq R$, the 5th step holds since $\max_{j_1 \in [n]} \| b_{[j_1]} \|_2 < 1$, the 6th step is given by Lemma~\ref{lem:lipschitz_f:x}.

For the first item ($C_2$) of Eq.~\eqref{eq:diff_f_b_f_A}, we have
\begin{align} \label{eq:diff_f_b_f_A_p2}
    C_2 = & ~ | \langle f(x)_{j_1}, b_{[j_1]} \rangle  \cdot \langle f(y)_{j_1}, \A_{[j_1],i}\rangle - \langle f(y)_{j_1}, b_{[j_1]} \rangle  \cdot \langle f(y)_{j_1}, \A_{[j_1],i}\rangle | \notag \\
    = & ~ | \langle f(x)_{j_1} - f(y)_{j_1}, b_{[j_1]} \rangle \cdot \langle f(y)_{j_1} , \A_{[j_1],i}\rangle | \notag \\
    \leq & ~ \| f(y)_{j_1} \|_2 \cdot \| \A_{[j_1],i} \|_2 \cdot \| f(x)_{j_1} - f(y)_{j_1} \|_2 \cdot \| b_{[j_1]} \|_2 \notag \\
    \leq & ~ \| \A_{[j_1],i} \|_2 \cdot \| f(x)_{j_1} - f(y)_{j_1} \|_2 \cdot \| b_{[j_1]} \|_2 \notag \\
    \leq & ~ \| \A_{[j_1],i} \|_2 \cdot \| f(x)_{j_1} - f(y)_{j_1} \|_2 \notag \\
    \leq & ~ R \cdot \| f(x)_{j_1} - f(y)_{j_1} \|_2 \cdot \| b_{[j_1]} \|_2 \notag \\
    \leq & ~ 4n^{1.5} R^2 \beta^{-2} \exp(2 R^2) \cdot \| x - y \|_2
\end{align}
where the 1st step is inner product calculation, the 2nd step uses Cauchy-Schwartz inequality, the 3rd step uses Lemma~\ref{lem:upper_bound:f}, the 4th step is due to $\max_{j_1 \in [n]} \| \A_{[j_1],*} \| \leq R$, the 5th step holds since $\max_{j_1 \in [n]} \| b_{[j_1]} \|_2 < 1$, the 6th step is given by Lemma~\ref{lem:lipschitz_f:x}.

Putting Eq.~\eqref{eq:diff_f_b_f_A_p1} and Eq.~\eqref{eq:diff_f_b_f_A_p2} into Eq.~\eqref{eq:diff_f_b_f_A} yields,
\begin{align} \label{eq:diff_f_b_f_A_whole}
    & | \langle f(x)_{j_1}, b_{[j_1]} \rangle  \cdot \langle f(x)_{j_1}, \A_{[j_1],i}\rangle - \langle f(y)_{j_1}, b_{[j_1]} \rangle  \cdot \langle f(y)_{j_1}, \A_{[j_1],i}\rangle | \notag \\
    \leq & ~ C_1 + C_2 \notag \\
    \leq & ~ 8n^{1.5} R^2 \beta^{-2} \exp(2 R^2) \cdot \| x - y \|_2
\end{align}

For the second term in Eq.~\eqref{eq:grad_L_cent}, we have
\begin{align} \label{eq:diff_f_A_b}
    & | \langle f(x)_{j_1} \circ \A_{[j_1],i}, b_{[j_1]} \rangle - \langle f(y)_{j_1} \circ \A_{[j_1],i}, b_{[j_1]} \rangle | \notag \\
    = & ~ | \langle f(x)_{j_1} \circ \A_{[j_1],i} - f(y)_{j_1} \circ \A_{[j_1],i}, b_{[j_1]} \rangle | \notag \\
    = & ~ | \langle (f(x)_{j_1} - f(y)_{j_1}) \circ \A_{[j_1],i}, b_{[j_1]} \rangle | \notag \\
    \leq & ~ \| (f(x)_{j_1} - f(y)_{j_1}) \circ \A_{[j_1],i} \|_2 \cdot \| b_{[j_1]} \|_2 \notag \\
    \leq & ~ \| (f(x)_{j_1} - f(y)_{j_1}) \circ \A_{[j_1],i} \|_2 \notag \\
    \leq & ~ \| (f(x)_{j_1} - f(y)_{j_1}) \|_\infty \cdot \| \A_{[j_1],i} \|_2 \notag \\
    \leq & ~ \| (f(x)_{j_1} - f(y)_{j_1}) \|_2 \cdot \| \A_{[j_1],i} \|_2 \notag \\
    \leq & ~ R \cdot \| f(x)_{j_1} - f(y)_{j_1} \|_2 \notag \\
    \leq & ~ 4n^{1.5} R^2 \beta^{-2} \exp(2 R^2) \cdot \| x - y \|_2
\end{align}
where the 1st step is inner product calculation, the 2nd step is Hadamard product calculation, the 3rd step is given by Cauchy-Schwartz inequality, the 4th step is because $\max_{j_1 \in [n]} \| b_{[j_1]} \|_2 \leq 1$, the 5th and the 6th step are due to Fact~\ref{fac:vector_norm}, the 7th step holds because $\max_{j_1 \in [n]} \| \A_{[j_1],*} \| \leq R$, the 8th step is from Lemma~\ref{lem:lipschitz_f:x}.

Now we put Eq.~\eqref{eq:diff_f_b_f_A_whole} and Eq.~\eqref{eq:diff_f_A_b} into Eq.~\ref{eq:grad_L_cent},
\begin{align*}
    \| \nabla L_{\cent}(x) - \nabla L_{\cent}(y) \|_2^2 \leq & ~ \sum_{i=1}^{d^2} \sum_{j_1=1}^n (12n^{1.5} R^2 \beta^{-2} \exp(2 R^2) \cdot \| x - y \|_2)^2 \\
    = & ~ d^2 n (12n^{1.5} R^2 \beta^{-2} \exp(2 R^2) \cdot \| x - y \|_2)^2
\end{align*}

Therefore,
\begin{align*}
    \| \nabla L_{\cent}(x) - \nabla L_{\cent}(y) \|_2 \leq & ~ 12 d n^2 R^2 \beta^{-2} \exp(2 R^2) \cdot \| x - y \|_2 \\
    \leq & ~ 12 d n^2 R^2 \exp(4 R^2) \cdot \| x - y \|_2 \\
    \leq & ~ d n^2 \exp(5 R^2) \cdot \| x - y \|_2
\end{align*}
where the 2nd step is derived from Lemma~\ref{lem:lower_bound:beta}, the 3rd step is because $R > 4$.

\end{proof}
\end{lemma}

\subsection{Lipschitz for \texorpdfstring{$\nabla L_{\cent}(\A)$}{} Function}\label{sec:lipschitz_grad_L_cent:A}

\begin{lemma}\label{lem:lipschitz_grad_L_cent:A}
Provided that the subsequent requirement are satisfied
\begin{itemize}
    \item Let $\A, \B \in \R^{n^2 \times d^2}$ satisfy $\max_{j_1 \in [n]}\| \A_{[j_1],*} \| \leq R$, $\max_{j_1 \in [n]} \| \B_{[j_1],*} \| \leq R$ \item Let $\max_{j_1 \in [n]} \| ( \A_{[j_1],*} - \B_{[j_1],*} ) x \|_{\infty} < 0.01$
    \item Let $x \in \R^{d^2}$ satisfy that $\| x \|_2 \leq R $
    \item Let $L_{\sparse}(\A)$ be defined as Definition~\ref{def:L_sparse}
    \item The greatest lower bound of $\langle u(\A)_{j_1}, {\bf 1}_n \rangle$ is denoted as $\beta$
    \item $\max_{j_1 \in [n]} \| b_{[j_1]} \|_2 \leq 1$
    \item Let $\| \A - \B \|_{\infty,2} = \max_{j_1 \in [n]}$
    \item $R > 4$
\end{itemize}
Then, for all $j_1 \in [n]$ we have
\begin{align*}
    \| \nabla L_{\cent}(\A) - \nabla L_{\cent}(\B) \|_2 \leq dn^2 \exp(5 R^2) \cdot \| \A - \B \|_{\infty,2}
\end{align*}
\begin{proof}
\begin{align} \label{eq:grad_L_cent:A}
    & \| \nabla L_{\cent}(\A) - \nabla L_{\cent}(\B) \|_2^2 \notag \\ 
    = & ~ \sum_{i=1}^{d^2} | \sum_{j_1=1}^n (\langle f(\A)_{j_1}, b_{[j_1]} \rangle  \cdot \langle f(\A)_{j_1}, \A_{[j_1],i}\rangle - \langle f(\A)_{j_1} \circ \A_{[j_1],i}, b_{[j_1]} \rangle) \notag \\
    - & ~ \sum_{j_1=1}^n (\langle f(\B)_{j_1}, b_{[j_1]} \rangle  \cdot \langle f(\B)_{j_1}, \B_{[j_1],i}\rangle - \langle f(\B)_{j_1} \circ \B_{[j_1],i}, b_{[j_1]} \rangle)| ^2 \notag \\
    \leq & ~ \sum_{i=1}^{d^2} \sum_{j_1=1}^n (|\langle f(\A)_{j_1}, b_{[j_1]} \rangle  \cdot \langle f(\A)_{j_1}, \A_{[j_1],i}\rangle - \langle f(\B)_{j_1}, b_{[j_1]} \rangle  \cdot \langle f(\B)_{j_1}, \B_{[j_1],i}\rangle | \notag \\
    + & ~| \langle f(\A)_{j_1} \circ \A_{[j_1],i}, b_{[j_1]} \rangle - \langle f(\B)_{j_1} \circ \B_{[j_1],i}, b_{[j_1]} \rangle)|) ^2 \notag \\
\end{align}

For the first term of Eq.~\eqref{eq:grad_L_cent:A}, by triangle inequality, we have
\begin{align} \label{eq:f_b_f_A:A}
    & ~ |\langle f(\A)_{j_1}, b_{[j_1]} \rangle  \cdot \langle f(\A)_{j_1}, \A_{[j_1],i}\rangle - \langle f(\B)_{j_1}, b_{[j_1]} \rangle  \cdot \langle f(\B)_{j_1}, \B_{[j_1],i}\rangle | \notag \\
    \leq & ~ |\langle f(\A)_{j_1}, b_{[j_1]} \rangle  \cdot \langle f(\A)_{j_1}, \A_{[j_1],i}\rangle - \langle f(\A)_{j_1}, b_{[j_1]} \rangle  \cdot \langle f(\A)_{j_1}, \B_{[j_1],i}\rangle | ~ + \notag \\
    & ~ |\langle f(\A)_{j_1}, b_{[j_1]} \rangle  \cdot \langle f(\A)_{j_1}, \B_{[j_1],i}\rangle - \langle f(\A)_{j_1}, b_{[j_1]} \rangle  \cdot \langle f(\B)_{j_1}, \B_{[j_1],i}\rangle | ~ +\notag \\
    & ~ |\langle f(\A)_{j_1}, b_{[j_1]} \rangle  \cdot \langle f(\B)_{j_1}, \B_{[j_1],i}\rangle - \langle f(\B)_{j_1}, b_{[j_1]} \rangle  \cdot \langle f(\B)_{j_1}, \B_{[j_1],i}\rangle | \notag \\
    := & ~ C_1 + C_2 + C_3
\end{align}

For the first term ($C_1$) of Eq.~\eqref{eq:f_b_f_A:A}, we have
\begin{align} \label{eq:f_b_f_A_p1:A}
    C_1 = & ~ |\langle f(\A)_{j_1}, b_{[j_1]} \rangle  \cdot \langle f(\A)_{j_1}, \A_{[j_1],i}\rangle - \langle f(\A)_{j_1}, b_{[j_1]} \rangle  \cdot \langle f(\A)_{j_1}, \B_{[j_1],i}\rangle | \notag \\
    = & ~ |\langle f(\A)_{j_1}, b_{[j_1]} \rangle \cdot \langle f(\A)_{j_1}, \A_{[j_1],i} - \B_{[j_1],i}\rangle | \notag \\
    \leq & ~ \| f(\A)_{j_1} \|_2^2 \cdot \| b_{[j_1]} \|_2 \cdot\| \A_{[j_1],i} - \B_{[j_1],i} \|_2 \notag \\
    \leq & ~ \| b_{[j_1]} \|_2 \cdot\| \A_{[j_1],i} - \B_{[j_1],i} \|_2 \notag \\
    \leq & ~ \| \A_{[j_1],i} - \B_{[j_1],i} \|_2 \notag \\
    \leq & ~ \| \A_{[j_1],*} - \B_{[j_1],*} \|
\end{align}
where the 1st step is inner product calculation, the 2nd step is because of Cauchy-Schwartz inequality, the 3rd step is due to Lemma~\ref{lem:upper_bound:f}, the 4th step is given by $\max_{j_1 \in [n]} \| b_{[j_1]} \|_2 \leq 1$, the last step is due to the definition of matrix norm.

For the second term ($C_2$) of Eq.~\eqref{eq:f_b_f_A:A}, we have
\begin{align} \label{eq:f_b_f_A_p2:A}
    C_2 = & ~ |\langle f(\A)_{j_1}, b_{[j_1]} \rangle  \cdot \langle f(\A)_{j_1}, \B_{[j_1],i}\rangle - \langle f(\A)_{j_1}, b_{[j_1]} \rangle  \cdot \langle f(\B)_{j_1}, \B_{[j_1],i}\rangle | \notag \\
    = & ~ |\langle f(\A)_{j_1}, b_{[j_1]} \rangle \cdot \langle f(\A)_{j_1} - f(\B)_{j_1}, \B_{[j_1],i} \rangle | \notag \\
    \leq & ~ \| f(\A)_{j_1} \|_2 \cdot \| b_{[j_1]} \|_2 \cdot \| f(\A)_{j_1} - f(\B)_{j_1} \|_2 \cdot \| \B_{[j_1],i} \|_2 \notag \\
    \leq & ~ \| b_{[j_1]} \|_2 \cdot \| f(\A)_{j_1} - f(\B)_{j_1} \|_2 \cdot \| \B_{[j_1],i} \|_2 \notag \\
    \leq & ~ \| f(\A)_{j_1} - f(\B)_{j_1} \|_2 \cdot \| \B_{[j_1],i} \|_2 \notag \\
    \leq & ~ R \cdot \| f(\A)_{j_1} - f(\B)_{j_1} \|_2 \notag \\
    \leq & ~ 4 n^{1.5} R^2 \beta^{-2} \exp(2 R^2) \cdot \| \A_{[j_1],*} - \B_{[j_1],*} \|
\end{align}
where the 1st step is inner product calculation, the 2nd step is because of Cauchy-Schwartz inequality, the 3rd step is due to Lemma~\ref{lem:upper_bound:f}, the 4th step is given by $\max_{j_1 \in [n]} \| b_{[j_1]} \|_2 \leq 1$, the 5th step holds since $\max_{j_1 \in [n]} \| \B_{[j_1],*} \| \leq R$,the 6th step follows by
Lemma~\ref{lem:lipschitz_f:A}.

For the third term ($C_3$) of Eq.~\eqref{eq:f_b_f_A:A}, we have
\begin{align} \label{eq:f_b_f_A_p3:A}
    C_3 = & ~ |\langle f(\A)_{j_1}, b_{[j_1]} \rangle  \cdot \langle f(\B)_{j_1}, \B_{[j_1],i}\rangle - \langle f(\B)_{j_1}, b_{[j_1]} \rangle  \cdot \langle f(\B)_{j_1}, \B_{[j_1],i}\rangle | \notag \\
    = & ~ |\langle f(\A)_{j_1} - f(\B)_{j_1}, b_{[j_1]} \rangle \cdot \langle f(\B)_{j_1}, \B_{[j_1],i} \rangle | \notag \\
    \leq & ~ \| f(\A)_{j_1} - f(\B)_{j_1} \|_2 \cdot \| b_{[j_1]} \|_2 \cdot \| f(\B)_{j_1} \|_2 \cdot \| \B_{[j_1],i} \|_2 \notag \\
    \leq & ~ \| b_{[j_1]} \|_2 \cdot \| f(\A)_{j_1} - f(\B)_{j_1} \|_2 \cdot \| \B_{[j_1],i} \|_2 \notag \\
    \leq & ~ \| f(\A)_{j_1} - f(\B)_{j_1} \|_2 \cdot \| \B_{[j_1],i} \|_2 \notag \\
    \leq & ~ R \cdot \| f(\A)_{j_1} - f(\B)_{j_1} \|_2 \notag \\
    \leq & ~ 4 n^{1.5} R^2 \beta^{-2} \exp(2 R^2) \cdot \| \A_{[j_1],*} - \B_{[j_1],*} \|
\end{align}
where the 1st step is inner product calculation, the 2nd step is because of Cauchy-Schwartz inequality, the 3rd step is due to Lemma~\ref{lem:upper_bound:f}, the 4th step is given by $\max_{j_1 \in [n]} \| b_{[j_1]} \|_2 \leq 1$, the 5th step holds since $\max_{j_1 \in [n]} \| \B_{[j_1],*} \| \leq R$,the 6th step follows by
Lemma~\ref{lem:lipschitz_f:A}.

Putting Eq.~\eqref{eq:f_b_f_A_p1:A}, Eq.~\eqref{eq:f_b_f_A_p2:A}, and Eq.~\eqref{eq:f_b_f_A_p3:A} into Eq.~\eqref{eq:f_b_f_A:A}, we have
\begin{align} \label{eq:f_b_f_A_whole:A}
    & ~ |\langle f(\A)_{j_1}, b_{[j_1]} \rangle  \cdot \langle f(\A)_{j_1}, \A_{[j_1],i}\rangle - \langle f(\B)_{j_1}, b_{[j_1]} \rangle  \cdot \langle f(\B)_{j_1}, \B_{[j_1],i}\rangle | \notag \\
    \leq & ~ C_1 + C_2 + C_3 \notag \\
    \leq & ~ (8 n^{1.5} R^2 \beta^{-2} \exp(2 R^2) + 1) \cdot \| \A_{[j_1],*} - \B_{[j_1],*} \|
\end{align}

For the second term of Eq.~\eqref{eq:grad_L_cent:A}, we have
\begin{align} \label{eq:f_A_b:A}
    & | \langle f(\A)_{j_1} \circ \A_{[j_1],i}, b_{[j_1]} \rangle - \langle f(\B)_{j_1} \circ \B_{[j_1],i}, b_{[j_1]} \rangle| \notag \\
    \leq & ~ | \langle f(\A)_{j_1} \circ \A_{[j_1],i}, b_{[j_1]} \rangle - \langle f(\A)_{j_1} \circ \B_{[j_1],i}, b_{[j_1]} \rangle| ~ + \notag \\
    & ~ | \langle f(\A)_{j_1} \circ \B_{[j_1],i}, b_{[j_1]} \rangle - \langle f(\B)_{j_1} \circ \B_{[j_1],i}, b_{[j_1]} \rangle| \notag \\
    := & ~ C_4 + C_5
\end{align}

For the first item ($C_4$) of Eq.~\eqref{eq:f_A_b:A}, we have,
\begin{align} \label{eq:f_A_b_p1:A}
    C_4 = & ~ | \langle f(\A)_{j_1} \circ \A_{[j_1],i}, b_{[j_1]} \rangle - \langle f(\A)_{j_1} \circ \B_{[j_1],i}, b_{[j_1]} \rangle| \notag \\
    = & ~ | \langle f(\A)_{j_1} \circ \A_{[j_1],i} - f(\A)_{j_1} \circ \B_{[j_1],i}, b_{[j_1]} \rangle|  \notag \\
    = & ~ | \langle f(\A)_{j_1} \circ (\A_{[j_1],i} - \B_{[j_1],i}), b_{[j_1]} \rangle| \notag \\
    \leq & ~ \| f(\A)_{j_1} \circ (\A_{[j_1],i} - \B_{[j_1],i}) \|_2 \cdot \| b_{[j_1]} \|_2 \notag \\
    \leq & ~ \| f(\A)_{j_1} \|_\infty \cdot \| \A_{[j_1],i} - \B_{[j_1],i} \|_2 \cdot \| b_{[j_1]} \|_2 \notag \\
    \leq & ~ \| f(\A)_{j_1} \|_2 \cdot \| \A_{[j_1],i} - \B_{[j_1],i} \|_2 \cdot \| b_{[j_1]} \|_2 \notag \\
    \leq & ~ \| \A_{[j_1],i} - \B_{[j_1],i} \|_2 \cdot \| b_{[j_1]} \|_2 \notag \\
    \leq & ~ \| \A_{[j_1],i} - \B_{[j_1],i} \|_2 \notag \\
    \leq & ~ \| \A_{[j_1],*} - \B_{[j_1],*} \|
\end{align}
where the 1st step is inner product calculation, the 2nd step is Hadamard product calculation, the 3rd step uses Cauchy-Schwartz inequality, the 4th and the 5th steps are due to Fact~\ref{fac:vector_norm}, the 6th step is from Lemma~\ref{lem:upper_bound:f}, the 7th step is because $\max_{j_1 \in [n]} \| b_{[j_1]} \|_2 \leq 1$, the 8th step follows by the definition of matrix norm.

For the second item ($C_5$) of Eq.~\eqref{eq:f_A_b:A}, we have,
\begin{align} \label{eq:f_A_b_p2:A}
    C_5 = & ~ | \langle f(\A)_{j_1} \circ \B_{[j_1],i}, b_{[j_1]} \rangle - \langle f(\B)_{j_1} \circ \B_{[j_1],i}, b_{[j_1]} \rangle| \notag \\
    = & ~ | \langle f(\A)_{j_1} \circ \B_{[j_1],i} - f(\B)_{j_1} \circ \B_{[j_1],i}, b_{[j_1]} \rangle|  \notag \\
    = & ~ | \langle (f(\A)_{j_1} - f(\B)_{j_1}) \circ \B_{[j_1],i}, b_{[j_1]} \rangle| \notag \\
    \leq & ~ \| f(\A)_{j_1} - f(\B)_{j_1}) \circ \B_{[j_1],i} \|_2 \cdot \| b_{[j_1]} \|_2 \notag \\
    \leq & ~ \| f(\A)_{j_1} - f(\B)_{j_1} \|_\infty \cdot \| \B_{[j_1],i} \|_2 \cdot \| b_{[j_1]} \|_2 \notag \\
    \leq & ~ \| f(\A)_{j_1} - f(\B)_{j_1} \|_2 \cdot \| \B_{[j_1],i} \|_2 \cdot \| b_{[j_1]} \|_2 \notag \\
    \leq & ~ R \cdot \| f(\A)_{j_1} - f(\B)_{j_1} \|_2 \cdot \| b_{[j_1]} \|_2 \notag \\
    \leq & ~ R \cdot \| f(\A)_{j_1} - f(\B)_{j_1} \|_2 \notag \\
    \leq & ~ 4 n^{1.5} R^2 \beta^{-2} \exp(2 R^2) \cdot \| \A_{[j_1],*} - \B_{[j_1],*} \|
\end{align}
where the 1st step is inner product calculation, the 2nd step is Hadamard product calculation, the 3rd step uses Cauchy-Schwartz inequality, the 4th and the 5th steps are due to Fact~\ref{fac:vector_norm}, the 6th step is from $\max_{j_1 \in [n]} \| \A_{[j_1],*} - \B_{[j_1],*} \| \leq R$, the 7th step is because $\max_{j_1 \in [n]} \| b_{[j_1]} \|_2 \leq 1$, the 8th step follows by Lemma~\ref{lem:lipschitz_f:A}.

Putting Eq.~\eqref{eq:f_A_b_p1:A} and Eq.~\eqref{eq:f_A_b_p2:A} into Eq.~\eqref{eq:f_A_b:A}, we have
\begin{align} \label{eq:f_A_b_whole:A}
    & | \langle f(\A)_{j_1} \circ \A_{[j_1],i}, b_{[j_1]} \rangle - \langle f(\B)_{j_1} \circ \B_{[j_1],i}, b_{[j_1]} \rangle| \notag \\
    \leq & ~ C_4 + C_5 \notag \\
    \leq & ~ (4 n^{1.5} R^2 \beta^{-2} \exp(2 R^2) + 1) \cdot \| \A_{[j_1],*} - \B_{[j_1],*} \|
\end{align}

Putting Eq.~\eqref{eq:f_b_f_A_whole:A} and Eq.\eqref{eq:f_A_b_whole:A} into Eq.~\eqref{eq:grad_L_cent:A} yields,
\begin{align*}
     & \| \nabla L_{\cent}(\A) - \nabla L_{\cent}(\B) \|_2^2 \notag \\ 
    \leq & ~ \sum_{i=1}^{d^2} \sum_{j_1=1}^n ((12 n^{1.5} R^2 \beta^{-2} \exp(2 R^2) + 2) \cdot \| \A_{[j_1],*} - \B_{[j_1],*} \|) ^2 \notag \\
    \leq & ~ \sum_{i=1}^{d^2} \sum_{j_1=1}^n ((12 n^{1.5} R^2 \exp(4 R^2) + 2) \cdot \| \A_{[j_1],*} - \B_{[j_1],*} \|) ^2 \notag \\
    \leq & ~ d^2 n (n^{1.5} \exp(5 R^2) \cdot \| \A - \B \|_{\infty,2}) ^2
\end{align*}
where the 2nd step is due to Lemma~\ref{lem:lower_bound_A:beta}, the 3rd step is because $R > 4$.

Therefore,
\begin{align*}
    \| \nabla L_{\cent}(\A) - \nabla L_{\cent}(\B) \|_2 \leq n^2 \exp(5 R^2) \cdot \| \A - \B \|_{\infty,2}
\end{align*}

\end{proof}
\end{lemma}

\section{Lipschitz for Entropy Loss Function} \label{sec:lipschitz_L_ent}

In this Section, we discuss Lipschitz conditions for function $L_{\ent}$ and $\nabla L_{\ent}$.

\subsection{Lipschitz for \texorpdfstring{$L_{\ent}(x)$}{} Function}\label{sec:lipschitz_L_ent:x}

\begin{lemma}\label{lem:lipschitz_L_ent:x}
Provided that the subsequent requirement are satisfied
\begin{itemize}
    \item Let $x \in \R^{d^2}, y \in \R^{d^2}$ satisfy $\| x \|_2 \leq R$ and $\| y \|_2 \leq R$
    \item Let $\A \in \R^{n^2 \times d^2}$
    \item Let $L_{\ent}(x)$ be defined as Definition~\ref{def:L_ent}
    \item Let $\max_{j_1 \in [n]} \| \A_{[j_1],*} (y-x) \|_{\infty} < 0.01$
    \item Let $\max_{j_1 \in [n]}\| \A_{[j_1],*} \| \leq R$ 
     \item The greatest lower bound of $\langle u(x)_{j_1}, {\bf 1}_n \rangle$ is denoted as $\beta$
    \item $\| {\bf 1}_n - \frac{f(x)_{j_1}}{f(y)_{j_1}} \|_\infty \leq 0.1$
    \item $R > 4$
\end{itemize}
Then, for all $j_1 \in [n]$ we have
\begin{align*}
    | L_{\ent}(x) - L_{\ent}(y) | \leq n^3 \beta^{-2} \exp(3 R^2) \cdot \| x - y \|_2
\end{align*}
\begin{proof}
\begin{align*}
    | L_{\ent} (x) - L_{\ent}(y) | = & ~ | \sum_{j_1=1}^n \langle f(x)_{j_1}, h(x)_{j_1} \rangle - \sum_{j_1=1}^n \langle f(y)_{j_1}, h(y)_{j_1} \rangle | \\
    \leq & ~ \sum_{j_1=1}^n (| \langle f(x)_{j_1}, h(x)_{j_1} \rangle -  \langle f(x)_{j_1}, h(y)_{j_1} \rangle | ~ + \\
    & ~ | \langle f(x)_{j_1}, h(y)_{j_1} \rangle - \langle f(y)_{j_1}, h(y)_{j_1} \rangle |) \\
    = & ~ \sum_{j_1=1}^n (| \langle f(x)_{j_1}, h(x)_{j_1} - h(y)_{j_1} \rangle | + | \langle f(x)_{j_1} - f(y)_{j_1}, h(y)_{j_1} \rangle |) \\
    \leq & ~ \sum_{j_1=1}^n (\| f(x)_{j_1} \|_2 \cdot \| h(x)_{j_1} - h(y)_{j_1} \|_2 + \| f(x)_{j_1} - f(y)_{j_1} \|_2 \cdot \| h(y)_{j_1} \|_2) \\
    \leq & ~ \sum_{j_1=1}^n (\| h(x)_{j_1} - h(y)_{j_1} \|_2 + \| f(x)_{j_1} - f(y)_{j_1} \|_2 \cdot \| h(y)_{j_1} \|_2) \\
    \leq & ~ \sum_{j_1=1}^n (\| f(x)_{j_1} - f(y)_{j_1} \|_2 + \| f(x)_{j_1} - f(y)_{j_1} \|_2 \cdot \| h(y)_{j_1} \|_2) \\
    \leq & ~ \sum_{j_1=1}^n (1 + 2 \sqrt{n} R^2) \cdot \| f(x)_{j_1} - f(y)_{j_1} \|_2 \\
    \leq & ~ \sum_{j_1=1}^n (1 + 2 \sqrt{n} R^2) \cdot 4 n^{1.5} R \beta^{-2} \exp(2 R^2) \cdot \| x - y \|_2 \\
    \leq & ~ n^3 \beta^{-2} \exp(3 R^2) \cdot \| x - y \|_2
\end{align*}
where the 1st step is due to the definition of$L_{\ent}$ (see Definition~\ref{def:L_ent}), the 2nd step uses triangle inequality, the 3rd step is inner product calculation, the 4th step follows from Cauchy-Schwartz inequality, the 5th step is from Lemma~\ref{lem:upper_bound:f}, the 6th step is from Lemma~\ref{lem:lipschitz_h:x}, the 7th step follows by Lemma~\ref{lem:upper_bound_h:x}, the 8th step follows by Lemma~\ref{lem:lipschitz_f:x}, the last step holds because $R > 4$.
\end{proof}
\end{lemma}

\subsection{Lipschitz for \texorpdfstring{$L_{\ent}(\A)$}{} Function}\label{sec:lipschitz_L_ent:A}

\begin{lemma}\label{lem:lipschitz_L_ent:A}
Provided that the subsequent requirement are satisfied
\begin{itemize}
    \item Let $\A, \B \in \R^{n^2 \times d^2}$ satisfy $\max_{j_1 \in [n]}\| \A_{[j_1],*} \| \leq R$, $\max_{j_1 \in [n]} \| \B_{[j_1],*} \| \leq R$ \item Let $\max_{j_1 \in [n]} \| ( \A_{[j_1],*} - \B_{[j_1],*} ) x \|_{\infty} < 0.01$
    \item Let $x \in \R^{d^2}$ satisfy that $\| x \|_2 \leq R $
    \item Let $L_{\ent}(\A)$ be defined as Definition~\ref{def:L_ent}
    \item The greatest lower bound of $\langle u(\A)_{j_1}, {\bf 1}_n \rangle$ is denoted as $\beta$
    \item Let $\| \A - \B \|_{\infty,2} = \max_{j_1 \in [n]}$
    \item $\| {\bf 1}_n - \frac{f(\A)_{j_1}}{f(\B)_{j_1}} \|_\infty \leq 0.1$
    \item $R > 4$
\end{itemize}
Then, for all $j_1 \in [n]$ we have
\begin{align*}
    | L_{\ent}(\A) - L_{\ent}(\B) | \leq n^3 \beta^{-2} \exp(3 R^2) \cdot \| \A - \B \|_{\infty,2}
\end{align*}
\begin{proof}
\begin{align*}
    | L_{\ent} (\A) - L_{\ent}(\B) | = & ~ | \sum_{j_1=1}^n \langle f(\A)_{j_1}, h(\A)_{j_1} \rangle - \sum_{j_1=1}^n \langle f(\B)_{j_1}, h(\B)_{j_1} \rangle | \\
    \leq & ~ \sum_{j_1=1}^n (| \langle f(\A)_{j_1}, h(\A)_{j_1} \rangle -  \langle f(\A)_{j_1}, h(\B)_{j_1} \rangle | ~ + \\
    & ~ | \langle f(\A)_{j_1}, h(\B)_{j_1} \rangle - \langle f(\B)_{j_1}, h(\B)_{j_1} \rangle |) \\
    = & ~ \sum_{j_1=1}^n (| \langle f(\A)_{j_1}, h(\A)_{j_1} - h(\B)_{j_1} \rangle | + | \langle f(\A)_{j_1} - f(\B)_{j_1}, h(\B)_{j_1} \rangle |) \\
    \leq & ~ \sum_{j_1=1}^n (\| f(\A)_{j_1} \|_2 \cdot \| h(\A)_{j_1} - h(\B)_{j_1} \|_2 + \| f(\A)_{j_1} - f(\B)_{j_1} \|_2 \cdot \| h(\B)_{j_1} \|_2) \\
    \leq & ~ \sum_{j_1=1}^n (\| h(\A)_{j_1} - h(\B)_{j_1} \|_2 + \| f(\A)_{j_1} - f(\B)_{j_1} \|_2 \cdot \| h(\B)_{j_1} \|_2) \\
    \leq & ~ \sum_{j_1=1}^n (\| f(\A)_{j_1} - f(\B)_{j_1} \|_2 + \| f(\A)_{j_1} - f(\B)_{j_1} \|_2 \cdot \| h(\B)_{j_1} \|_2) \\
    \leq & ~ \sum_{j_1=1}^n (1 + 2 \sqrt{n} R^2) \cdot \| f(\A)_{j_1} - f(\B)_{j_1} \|_2 \\
    \leq & ~ \sum_{j_1=1}^n (1 + 2 \sqrt{n} R^2) \cdot 4 n^{1.5} R \beta^{-2} \exp(2 R^2) \cdot \| \A_{[j_1],*} - \B_{[j_1],*} \| \\
    \leq & ~ n^3 \beta^{-2} \exp(3 R^2) \cdot \| \A - \B \|_{\infty,2}
\end{align*}
where the 1st step is due to the definition of$L_{\ent}$ (see Definition~\ref{def:L_ent}), the 2nd step uses triangle inequality, the 3rd step is inner product calculation, the 4th step follows from Cauchy-Schwartz inequality, the 5th step is from Lemma~\ref{lem:upper_bound:f}, the 6th step is from Lemma~\ref{lem:lipschitz_h:A}, the 7th step follows by Lemma~\ref{lem:upper_bound_h:x}, the 8th step is due to Lemma~\ref{lem:lipschitz_f:A}, the last step holds because $R > 4$.
\end{proof}
\end{lemma}

\subsection{Lipschitz for \texorpdfstring{$\nabla L_{\ent}(x)$}{} Function}\label{sec:lipschitz_grad_L_ent:x}

\begin{lemma}\label{lem:lipschitz_grad_L_ent:x}
Provided that the subsequent requirement are satisfied
\begin{itemize}
    \item Let $x \in \R^{d^2}, y \in \R^{d^2}$ satisfy $\| x \|_2 \leq R$ and $\| y \|_2 \leq R$
    \item Let $\A \in \R^{n^2 \times d^2}$
    \item Let $L_{\ent}(x)$ be defined as Definition~\ref{def:L_ent}
    \item Let $\max_{j_1 \in [n]} \| \A_{[j_1],*} (y-x) \|_{\infty} < 0.01$
    \item Let $\max_{j_1 \in [n]}\| \A_{[j_1],*} \| \leq R$ 
    \item $\| {\bf 1}_n - \frac{f(x)_{j_1}}{f(y)_{j_1}} \|_\infty \leq 0.1$
     \item The greatest lower bound of $\langle u(\A)_{j_1}, {\bf 1}_n \rangle$ is denoted as $\beta$
     \item $R > 4$
\end{itemize}
Then, for all $j_1 \in [n]$ we have
\begin{align*}
     \| \nabla L_{\ent}(x) - \nabla L_{\ent}(y) \|_2 \leq dn^{2.5} \exp(5 R^2) \cdot \| x - y \|_2
\end{align*}
\begin{proof}
Utilizing triangle inequality, we have
\begin{align} \label{eq:grad_L_ent}
    & \| \nabla L_{\ent}(x) - \nabla L_{\ent}(y) \|_2^2 \notag \\
    = & ~ \sum_{i=1}^{d^2} |\sum_{j_1=1}^n (\langle f(x)_{j_1}  , h(x)_{j_1} \rangle \cdot \langle f(x)_{j_1}, \A_{[j_1],i}\rangle - \langle f(x)_{j_1} \circ \A_{[j_1],i}, h(x)_{j_1} \rangle ~ + \notag \\
    & ~ \langle f(x)_{j_1}, \A_{[j_1],i}\rangle \cdot \langle f(x)_{j_1}, {\bf 1}_n \rangle - \langle f(x)_{j_1}, \A_{[j_1],i} \rangle) \notag ~ - \\
    & ~ \sum_{j_1=1}^n (\langle f(y)_{j_1}  , h(y)_{j_1} \rangle \cdot \langle f(y)_{j_1}, \A_{[j_1],i}\rangle - \langle f(y)_{j_1} \circ \A_{[j_1],i}, h(y)_{j_1} \rangle ~ + \notag \\
    & ~ \langle f(y)_{j_1}, \A_{[j_1],i}\rangle \cdot \langle f(y)_{j_1}, {\bf 1}_n \rangle - \langle f(y)_{j_1}, \A_{[j_1],i} \rangle) \notag |^2\\
    \leq & ~ \sum_{i=1}^{d^2} \sum_{j_1=1}^n ( |\langle f(x)_{j_1}  , h(x)_{j_1} \rangle \cdot \langle f(x)_{j_1}, \A_{[j_1],i}\rangle -\langle f(y)_{j_1}  , h(y)_{j_1} \rangle \cdot \langle f(y)_{j_1}, \A_{[j_1],i}\rangle | ~ + \notag \\
    & ~ | \langle f(x)_{j_1} \circ \A_{[j_1],i}, h(x)_{j_1} \rangle - \langle f(y)_{j_1} \circ \A_{[j_1],i}, h(y)_{j_1} \rangle| \notag ~ + \\
    & ~ |\langle f(x)_{j_1}, \A_{[j_1],i}\rangle \cdot \langle f(x)_{j_1}, {\bf 1}_n \rangle - \langle f(y)_{j_1}, \A_{[j_1],i}\rangle \cdot \langle f(y)_{j_1}, {\bf 1}_n \rangle | ~ + \notag \\
    & ~ |\langle f(x)_{j_1}, \A_{[j_1],i} \rangle - \langle f(y)_{j_1}, \A_{[j_1],i} \rangle| )^2
\end{align}

For the first item of Eq.~\eqref{eq:grad_L_ent}, by triangle inequality, we have
\begin{align} \label{eq:f_h_f_A}
    & |\langle f(x)_{j_1}  , h(x)_{j_1} \rangle \cdot \langle f(x)_{j_1}, \A_{[j_1],i}\rangle -\langle f(y)_{j_1}  , h(y)_{j_1} \rangle \cdot \langle f(y)_{j_1}, \A_{[j_1],i}\rangle | \notag \\
    \leq & ~ |\langle f(x)_{j_1}  , h(x)_{j_1} \rangle \cdot \langle f(x)_{j_1}, \A_{[j_1],i}\rangle -\langle f(x)_{j_1}  , h(x)_{j_1} \rangle \cdot \langle f(y)_{j_1}, \A_{[j_1],i}\rangle | ~ + \notag \\
    & ~ |\langle f(x)_{j_1}  , h(x)_{j_1} \rangle \cdot \langle f(y)_{j_1}, \A_{[j_1],i}\rangle -\langle f(x)_{j_1}  , h(y)_{j_1} \rangle \cdot \langle f(y)_{j_1}, \A_{[j_1],i}\rangle | ~ + \notag \\
    & ~ |\langle f(x)_{j_1}  , h(y)_{j_1} \rangle \cdot \langle f(y)_{j_1}, \A_{[j_1],i}\rangle -\langle f(y)_{j_1}  , h(y)_{j_1} \rangle \cdot \langle f(y)_{j_1}, \A_{[j_1],i}\rangle | \notag \\
    := & ~ C_1 + C_2 + C_3
\end{align}

For the first item ($C_1$) of Eq.~\eqref{eq:f_h_f_A}, we have
\begin{align} \label{eq:f_h_f_A_p1}
    C_1 = & ~ |\langle f(x)_{j_1}  , h(x)_{j_1} \rangle \cdot \langle f(x)_{j_1}, \A_{[j_1],i}\rangle -\langle f(x)_{j_1}  , h(x)_{j_1} \rangle \cdot \langle f(y)_{j_1}, \A_{[j_1],i}\rangle | \notag \\
    = & ~ |\langle f(x)_{j_1}  , h(x)_{j_1} \rangle \cdot \langle f(x)_{j_1} - f(y)_{j_1}, \A_{[j_1],i}\rangle | \notag \\
    \leq & ~ \| f(x)_{j_1} \|_2 \cdot \| h(x)_{j_1} \|_2 \cdot \| f(x)_{j_1} - f(y)_{j_1} \|_2 \cdot \| \A_{[j_1],i} \|_2 \notag \\
    \leq & ~ \| h(x)_{j_1} \|_2 \cdot \| f(x)_{j_1} - f(y)_{j_1} \|_2 \cdot \| \A_{[j_1],i} \|_2 \notag \\
    \leq & ~ 2 \sqrt{n} R^2 \cdot \| f(x)_{j_1} - f(y)_{j_1} \|_2 \cdot \| \A_{[j_1],i} \|_2 \notag \\
    \leq & ~ 2 \sqrt{n} R^3 \cdot \| f(x)_{j_1} - f(y)_{j_1} \|_2 \notag \\
    \leq & ~ 8 n^2 R^4 \beta^{-2} \exp(2 R^2) \cdot \| x - y \|_2
\end{align}
where the 1st step is inner product calculation, the 2nd step is derived from Cauchy-Schwartz inequality, the 3rd step follows by Lemma~\ref{lem:upper_bound:f}, the 4th step follows by Lemma~\ref{lem:upper_bound_h:x}, the 5th step is due to $\max{j_1 \in [n]} \| \A_{[j_1],*} \| \leq R$, the last step follows from Lemma~\ref{lem:lipschitz_f:x}.

For the second item ($C_2$) of Eq.~\eqref{eq:f_h_f_A}, we have
\begin{align} \label{eq:f_h_f_A_p2}
    C_2 = & ~ |\langle f(x)_{j_1}  , h(x)_{j_1} \rangle \cdot \langle f(y)_{j_1}, \A_{[j_1],i}\rangle -\langle f(x)_{j_1}  , h(y)_{j_1} \rangle \cdot \langle f(y)_{j_1}, \A_{[j_1],i}\rangle | \notag \\
    = & ~ |\langle f(x)_{j_1}  , h(x)_{j_1} - h(y)_{j_1} \rangle \cdot \langle f(y)_{j_1}, \A_{[j_1],i}\rangle | \notag \\
    \leq & ~ \| f(x)_{j_1} \|_2 \cdot \| h(x)_{j_1} - h(y)_{j_1}  \|_2 \cdot \| f(y)_{j_1} \|_2 \cdot \| \A_{[j_1],i} \|_2 \notag \\
    \leq & ~ \| h(x)_{j_1} - h(y)_{j_1}\|_2 \cdot \| \A_{[j_1],i} \|_2 \notag \\
    \leq & ~ R \cdot \| h(x)_{j_1} - h(y)_{j_1} \|_2  \notag \\
    \leq & ~ R \cdot \| f(x)_{j_1} - f(y)_{j_1} \|_2 \notag \\
    \leq & ~ 4 n^{1.5} R^2 \beta^{-2} \exp(2 R^2) \cdot \| x - y \|_2
\end{align}
where the 1st step is inner product calculation, the 2nd step is derived from Cauchy-Schwartz inequality, the 3rd step follows by Lemma~\ref{lem:upper_bound:f}, the 4th step follows by $\max{j_1 \in [n]} \| \A_{[j_1],*} \| \leq R$, the 5th step is due to Lemma~\ref{lem:lipschitz_h:x}, the last step follows from Lemma~\ref{lem:lipschitz_f:x}.

For the first item ($C_3$) of Eq.~\eqref{eq:f_h_f_A}, we have
\begin{align} \label{eq:f_h_f_A_p3}
    C_3 = & ~ |\langle f(x)_{j_1}  , h(y)_{j_1} \rangle \cdot \langle f(y)_{j_1}, \A_{[j_1],i}\rangle -\langle f(y)_{j_1}  , h(y)_{j_1} \rangle \cdot \langle f(y)_{j_1}, \A_{[j_1],i}\rangle | \notag \\
    = & ~ |\langle f(x)_{j_1} - f(y)_{j_1}  , h(x)_{j_1} \rangle \cdot \langle f(y)_{j_1}, \A_{[j_1],i}\rangle | \notag \\
    \leq & ~ \| f(x)_{j_1} - f(y)_{j_1} \|_2 \cdot \| h(x)_{j_1} \|_2 \cdot \| f(y)_{j_1} \|_2 \cdot \| \A_{[j_1],i} \|_2 \notag \\
    \leq & ~ \| h(x)_{j_1} \|_2 \cdot \| f(x)_{j_1} - f(y)_{j_1} \|_2 \cdot \| \A_{[j_1],i} \|_2 \notag \\
    \leq & ~ 2 \sqrt{n} R^2 \cdot \| f(x)_{j_1} - f(y)_{j_1} \|_2 \cdot \| \A_{[j_1],i} \|_2 \notag \\
    \leq & ~ 2 \sqrt{n} R^3 \cdot \| f(x)_{j_1} - f(y)_{j_1} \|_2 \notag \\
    \leq & ~ 8 n^2 R^4 \beta^{-2} \exp(2 R^2) \cdot \| x - y \|_2
\end{align}
where the 1st step is inner product calculation, the 2nd step is derived from Cauchy-Schwartz inequality, the 3rd step follows by Lemma~\ref{lem:upper_bound:f}, the 4th step follows by Lemma~\ref{lem:upper_bound_h:x}, the 5th step is due to $\max{j_1 \in [n]} \| \A_{[j_1],*} \| \leq R$, the last step follows from Lemma~\ref{lem:lipschitz_f:x}.

Putting Eq.~\eqref{eq:f_h_f_A_p1}, Eq.~\eqref{eq:f_h_f_A_p2}, and Eq.~\eqref{eq:f_h_f_A_p3} into Eq.~\eqref{eq:f_h_f_A}, we have
\begin{align} \label{eq:f_h_f_A_whole}
    & |\langle f(x)_{j_1}  , h(x)_{j_1} \rangle \cdot \langle f(x)_{j_1}, \A_{[j_1],i}\rangle -\langle f(y)_{j_1}  , h(y)_{j_1} \rangle \cdot \langle f(y)_{j_1}, \A_{[j_1],i}\rangle | \notag \\
    \leq & ~ C_1 + C_2 + C_3 \notag \\
    \leq & ~ (4 n^{0.5} R^2 +1) \cdot 4n^{1.5} R^2 \beta^{-2} \exp(2 R^2) \cdot \| x - y \|_2
\end{align}

For the second term of Eq.~\eqref{eq:grad_L_ent}, we have
\begin{align} \label{eq:f_A_h}
    & | \langle f(x)_{j_1} \circ \A_{[j_1],i}, h(x)_{j_1} \rangle - \langle f(y)_{j_1} \circ \A_{[j_1],i}, h(y)_{j_1} \rangle| \notag \\
    \leq & ~ | \langle f(x)_{j_1} \circ \A_{[j_1],i}, h(x)_{j_1} \rangle - \langle f(x)_{j_1} \circ \A_{[j_1],i}, h(y)_{j_1} \rangle| ~ + \notag \\
    & ~ | \langle f(x)_{j_1} \circ \A_{[j_1],i}, h(y)_{j_1} \rangle - \langle f(y)_{j_1} \circ \A_{[j_1],i}, h(y)_{j_1} \rangle| \notag \\
    & ~ := C_4 + C_5
\end{align}

For the first term ($C_4$) of Eq.~\eqref{eq:f_A_h}, we have
\begin{align} \label{eq:f_A_h_p1}
    C_4 = & ~ | \langle f(x)_{j_1} \circ \A_{[j_1],i}, h(x)_{j_1} \rangle - \langle f(x)_{j_1} \circ \A_{[j_1],i}, h(y)_{j_1} \rangle| \notag \\
    = & ~ | \langle f(x)_{j_1} \circ \A_{[j_1],i}, h(x)_{j_1} - h(y)_{j_1} \rangle| \notag \\
    \leq & ~ \| f(x)_{j_1} \circ \A_{[j_1],i} \|_2 \cdot \| h(x)_{j_1} - h(y)_{j_1} \|_2 \notag \\
    \leq & ~ \| f(x)_{j_1} \|_\infty \cdot \| \A_{[j_1],i} \|_2 \cdot \| h(x)_{j_1} - h(y)_{j_1} \|_2 \notag \\
    \leq & ~ \| f(x)_{j_1} \|_2 \cdot \| \A_{[j_1],i} \|_2 \cdot \| h(x)_{j_1} - h(y)_{j_1} \|_2 \notag \\
    \leq & ~ \| \A_{[j_1],i} \|_2 \cdot \| h(x)_{j_1} - h(y)_{j_1} \|_2 \notag \\
    \leq & ~ R \cdot \| h(x)_{j_1} - h(y)_{j_1} \|_2 \notag \\
    \leq & ~ R \cdot \| f(x)_{j_1} - f(y)_{j_1} \|_2 \notag \\
    \leq & ~ 4 n^{1.5} R^2 \beta^{-2} \exp(2 R^2) \cdot \| x - y \|_2
\end{align}
where the 1st step is inner product calculation, the 2nd is due to Cauchy-Schwartz inequality, the 3rd step and the 4th step are from Fact.~\ref{fac:vector_norm}, the 5th step is because of Lemma~\ref{lem:upper_bound:f}, the 6th step is because $\max_{j_1 \in [n]} \| \A_{[j_1],*} \| \leq R$, the 7th step follows by Lemma~\ref{lem:lipschitz_h:x}, the 8th step is given by Lemma~\ref{lem:lipschitz_f:x}.

For the second term ($C_5$) of Eq.~\eqref{eq:f_A_h}, we have
\begin{align} \label{eq:f_A_h_p2}
    C_5 = & ~ | \langle f(x)_{j_1} \circ \A_{[j_1],i}, h(y)_{j_1} \rangle - \langle f(y)_{j_1} \circ \A_{[j_1],i}, h(y)_{j_1} \rangle| \notag \\
    = & ~ | \langle f(x)_{j_1} \circ \A_{[j_1],i} -f(y)_{j_1} \circ \A_{[j_1],i}, h(y)_{j_1} \rangle| \notag \\
    = & ~ | \langle (f(x)_{j_1} - f(y)_{j_1}) \circ \A_{[j_1],i}, h(y)_{j_1} \rangle| \notag \\
    \leq & ~ \| (f(x)_{j_1} - f(y)_{j_1}) \circ \A_{[j_1],i} \|_2 \cdot \| h(y)_{j_1} \|_2 \notag \\
    \leq & ~ \| f(x)_{j_1} - f(y)_{j_1} \|_\infty \cdot \| \A_{[j_1],i} \|_2 \cdot \| h(y)_{j_1} \|_2 \notag \\
    \leq & ~ \| f(x)_{j_1} - f(y)_{j_1} \|_2 \cdot \| \A_{[j_1],i} \|_2 \cdot \| h(y)_{j_1} \|_2 \notag \\
    \leq & ~ R \cdot \| f(x)_{j_1} - f(y)_{j_1} \|_2  \cdot \| h(y)_{j_1} \|_2 \notag \\
    \leq & ~ 2 \sqrt{n} R^3 \cdot \| f(x)_{j_1} - f(y)_{j_1} \|_2 \notag \\
    \leq & ~ 8 n^2 R^4 \beta^{-2} \exp(2 R^2) \cdot \| x - y \|_2
\end{align}
where the 1st step is inner product calculation, the 2nd is Hadamard product calculation, the 3rd step is due to Cauchy-Schwartz inequality, the 4th step and the 5th step are from Fact.~\ref{fac:vector_norm}, the 6th step is because $\max_{j_1 \in [n]} \| \A_{[j_1],*} \| \leq R$, the 7th step follows by Lemma~\ref{lem:upper_bound_h:x}, the 8th step is given by Lemma~\ref{lem:lipschitz_f:x}.

Putting Eq.~\eqref{eq:f_A_h_p1} and Eq.~\eqref{eq:f_A_h_p2} into Eq.~\eqref{eq:f_A_h}, we have
\begin{align} \label{eq:f_A_h_whole}
    & |\langle f(x)_{j_1}  , h(x)_{j_1} \rangle \cdot \langle f(x)_{j_1}, \A_{[j_1],i}\rangle -\langle f(y)_{j_1}  , h(y)_{j_1} \rangle \cdot \langle f(y)_{j_1}, \A_{[j_1],i}\rangle | \notag \\
    \leq & ~ C_4 + C_5 \notag \\
    \leq & ~ (2 n^{0.5} R^2 +1) \cdot 4n^{1.5} R^2 \beta^{-2} \exp(2 R^2) \cdot \| x - y \|_2
\end{align}

For the third term of Eq.~\eqref{eq:grad_L_ent}, we have
\begin{align} \label{eq:f_A_f_1}
    & |\langle f(x)_{j_1}, \A_{[j_1],i}\rangle \cdot \langle f(x)_{j_1}, {\bf 1}_n \rangle - \langle f(y)_{j_1}, \A_{[j_1],i}\rangle \cdot \langle f(y)_{j_1}, {\bf 1}_n \rangle | \notag \\
    \leq & ~ |\langle f(x)_{j_1}, \A_{[j_1],i}\rangle \cdot \langle f(x)_{j_1}, {\bf 1}_n \rangle - \langle f(x)_{j_1}, \A_{[j_1],i}\rangle \cdot \langle f(y)_{j_1}, {\bf 1}_n \rangle | ~ + \notag \\
    & ~ |\langle f(x)_{j_1}, \A_{[j_1],i}\rangle \cdot \langle f(y)_{j_1}, {\bf 1}_n \rangle - \langle f(y)_{j_1}, \A_{[j_1],i}\rangle \cdot \langle f(y)_{j_1}, {\bf 1}_n \rangle | \notag \\
    := & ~ C_6 + C_7
\end{align}

For the first item ($C_6$) of Eq.~\eqref{eq:f_A_f_1}, we have
\begin{align} \label{eq:f_A_f_1_p1}
    C_6 = & ~ |\langle f(x)_{j_1}, \A_{[j_1],i}\rangle \cdot \langle f(x)_{j_1}, {\bf 1}_n \rangle - \langle f(x)_{j_1}, \A_{[j_1],i}\rangle \cdot \langle f(y)_{j_1}, {\bf 1}_n \rangle | \notag \\
    \leq & ~ |\langle f(x)_{j_1}, \A_{[j_1],i}\rangle \cdot \langle f(x)_{j_1} -  f(y)_{j_1}, {\bf 1}_n \rangle | \notag \\
    \leq & ~ \| f(x)_{j_1} \|_2 \cdot \| \A_{[j_1],i} \|_2 \cdot \| f(x)_{j_1} -  f(y)_{j_1} \|_2 \cdot \| {\bf 1}_n \|_2 \notag \\
    \leq & ~ \| \A_{[j_1],i} \|_2 \cdot \| f(x)_{j_1} -  f(y)_{j_1} \|_2 \cdot \| {\bf 1}_n \| \notag \\
    \leq & ~ R \cdot \| f(x)_{j_1} -  f(y)_{j_1} \|_2 \cdot \| {\bf 1}_n \|_2 \notag \\
    \leq & ~ 4n^{1.5} R^2 \beta{-2} \exp(2 R^2) \cdot \| x - y \|_2 \cdot \| {\bf 1}_n \|_2 \notag \\
    \leq & ~ 4n^2 R^2 \beta{-2} \exp(2 R^2) \cdot \| x - y \|_2
\end{align}
where the 1st step is inner product calculation, the 2nd step is due to Cauchy-Schwartz inequality, the 3rd step is derived from Lemma~\ref{lem:upper_bound:f}, the 4th step holds because $\max_{j_1 \in [n]} \| \A_{[j_1],*} \| \leq R$, the 5th step uses Lemma~\ref{lem:lipschitz_f:x}.

For the second item ($C_7$) of Eq.~\eqref{eq:f_A_f_1}, we have
\begin{align} \label{eq:f_A_f_1_p2}
    C_7 = & ~ |\langle f(x)_{j_1}, \A_{[j_1],i}\rangle \cdot \langle f(y)_{j_1}, {\bf 1}_n \rangle - \langle f(y)_{j_1}, \A_{[j_1],i}\rangle \cdot \langle f(y)_{j_1}, {\bf 1}_n \rangle | \notag \\
    \leq & ~ |\langle f(x)_{j_1} - f(y)_{j_1}, \A_{[j_1],i}\rangle \cdot \langle f(y)_{j_1} , {\bf 1}_n \rangle | \notag \\
    \leq & ~ \| f(x)_{j_1} -  f(y)_{j_1}  \|_2 \cdot \| \A_{[j_1],i} \|_2 \cdot \| f(y)_{j_1}\|_2 \cdot \| {\bf 1}_n \|_2 \notag \\
    \leq & ~ \| \A_{[j_1],i} \|_2 \cdot \| f(x)_{j_1} -  f(y)_{j_1} \|_2 \cdot \| {\bf 1}_n \| \notag \\
    \leq & ~ R \cdot \| f(x)_{j_1} -  f(y)_{j_1} \|_2 \cdot \| {\bf 1}_n \|_2 \notag \\
    \leq & ~ 4n^{1.5} R^2 \beta{-2} \exp(2 R^2) \cdot \| x - y \|_2 \cdot \| {\bf 1}_n \|_2 \notag \\
    \leq & ~ 4n^2 R^2 \beta{-2} \exp(2 R^2) \cdot \| x - y \|_2
\end{align}
where the 1st step is inner product calculation, the 2nd step is due to Cauchy-Schwartz inequality, the 3rd step is derived from Lemma~\ref{lem:upper_bound:f}, the 4th step holds because $\max_{j_1 \in [n]} \| \A_{[j_1],*} \| \leq R$, the 5th step uses Lemma~\ref{lem:lipschitz_f:x}.

Putting Eq.~\eqref{eq:f_A_f_1_p1} and Eq.~\eqref{eq:f_A_f_1_p2} into Eq.~\eqref{eq:f_A_f_1}, we have
\begin{align} \label{eq:f_A_f_1_whole}
    & |\langle f(x)_{j_1}, \A_{[j_1],i}\rangle \cdot \langle f(x)_{j_1}, {\bf 1}_n \rangle - \langle f(y)_{j_1}, \A_{[j_1],i}\rangle \cdot \langle f(y)_{j_1}, {\bf 1}_n \rangle | \notag \\
    \leq & ~ C_6 + C_7 \notag \\
    \leq & ~ 8n^2 R^2 \beta{-2} \exp(2 R^2) \cdot \| x - y \|_2
\end{align}

For the fourth item of Eq.~\eqref{eq:grad_L_ent}, we have
\begin{align} \label{eq:f_A}
    |\langle f(x)_{j_1}, \A_{[j_1],i} \rangle - \langle f(y)_{j_1}, \A_{[j_1],i} \rangle| \notag \leq & ~ |\langle f(x)_{j_1} - f(y)_{j_1}, \A_{[j_1],i} \rangle| \notag \\
    \leq & ~ \| f(x)_{j_1} - f(y)_{j_1} \|_2 \cdot \| \A_{[j_1],i} \|_2 \notag \\
    \leq & ~ R \cdot \| f(x)_{j_1} - f(y)_{j_1} \|_2  \notag \\
    \leq & ~ 4n^2 R^2 \beta{-2} \exp(2 R^2) \cdot \| x - y \|_2
\end{align}
where the 1st step is inner product calculation, the 2nd step is due to Cauchy-Schwartz inequality, the 3th step holds because $\max_{j_1 \in [n]} \| \A_{[j_1],*} \| \leq R$, the 4th step uses Lemma~\ref{lem:lipschitz_f:x}.

Putting Eq.~\eqref{eq:f_h_f_A_whole}, Eq.~\eqref{eq:f_A_h_whole} , Eq.~\eqref{eq:f_A_f_1_whole} and Eq.~\eqref{eq:f_A} into Eq.~\eqref{eq:grad_L_ent}, we have
\begin{align*}
    & \| \nabla L_{\ent}(x) - \nabla L_{\ent}(y) \|_2^2 \notag \\
    \leq & ~ \sum_{i=1}^{d^2} \sum_{j_1=1}^n ((6 n^{0.5} R^2 +5) \cdot 4n^{1.5} R^2 \beta^{-2} \exp(2 R^2) \cdot \| x - y \|_2)^2 \\
    \leq & ~ \sum_{i=1}^{d^2} \sum_{j_1=1}^n ((6 n^{0.5} R^2 +5) \cdot 4n^{1.5} R^2 \exp(4 R^2) \cdot \| x - y \|_2)^2 \\
    \leq & ~ d^2n \cdot ( n^2 \exp(5 R^2) \cdot \| x - y \|_2)^2
\end{align*}

Therefore, we have
\begin{align*}
     \| \nabla L_{\ent}(x) - \nabla L_{\ent}(y) \|_2 \leq dn^{2.5} \exp(5 R^2) \cdot \| x - y \|_2
\end{align*}

\end{proof}
\end{lemma}

\subsection{Lipschitz for \texorpdfstring{$\nabla L_{\ent}(\A)$}{} Function}\label{sec:lipschitz_grad_L_ent:A}

\begin{lemma}\label{lem:lipschitz_grad_L_ent:A}
Provided that the subsequent requirement are satisfied
\begin{itemize}
    \item Let $\A, \B \in \R^{n^2 \times d^2}$ satisfy $\max_{j_1 \in [n]}\| \A_{[j_1],*} \| \leq R$, $\max_{j_1 \in [n]} \| \B_{[j_1],*} \| \leq R$ \item Let $\max_{j_1 \in [n]} \| ( \A_{[j_1],*} - \B_{[j_1],*} ) x \|_{\infty} < 0.01$
    \item Let $x \in \R^{d^2}$ satisfy that $\| x \|_2 \leq R $
    \item Let $L_{\ent}(\A)$ be defined as Definition~\ref{def:L_ent}
    \item The greatest lower bound of $\langle u(\A)_{j_1}, {\bf 1}_n \rangle$ is denoted as $\beta$
    \item Let $\| \A - \B \|_{\infty,2} = \max_{j_1 \in [n]}$
    \item $\| {\bf 1}_n - \frac{f(\A)_{j_1}}{f(\B)_{j_1}} \|_\infty \leq 0.1$
    \item $R > 4$
    \item Let $\| \A - \B \|_{\infty,2} = \max_{j_1 \in [n]} \| \A_{[j_1],*} - \B_{[j_1],*} \|$
\end{itemize}
Then, for all $j_1 \in [n]$ we have
\begin{align*}
    \| \nabla L_{\ent}(\A) - \nabla L_{\ent}(\B) \|_2 \leq dn^{2.5} \exp(5 R^2) \cdot \| \A - \B \|_{\infty,2}
\end{align*}

\begin{proof}
Applying triangle inequality, we have
\begin{align} \label{eq:grad_L_ent:A}
    & \| \nabla L_{\ent}(\A) - \nabla L_{\ent}(\B) \|_2^2 \notag \\
    = & ~ \sum_{i=1}^{d^2} |\sum_{j_1=1}^n (\langle f(\A)_{j_1}  , h(\A)_{j_1} \rangle \cdot \langle f(\A)_{j_1}, \A_{[j_1],i}\rangle - \langle f(\A)_{j_1} \circ \A_{[j_1],i}, h(\A)_{j_1} \rangle ~ + \notag \\
    & ~ \langle f(\A)_{j_1}, \A_{[j_1],i}\rangle \cdot \langle f(\A)_{j_1}, {\bf 1}_n \rangle - \langle f(\A)_{j_1}, \A_{[j_1],i} \rangle) \notag ~ - \\
    & ~ \sum_{j_1=1}^n (\langle f(\B)_{j_1}  , h(\B)_{j_1} \rangle \cdot \langle f(\B)_{j_1}, \B_{[j_1],i}\rangle - \langle f(\B)_{j_1} \B_{[j_1],i}, h(\B)_{j_1} \rangle ~ + \notag \\
    & ~ \langle f(\B)_{j_1}, \B_{[j_1],i}\rangle \cdot \langle f(\B)_{j_1}, {\bf 1}_n \rangle - \langle f(\B)_{j_1}, \B_{[j_1],i} \rangle) \notag |^2 \\
    \leq & ~ \sum_{i=1}^{d^2} \sum_{j_1=1}^n ( |\langle f(\A)_{j_1}  , h(\A)_{j_1} \rangle \cdot \langle f(\A)_{j_1}, \A_{[j_1],i}\rangle -\langle f(\B)_{j_1}  , h(\B)_{j_1} \rangle \cdot \langle f(\B)_{j_1}, \B_{[j_1],i}\rangle | ~ + \notag \\
    & ~ | \langle f(\A)_{j_1} \circ \A_{[j_1],i}, h(\A)_{j_1} \rangle - \langle f(\B)_{j_1} \circ \B_{[j_1],i}, h(\B)_{j_1} \rangle| \notag ~ + \\
    & ~ |\langle f(\A)_{j_1}, \A_{[j_1],i}\rangle \cdot \langle f(\A)_{j_1}, {\bf 1}_n \rangle - \langle f(\B)_{j_1}, \B_{[j_1],i}\rangle \cdot \langle f(\B)_{j_1}, {\bf 1}_n \rangle | ~ + \notag \\
    & ~ |\langle f(\A)_{j_1}, \A_{[j_1],i} \rangle - \langle f(\B)_{j_1}, \B_{[j_1],i} \rangle| )^2
\end{align}

For the first term of Eq.~\eqref{eq:grad_L_cent:A}, we have
\begin{align} \label{eq:f_h_f_A:A}
    & |\langle f(\A)_{j_1}  , h(\A)_{j_1} \rangle \cdot \langle f(\A)_{j_1}, \A_{[j_1],i}\rangle -\langle f(\B)_{j_1}  , h(\B)_{j_1} \rangle \cdot \langle f(\B)_{j_1}, \B_{[j_1],i}\rangle | \notag \\
    \leq & ~ |\langle f(\A)_{j_1}  , h(\A)_{j_1} \rangle \cdot \langle f(\A)_{j_1}, \A_{[j_1],i}\rangle -\langle f(\A)_{j_1}  , h(\A)_{j_1} \rangle \cdot \langle f(\A)_{j_1}, \B_{[j_1],i}\rangle | ~ +\notag \\
    & ~ |\langle f(\A)_{j_1}  , h(\A)_{j_1} \rangle \cdot \langle f(\A)_{j_1}, \B_{[j_1],i}\rangle -\langle f(\A)_{j_1}  , h(\A)_{j_1} \rangle \cdot \langle f(\B)_{j_1}, \B_{[j_1],i}\rangle | ~ + \notag \\
    & ~ |\langle f(\A)_{j_1}  , h(\A)_{j_1} \rangle \cdot \langle f(\B)_{j_1}, \B_{[j_1],i}\rangle -\langle f(\A)_{j_1}  , h(\B)_{j_1} \rangle \cdot \langle f(\B)_{j_1}, \B_{[j_1],i}\rangle | ~ + \notag \\
    & ~ |\langle f(\A)_{j_1}  , h(\B)_{j_1} \rangle \cdot \langle f(\B)_{j_1}, \B_{[j_1],i}\rangle -\langle f(\B)_{j_1}  , h(\B)_{j_1} \rangle \cdot \langle f(\B)_{j_1}, \B_{[j_1],i}\rangle | \notag \\ 
    := & ~ C_1 + C_2 + C_3 + C_4
\end{align}

For the first term ($C_1$) of Eq.~\eqref{eq:f_h_f_A:A}, we have
\begin{align} \label{eq:f_h_f_A_p1:A}
    C_1 = & ~ |\langle f(\A)_{j_1}  , h(\A)_{j_1} \rangle \cdot \langle f(\A)_{j_1}, \A_{[j_1],i}\rangle -\langle f(\A)_{j_1}  , h(\A)_{j_1} \rangle \cdot \langle f(\A)_{j_1}, \B_{[j_1],i}\rangle | \notag \\
    = & ~ |\langle f(\A)_{j_1}  , h(\A)_{j_1} \rangle \cdot \langle f(\A)_{j_1}, \A_{[j_1],i} - \B_{[j_1],i}\rangle | \notag \\
    \leq & ~ \| f(\A)_{j_1} \|_2^2 \cdot \| h(\A)_{j_1} \|_2 \cdot \| \A_{[j_1],i} - \B_{[j_1],i} \|_2 \notag \\
    \leq & ~ \| h(\A)_{j_1} \|_2 \cdot \| \A_{[j_1],i} - \B_{[j_1],i} \|_2 \notag \\
    \leq & ~ 2 \sqrt{n} R^2 \cdot \| \A_{[j_1],i} - \B_{[j_1],i} \|_2 \notag \\
    \leq & ~ 2 \sqrt{n} R^2 \cdot \| \A_{[j_1],*} - \B_{[j_1],*} \|
\end{align}
where 1st step is inner product calculation, the 2nd step is due to Cauchy-Schwartz inequality, the 3rd step is from Lemma~\ref{lem:upper_bound:f}, the 4th step is from Lemma~\ref{lem:upper_bound_h:x}, the 5th step is because of definition of matrix norm.

For the second term ($C_2$) of Eq.~\eqref{eq:f_h_f_A:A}, we have
\begin{align} \label{eq:f_h_f_A_p2:A}
    C_2 = & ~ |\langle f(\A)_{j_1}  , h(\A)_{j_1} \rangle \cdot \langle f(\A)_{j_1}, \B_{[j_1],i}\rangle -\langle f(\A)_{j_1}  , h(\A)_{j_1} \rangle \cdot \langle f(\B)_{j_1}, \B_{[j_1],i}\rangle | \notag \\
    = & ~ |\langle f(\A)_{j_1}  , h(\A)_{j_1} \rangle \cdot \langle f(\A)_{j_1} - f(\B)_{j_1}, \B_{[j_1],i}\rangle | \notag \\
    \leq & ~ \| f(\A)_{j_1} \|_2 \cdot \| h(\A)_{j_1} \|_2 \cdot \| f(\A)_{j_1} - f(\B)_{j_1} \|_2 \cdot \| \B_{[j_1],i} \| \notag \\
    \leq & ~ \| h(\A)_{j_1} \|_2 \cdot \| f(\A)_{j_1} - f(\B)_{j_1} \|_2 \cdot \| \B_{[j_1],i} \notag \| \\
    \leq & ~ 2 \sqrt{n} R^2 \cdot \| f(\A)_{j_1} - f(\B)_{j_1} \|_2 \cdot \| \B_{[j_1],i} \notag \| \\
    \leq & ~ 2 \sqrt{n} R^3 \cdot \| f(\A)_{j_1} - f(\B)_{j_1} \|_2 \notag \\
    \leq & ~ 8 n^2 R^4 \beta^{-2} \exp(2 R^2) \cdot \| \A_{[j_1],*} - \B_{[j_1],*} \|
\end{align}
where 1st step is inner product calculation, the 2nd step is due to Cauchy-Schwartz inequality, the 3rd step is from Lemma~\ref{lem:upper_bound:f}, the 4th step is from Lemma~\ref{lem:upper_bound_h:x}, the 5th step is because of $\max_{j_1 \in [n]} \| \B_{[j_1],*} \| \leq R$, the 6th step is due to Lemma~\ref{lem:lipschitz_f:A}.

For the third term ($C_3$) of Eq.~\eqref{eq:f_h_f_A:A}, we have
\begin{align} \label{eq:f_h_f_A_p3:A}
    C_3 = & ~ |\langle f(\A)_{j_1}  , h(\A)_{j_1} \rangle \cdot \langle f(\B)_{j_1}, \B_{[j_1],i}\rangle -\langle f(\A)_{j_1}  , h(\B)_{j_1} \rangle \cdot \langle f(\B)_{j_1}, \B_{[j_1],i}\rangle | \notag \\
    = & ~ |\langle f(\A)_{j_1}  , h(\A)_{j_1} - h(\B)_{j_1} \rangle \cdot \langle f(\B)_{j_1} , \B_{[j_1],i}\rangle | \notag \\
    \leq & ~ \| f(\A)_{j_1} \|_2 \cdot \| h(\A)_{j_1}  - h(\B)_{j_1}\|_2 \cdot \| f(\B)_{j_1} \|_2 \cdot \| \B_{[j_1],i} \| \notag \\
    \leq & ~ \| h(\A)_{j_1} - h(\B)_{j_1} \|_2 \cdot \| \B_{[j_1],i} \| \notag \\
    \leq & ~ \| f(\A)_{j_1} - f(\B)_{j_1} \|_2 \cdot \| \B_{[j_1],i} \| \notag \\
    \leq & ~ R \cdot \| f(\A)_{j_1} - f(\B)_{j_1} \|_2 \notag \\
    \leq & ~ 4 n^{1.5} R^2 \beta^{-2} \exp(2 R^2) \cdot \| \A_{[j_1],*} - \B_{[j_1],*} \|
\end{align}
where 1st step is inner product calculation, the 2nd step is due to Cauchy-Schwartz inequality, the 3rd step is from Lemma~\ref{lem:upper_bound:f}, the 4th step is from Lemma~\ref{lem:lipschitz_h:A}, the 5th step is because of $\max_{j_1 \in [n]} \| \B_{[j_1],*} \| \leq R$, the 6th step is due to Lemma~\ref{lem:lipschitz_f:A}.

For the fourth term ($C_4$) of Eq.~\eqref{eq:f_h_f_A:A}, we have
\begin{align} \label{eq:f_h_f_A_p4:A}
    C_4 = & ~ |\langle f(\A)_{j_1}  , h(\B)_{j_1} \rangle \cdot \langle f(\B)_{j_1}, \B_{[j_1],i}\rangle -\langle f(\B)_{j_1}  , h(\B)_{j_1} \rangle \cdot \langle f(\B)_{j_1}, \B_{[j_1],i}\rangle | \notag \\
    = & ~ |\langle f(\A)_{j_1} - f(\B)_{j_1} , h(\B)_{j_1} \rangle \cdot \langle f(\B)_{j_1} , \B_{[j_1],i}\rangle | \notag \\
    \leq & ~ \| f(\A)_{j_1} - f(\B)_{j_1} \|_2 \cdot \| h(\B)_{j_1} \|_2 \cdot \| f(\B)_{j_1} \|_2 \cdot \| \B_{[j_1],i} \| \notag \\
    \leq & ~ \| h(\B)_{j_1} \|_2 \cdot \| f(\A)_{j_1} - f(\B)_{j_1} \|_2 \cdot \| \B_{[j_1],i} \notag \| \\
    \leq & ~ 2 \sqrt{n} R^2 \cdot \| f(\A)_{j_1} - f(\B)_{j_1} \|_2 \cdot \| \B_{[j_1],i} \| \notag \\
    \leq & ~ 2 \sqrt{n} R^3 \cdot \| f(\A)_{j_1} - f(\B)_{j_1} \|_2 \notag \\
    \leq & ~ 8 n^2 R^4 \beta^{-2} \exp(2 R^2) \cdot \| \A_{[j_1],*} - \B_{[j_1],*} \|
\end{align}
where 1st step is inner product calculation, the 2nd step is due to Cauchy-Schwartz inequality, the 3rd step is from Lemma~\ref{lem:upper_bound:f}, the 4th step is from Lemma~\ref{lem:upper_bound_h:x}, the 5th step is because of $\max_{j_1 \in [n]} \| \B_{[j_1],*} \| \leq R$, the 6th step is due to Lemma~\ref{lem:lipschitz_f:A}.

Putting Eq.~\eqref{eq:f_h_f_A_p1:A}, Eq.~\eqref{eq:f_h_f_A_p2:A}, Eq.~\eqref{eq:f_h_f_A_p3:A} and Eq.~\eqref{eq:f_h_f_A_p4:A} into Eq.~\eqref{eq:f_h_f_A:A}, we have
\begin{align} \label{eq:f_h_f_A_whole:A}
    & |\langle f(\A)_{j_1}  , h(\A)_{j_1} \rangle \cdot \langle f(\A)_{j_1}, \A_{[j_1],i}\rangle -\langle f(\B)_{j_1}  , h(\B)_{j_1} \rangle \cdot \langle f(\B)_{j_1}, \B_{[j_1],i}\rangle | \notag \\
    \leq & ~ C_1 + C_2 + C_3 + C_4 \notag \\
    \leq & ~ (1+ 4 n^{1.5} R^2 \beta^{-2} \exp(2 R^2) + 16 n^2 R^4 \beta^{-2} \exp(2 R^2)) \cdot \| \A_{[j_1],*} - \B_{[j_1],*} \|
\end{align}

For the second term of Eq.~\eqref{eq:grad_L_ent:A}, we have
\begin{align} \label{eq:f_A_h:A}
    & | \langle f(\A)_{j_1} \circ \A_{[j_1],i}, h(\A)_{j_1} \rangle - \langle f(\B)_{j_1} \circ \B_{[j_1],i}, h(\B)_{j_1} \rangle| \notag \\
    \leq & ~ | \langle f(\A)_{j_1} \circ \A_{[j_1],i}, h(\A)_{j_1} \rangle - \langle f(\A)_{j_1} \circ \A_{[j_1],i}, h(\B)_{j_1} \rangle| ~ + \notag \\
    & ~ | \langle f(\A)_{j_1} \circ \A_{[j_1],i}, h(\B)_{j_1} \rangle - \langle f(\A)_{j_1} \circ \B_{[j_1],i}, h(\B)_{j_1} \rangle| ~ + \notag \\
    & ~ | \langle f(\A)_{j_1} \circ \B_{[j_1],i}, h(\B)_{j_1} \rangle - \langle f(\B)_{j_1} \circ \B_{[j_1],i}, h(\B)_{j_1} \rangle| \notag \\
    := & ~ C_5 + C_6 + C_7
\end{align}

For the first item ($C_5$) of Eq.~\eqref{eq:f_A_h:A}, we have
\begin{align} \label{eq:f_A_h_p1:A}
    C_5 = & ~ | \langle f(\A)_{j_1} \circ \A_{[j_1],i}, h(\A)_{j_1} \rangle - \langle f(\A)_{j_1} \circ \A_{[j_1],i}, h(\B)_{j_1} \rangle| \notag \\
    = & ~ | \langle f(\A)_{j_1} \circ \A_{[j_1],i}, h(\A)_{j_1} - h(\B)_{j_1} \rangle| \notag \\
    \leq & ~ \| f(\A)_{j_1} \circ \A_{[j_1],i} \|_2 \cdot \| h(\A)_{j_1} - h(\B)_{j_1} \|_2 \notag \\
    \leq & ~ \| f(\A)_{j_1} \|_\infty \cdot \| \A_{[j_1],i} \|_2 \cdot \| h(\A)_{j_1} - h(\B)_{j_1} \|_2 \notag \\
    \leq & ~ \| f(\A)_{j_1} \|_2 \cdot \| \A_{[j_1],i} \|_2 \cdot \| h(\A)_{j_1} - h(\B)_{j_1} \|_2 \notag \\
    \leq & ~  \| \A_{[j_1],i} \|_2 \cdot \| h(\A)_{j_1} - h(\B)_{j_1} \|_2 \notag \\
    \leq & ~ R \cdot \| h(\A)_{j_1} - h(\B)_{j_1} \|_2 \notag \\
    \leq & ~ R \cdot \| f(\A)_{j_1} - f(\B)_{j_1} \|_2 \notag \\
    \leq & ~ 4 n^{1.5} R^2 \beta^{-2} \exp(2 R^2) \cdot \| \A_{[j_1],*} - \B_{[j_1],*} \|
\end{align}
where the 1st step is inner product calculation, the 2nd step is due to Cauchy-Schwartz inequality, the 3rd and the 4th step is owing to Fact~\ref{fac:vector_norm}, the 5th step follows from Lemma~\ref{lem:upper_bound:f}, the 6th step holds because $\max_{j_1 \in [n]} \| \A_{[j_1],*} \| \leq R$, the last step follows by Lemma~\ref{lem:lipschitz_f:A}.

For the second item ($C_6$) of Eq.~\eqref{eq:f_A_h:A}, we have
\begin{align} \label{eq:f_A_h_p2:A}
    C_6 = & ~ | \langle f(\A)_{j_1} \circ \A_{[j_1],i}, h(\B)_{j_1} \rangle - \langle f(\A)_{j_1} \circ \B_{[j_1],i}, h(\B)_{j_1} \rangle| \notag \\
    = & ~ | \langle f(\A)_{j_1} \circ \A_{[j_1],i} - f(\A)_{j_1} \circ \B_{[j_1],i}, h(\B)_{j_1} \rangle| \notag \\
    = & ~ | \langle f(\A)_{j_1} \circ (\A_{[j_1],i} - \B_{[j_1],i}), h(\B)_{j_1} \rangle| \notag \\
    \leq & ~ \| f(\A)_{j_1} \circ (\A_{[j_1],i} - \B_{[j_1],i}) \|_2 \cdot \| h(\B)_{j_1} \|_2 \notag \\
    \leq & ~ \| f(\A)_{j_1} \|_\infty \cdot \| \A_{[j_1],i} - \B_{[j_1],i} \|_2 \cdot \|  h(\B)_{j_1} \|_2 \notag \\
    \leq & ~ \| f(\A)_{j_1} \|_2 \cdot \| \A_{[j_1],i} - \B_{[j_1],i} \|_2 \cdot \| h(\B)_{j_1} \|_2 \notag \\
    \leq & ~  \| \A_{[j_1],i} - \B_{[j_1],i} \|_2 \cdot \| h(\B)_{j_1} \|_2 \notag \\
    \leq & ~  2 \sqrt{n} R^2 \cdot \| \A_{[j_1],i} - \B_{[j_1],i} \|_2 \notag \\
    \leq & ~ 2 \sqrt{n} R^2 \cdot \| \A_{[j_1],*} - \B_{[j_1],*} \|
\end{align}
where the 1st step is inner product calculation, the 2nd step is Hadamard product calculation, the 3rd step is due to Cauchy-Schwartz inequality, the 4th and the 5th step is owing to Fact~\ref{fac:vector_norm}, the 6th step follows from Lemma~\ref{lem:upper_bound:f}, the 7th step is derived from Lemma~\ref{sec:upper_bound:h}, the last step follows by definition of matrix norm.

For the third item ($C_7$) of Eq.~\eqref{eq:f_A_h:A}, we have
\begin{align} \label{eq:f_A_h_p3:A}
    C_7 = & ~ | \langle f(\A)_{j_1} \circ \B_{[j_1],i}, h(\B)_{j_1} \rangle - \langle f(\B)_{j_1} \circ \B_{[j_1],i}, h(\B)_{j_1} \rangle| \notag \\
    = & ~ | \langle f(\A)_{j_1} \circ \B_{[j_1],i} - f(\B)_{j_1} \circ \B_{[j_1],i}, h(\B)_{j_1} \rangle| \notag \\
    = & ~ | \langle (f(\A)_{j_1} - f(\B)_{j_1}) \circ \B_{[j_1],i}, h(\B)_{j_1} \rangle| \notag \\
    \leq & ~ \| f(\A)_{j_1}- f(\B)_{j_1}) \circ \B_{[j_1],i}  \|_2 \cdot \| h(\B)_{j_1} \|_2 \notag \\
    \leq & ~ \| f(\A)_{j_1} - f(\B)_{j_1} \|_\infty \cdot \| \B_{[j_1],i}  \|_2 \cdot \|  h(\B)_{j_1} \|_2 \notag \\
    \leq & ~ \| f(\A)_{j_1} - f(\B)_{j_1} \|_2 \cdot \| \B_{[j_1],i}  \|_2 \cdot \|  h(\B)_{j_1} \|_2 \notag \\
    \leq & ~ R \cdot \| f(\A)_{j_1} - f(\B)_{j_1} \|_2  \cdot \|  h(\B)_{j_1} \|_2 \notag \\
    \leq & ~ 2 \sqrt{n} R^3 \cdot \| f(\A)_{j_1} - f(\B)_{j_1} \|_2 \notag \\
    \leq & ~ 8 n^2 R^4 \beta^{-2} \exp(2 R^2) \cdot \| \A_{[j_1],*} - \B_{[j_1],*} \|
\end{align}
where the 1st step is inner product calculation, the 2nd step is Hadamard product calculation, the 3rd step is due to Cauchy-Schwartz inequality, the 4th and the 5th step is owing to Fact~\ref{fac:vector_norm}, the 6th step follows from $\max_{j_1 \in [n]} \| \B_{[j_1],*} \| \leq R$, the 7th step is derived from Lemma~\ref{sec:upper_bound:h}, the last step follows by Lemma~\ref{lem:lipschitz_f:A}.

Putting Eq.~\eqref{eq:f_A_h_p1:A}, Eq.~\eqref{eq:f_A_h_p2:A}, and Eq.~\eqref{eq:f_A_h_p3:A} into Eq.~\eqref{eq:f_A_h:A}, we have
\begin{align} \label{eq:f_A_h_whole:A}
    & | \langle f(\A)_{j_1} \circ \A_{[j_1],i}, h(\A)_{j_1} \rangle - \langle f(\B)_{j_1} \circ \B_{[j_1],i}, h(\B)_{j_1} \rangle| \notag \\
    \leq & ~ C_5 + C_6 + C_7 \notag \\
    \leq & ~ (2 \sqrt{n} R^2 + 4 n^{1.5} R^2 \beta^{-2} \exp(2 R^2) + 8 n^2 R^4 \beta^{-2} \exp(2 R^2)) \cdot \| \A_{[j_1],*} - \B_{[j_1],*} \|
\end{align}

For the third term of Eq.~\eqref{eq:grad_L_ent:A}, we have
\begin{align} \label{eq:f_A_f_1:A}
    & |\langle f(\A)_{j_1}, \A_{[j_1],i}\rangle \cdot \langle f(\A)_{j_1}, {\bf 1}_n \rangle - \langle f(\B)_{j_1}, \B_{[j_1],i}\rangle \cdot \langle f(\B)_{j_1}, {\bf 1}_n \rangle | \notag \\
    \leq & ~ |\langle f(\A)_{j_1}, \A_{[j_1],i}\rangle \cdot \langle f(\A)_{j_1}, {\bf 1}_n \rangle - \langle f(\A)_{j_1}, \A_{[j_1],i}\rangle \cdot \langle f(\B)_{j_1}, {\bf 1}_n \rangle | ~ + \notag \\
    & ~ |\langle f(\A)_{j_1}, \A_{[j_1],i}\rangle \cdot \langle f(\B)_{j_1}, {\bf 1}_n \rangle - \langle f(\A)_{j_1}, \A_{[j_1],i}\rangle \cdot \langle f(\B)_{j_1}, {\bf 1}_n \rangle | ~ + \notag \\
    & ~ |\langle f(\A)_{j_1}, \B_{[j_1],i}\rangle \cdot \langle f(\B)_{j_1}, {\bf 1}_n \rangle - \langle f(\B)_{j_1}, \B_{[j_1],i}\rangle \cdot \langle f(\B)_{j_1}, {\bf 1}_n \rangle | \notag \\
    := & ~ C_8 + C_9 + C_{10}
\end{align}

For the first item ($C_8$) of Eq.~\eqref{eq:f_A_f_1:A}, we have
\begin{align} \label{eq:f_A_f_1_p1:A}
    C_8 = & ~ |\langle f(\A)_{j_1}, \A_{[j_1],i}\rangle \cdot \langle f(\A)_{j_1}, {\bf 1}_n \rangle - \langle f(\A)_{j_1}, \A_{[j_1],i}\rangle \cdot \langle f(\B)_{j_1}, {\bf 1}_n \rangle | \notag \\
    = & ~ |\langle f(\A)_{j_1}, \A_{[j_1],i}\rangle \cdot \langle f(\A)_{j_1} - f(\B)_{j_1}, {\bf 1}_n \rangle | \notag \\
    \leq & ~ \| f(\A)_{j_1} \|_2 \cdot \| \A_{[j_1],i} \|_2 \cdot \| f(\A)_{j_1} - f(\B)_{j_1} \|_2 \cdot \| {\bf 1}_n \|_2 \notag \\
    \leq & ~ \| \A_{[j_1],i} \|_2 \cdot \| f(\A)_{j_1} - f(\B)_{j_1} \|_2 \cdot \| {\bf 1}_n \|_2 \notag \\
    \leq & ~ R \cdot \| f(\A)_{j_1} - f(\B)_{j_1} \|_2 \cdot \| {\bf 1}_n \|_2 \notag \\
    \leq & ~ \sqrt{n} R \cdot \| f(\A)_{j_1} - f(\B)_{j_1} \|_2 \notag \\
    \leq & ~ 4 n^2 R^2 \beta^{-2} \exp(2 R^2) \cdot \| \A_{[j_1],*} - \B_{[j_1],*} \|
\end{align}
where the 1st step is inner product calculation, the 2nd step is given by Cauchy-Schwartz inequality, the 3rd step is due to Lemma~\ref{sec:upper_bound:f}, the 4th step holds because $\max_{j_1 \in [n]} \| \A_{[j_1],*} \| \leq R$, the last step is due to Lemma~\ref{lem:lipschitz_f:A}.

For the second item ($C_9$) of Eq.~\eqref{eq:f_A_f_1:A}, we have
\begin{align} \label{eq:f_A_f_1_p2:A}
    C_9 = & ~ |\langle f(\A)_{j_1}, \A_{[j_1],i}\rangle \cdot \langle f(\B)_{j_1}, {\bf 1}_n \rangle - \langle f(\A)_{j_1}, \B_{[j_1],i}\rangle \cdot \langle f(\B)_{j_1}, {\bf 1}_n \rangle | \notag \\
    = & ~ |\langle f(\A)_{j_1}, \A_{[j_1],i} - \B_{[j_1],i} \rangle \cdot \langle f(\B)_{j_1}, {\bf 1}_n \rangle | \notag \\
    \leq & ~ \| f(\A)_{j_1} \|_2 \cdot \| \A_{[j_1],i} - \B_{[j_1],i} \|_2 \cdot \| f(\B)_{j_1} \|_2 \cdot \| {\bf 1}_n \|_2 \notag \\
    \leq & ~ \| \A_{[j_1],i} \|_2 \cdot \| \A_{[j_1],i} - \B_{[j_1],*} \|_2 \cdot \| {\bf 1}_n \|_2 \notag \\
    \leq & ~ R \cdot\| \A_{[j_1],i} - \B_{[j_1],*} \|_2 \cdot \| {\bf 1}_n \|_2 \notag \\
    \leq & ~ \sqrt{n} R \cdot \| \A_{[j_1],i} - \B_{[j_1],i} \|_2 \notag \\
    \leq & ~ \sqrt{n} R \cdot \| \A_{[j_1],*} - \B_{[j_1],*}  \|
\end{align}
where the 1st step is inner product calculation, the 2nd step is given by Cauchy-Schwartz inequality, the 3rd step is due to Lemma~\ref{sec:upper_bound:f}, the 4th step holds because $\max_{j_1 \in [n]} \| \A_{[j_1],*} \| \leq R$, the last step uses the definition of matrix norm.

For the third item ($C_10$) of Eq.~\eqref{eq:f_A_f_1:A}, we have
\begin{align} \label{eq:f_A_f_1_p3:A}
    C_{10} = & ~ |\langle f(\A)_{j_1}, \B_{[j_1],i}\rangle \cdot \langle f(\B)_{j_1}, {\bf 1}_n \rangle - \langle f(\B)_{j_1}, \B_{[j_1],i}\rangle \cdot \langle f(\B)_{j_1}, {\bf 1}_n \rangle | \notag \\
    = & ~ |\langle f(\A)_{j_1} - f(\B)_{j_1}, \B_{[j_1],i}\rangle \cdot \langle f(\B)_{j_1}, {\bf 1}_n \rangle | \notag \\
    \leq & ~ \| f(\A)_{j_1} - f(\B)_{j_1} \|_2 \cdot \| \B_{[j_1],i} \|_2 \cdot \| f(\B)_{j_1} \|_2 \cdot \| {\bf 1}_n \|_2 \notag \\
    \leq & ~ \| \B_{[j_1],i} \|_2 \cdot \| f(\A)_{j_1} - f(\B)_{j_1} \|_2 \cdot \| {\bf 1}_n \|_2 \notag \\
    \leq & ~ R \cdot \| f(\A)_{j_1} - f(\B)_{j_1} \|_2 \cdot \| {\bf 1}_n \|_2 \notag \\
    \leq & ~ \sqrt{n} R \cdot \| f(\A)_{j_1} - f(\B)_{j_1} \|_2 \notag \\
    \leq & ~ 4 n^2 R^2 \beta^{-2} \exp(2 R^2) \cdot \| \A_{[j_1],*} - \B_{[j_1],*} \|
\end{align}
where the 1st step is inner product calculation, the 2nd step is given by Cauchy-Schwartz inequality, the 3rd step is due to Lemma~\ref{sec:upper_bound:f}, the 4th step holds because $\max_{j_1 \in [n]} \| \B_{[j_1],*} \| \leq R$, the last step is due to Lemma~\ref{lem:lipschitz_f:A}.

Putting Eq.~\eqref{eq:f_A_f_1_p1:A}, Eq.~\eqref{eq:f_A_f_1_p2:A}, and Eq.~\eqref{eq:f_A_f_1_p3:A} into Eq.~\eqref{eq:f_A_f_1:A}, we have
\begin{align} \label{eq:f_A_f_1_whole:A}
    & |\langle f(\A)_{j_1}, \A_{[j_1],i}\rangle \cdot \langle f(\A)_{j_1}, {\bf 1}_n \rangle - \langle f(\B)_{j_1}, \B_{[j_1],i}\rangle \cdot \langle f(\B)_{j_1}, {\bf 1}_n \rangle | \notag \\
    \leq & ~ C_8 + C_9 + C_{10} \notag \\
    \leq & ~ (\sqrt{n} R + 8 n^2 R^2 \beta^{-2} \exp(2 R^2) ) \cdot \| \A_{[j_1],*} - \B_{[j_1],*} \|
\end{align}

For the fourth term of Eq.~\eqref{eq:grad_L_ent:A}, we have
\begin{align} \label{eq:f_A:A}
    & |\langle f(\A)_{j_1}, \A_{[j_1],i} \rangle - \langle f(\B)_{j_1}, \B_{[j_1],i} \rangle| \notag \\
    \leq & ~ |\langle f(\A)_{j_1}, \A_{[j_1],i} \rangle - \langle f(\A)_{j_1}, \B_{[j_1],i} \rangle| + |\langle f(\A)_{j_1}, \B_{[j_1],i} \rangle - \langle f(\B)_{j_1}, \B_{[j_1],i} \rangle| \notag \\
    = & ~ |\langle f(\A)_{j_1}, \A_{[j_1],i} - \B_{[j_1],i} \rangle| + |\langle f(\A)_{j_1} - f(\B)_{j_1}, \B_{[j_1],i} \rangle| \notag \\
    \leq & ~ \| f(\A)_{j_1} \|_2 \cdot \| \A_{[j_1],i} - \B_{[j_1],i} \|_2 + \| f(\A)_{j_1} - f(\B)_{j_1} \|_2 \cdot \| \B_{[j_1],i} \|_2 \notag \\
    \leq & ~ \| \A_{[j_1],i} - \B_{[j_1],i} \|_2 + \| f(\A)_{j_1} - f(\B)_{j_1} \|_2 \cdot \| \B_{[j_1],i} \|_2 \notag \\
    \leq & ~ \| \A_{[j_1],i} - \B_{[j_1],i} \|_2 + R \cdot \| f(\A)_{j_1} -  f(\B)_{j_1} \|_2 \ \notag \\
    \leq & ~ \| \A_{[j_1],*} - \B_{[j_1],*} \| + \| f(\A)_{j_1} - f(\B)_{j_1} \|_2 \notag \\
    \leq & ~ (1 + 4 n^{1.5} R^2 \beta^{-2} \exp(2R^2)) \cdot \| \A_{[j_1],*} - \B_{[j_1],*} \|
\end{align}
where the 1st step is from triangle inequality, the 2nd step is inner product calculation, the 3rd step uses Cauchy-Schwartz inequality, the 4th step is due to Lemma~\ref{sec:upper_bound:f}, the 5th step follows by $\max_{j_1 \in [n]} \| \B_{[j_1],*} \| \leq R$, the 6th step is given by matrix norm definition, the 7th step follows by Lemma~\ref{lem:lipschitz_f:A}. 

Putting Eq.~\eqref{eq:f_h_f_A_whole:A}, Eq.~\eqref{eq:f_A_h_whole:A} , Eq.~\eqref{eq:f_A_f_1_whole:A} and Eq.~\eqref{eq:f_A:A} into Eq.~\eqref{eq:grad_L_ent:A}, we have
\begin{align*}
    & \| \nabla L_{\ent}(\A) - \nabla L_{\ent}(\B) \|_2^2 \notag \\
    \leq & ~ \sum_{i=1}^{d^2} \sum_{j_1=1}^n (2 + \sqrt{n} R + 2 \sqrt{n} R^2 + 12 n^{1.5} R^2 \beta^{-2} \exp(2 R^2) ~ + \notag \\ 
    & ~ 4 n^2 R^2 \beta^{-2} \exp(2R^2) + 24 n^2 R^4 \beta^{-2} \exp(2 R^2))  \cdot \| \A_{[J_1],*} - \B_{[j_1],*} \|)^2 \\
    \leq & ~ \sum_{i=1}^{d^2} \sum_{j_1=1}^n (2 + \sqrt{n} R + 2 \sqrt{n} R^2 + 12 n^{1.5} R^2 \exp(4 R^2) ~ + \notag \\ 
    & ~ 4 n^2 R^2 \exp(4R^2) + 24 n^2 R^4 \exp(4 R^2))  \cdot \| \A_{[j_1],*} - \B_{[j_1],*} \|)^2 \\
    \leq & ~ d^2n \cdot ( n^2 \exp(5 R^2) \cdot \| \A - \B \|_{\infty,2})^2
\end{align*}
where the 2nd step is from Lemma~\ref{lem:lower_bound_A:beta}, the 3rd step is because $R > 4$.

Therefore, we have
\begin{align*}
    \| \nabla L_{\ent}(x) - \nabla L_{\ent}(y) \|_2 \leq dn^{2.5} \exp(5 R^2) \cdot \| \A - \B \|_{\infty,2}
\end{align*}

\end{proof}
\end{lemma}

\section{In-Context Learning for Rescaled Version} \label{sec:application_rescaled}

To bring the Lipschitz property into an in-context learning application, we will introduce two lemmas in Section~\ref{sec:application_rescaled:x} and Section~\ref{sec:application_rescaled:A}. Additionally, a new definition will be presented to assist in establishing the main result in Section~\ref{sec:application_rescaled:main_result}.

\subsection{For the \texorpdfstring{$x$}{} case}\label{sec:application_rescaled:x}

\begin{definition}\label{def:delta_b:x}
Let $\delta_q \in \R^{n^2}$ be the vector that fulfills the following condition
\begin{align*}
    \|  \exp(\A x_{t+1}) - D(x_{t+1}) b \|_2^2 = \|   \exp(\A x_{t}) - D(x_{t}) (b-\delta_q) \|_2^2
\end{align*}
\end{definition}

\begin{lemma} \label{lem:upper_bound:x}
Provided that the subsequent requirement are satisfied
\begin{itemize}
    \item Let $\delta_q$ be defined in Definition~\ref{def:delta_b:x}.
     \item Let $R \geq 4$.
    \item Let $x_t \in \R^{d^2}, x_{t+1} \in \R^{d^2}$ satisfy $\| x_{t} \|_2 \leq R$ and $\| x_{t+1} \|_2 \leq R$
    \item Let $\A \in \R^{n^2 \times d^2}$
    \item Let $\max_{j_1 \in [n]} \| \A_{[j_1],*} (x_{t}-x_{t+1}) \|_{\infty} < 0.01$
    \item Let $\max_{j_1 \in [n]}\| \A_{[j_1],*} \| \leq R$
    \item Let $\max_{j_1 \in [n]} \| b_{[j_1]} \|_2 \leq 1$
    \item Let $M = \exp(O(R^2 + \log n))$.
\end{itemize}
We have
\begin{align*}
   \| \delta_q \|_2 \leq M \cdot \| x_{t+1} - x_t \|_2
\end{align*}
\end{lemma}
\begin{proof}
    We have 
    \begin{align*}
         \exp(\A x_{t+1}) -  D(x_{t+1}) \cdot b  = (   \exp(\A x_{t}) - D(x_{t}) \cdot (b-\delta_q) ) \circ \{ -1 , +1 \}^{n^2}
    \end{align*}
   Since all solutions have the same norm, we can simplify the problem by focusing on a single solution, namely the one presented below.
    \begin{align*}
         \underbrace{ \exp(\A x_{t+1}) }_{n^2 \times 1 \mathrm{~vector}} - \underbrace{ D(x_{t+1}) }_{n^2 \times n^2 \mathrm{~matrix}} \cdot \underbrace{ b }_{n^2 \times 1 \mathrm{~vector}}  = \exp(\A x_{t}) - D(x_{t}) \cdot (b-\delta_q)
    \end{align*}

    One possible choice for $\delta_q \in \R^{n^2}$ is
    \begin{align}\label{eq:delta_b}
     \delta_q = D(x_t)^{-1} (\exp(\A x_{t+1}) - D(x_{t+1}) \cdot b - (\exp(\A x_{t}) - D(x_{t})\cdot b))
    \end{align}

In particular, we have
\begin{align}\label{eq:delta_b_j}
   \underbrace{ (\delta_q)_{[j]} }_{n \times 1 \mathrm{~vector}} = \alpha(x_t)^{-1}_{j} \cdot ( u(x_{t+1})_j - \alpha(x_{t+1})_j \cdot b_{[j]} ) - \alpha(x_t)^{-1}_j \cdot ( u(x_{t})_j - \alpha(x_{t})_j \cdot b_{[j]})
\end{align}

    Thus,
    \begin{align*}
        \| \delta_q \|_2^2
        = & ~ \| D(x_t)^{-1} (\exp(\A x_{t+1}) - D(x_{t+1}) \cdot b - (\exp(\A x_{t}) - D(x_{t})\cdot b)) \|_2^2 \\
        = & ~ \sum_{j=1}^n \| \alpha(x_t)^{-1}_{j} \cdot ( u(x_{t+1})_j - \alpha(x_{t+1})_j \cdot b_{[j]} ) - \alpha(x_t)^{-1}_j \cdot ( u(x_{t})_j - \alpha(x_{t})_j \cdot b_{[j]}) \|_2^2 \\ 
        = & ~ \sum_{j=1}^n \| \alpha(x_t)^{-1}_j \cdot (q(x_{t+1})_j - q(x_t)_j ) \|_2^2 \\
        = & ~ \sum_{j=1}^n | \alpha(x_t)^{-1}_{j} |^2 \cdot \| q(x_{t+1} )_j - q(x_{t})_j\|_2^2 \\ 
        \leq & ~ \sum_{j=1}^n (\exp(R^2))^2 \cdot \| q(x_{t+1} )_j - q(x_{t})_j\|_2^2 \\
        \leq & ~ n \cdot (\exp(R^2))^2 \cdot (4nR \exp(R^2))^2 \cdot \| x_{t+1} - x_t \|_2^2 \\
        \leq & ~ M^2 \cdot \| x_{t+1} - x_t \|_2^2   
    \end{align*}
    where the first step follows from Eq.~\eqref{eq:delta_b}, 
    the 2nd step follows from Eq.~\eqref{eq:delta_b_j}, 
    the 3rd step follows from definition of function $q(x)$ (see Definition~\ref{def:q}), 
    the 4th step is based on Fact~\ref{fac:matrix_norm},
    the 5th step is because of Lemma~\ref{lem:upper_bound:alpha_inverse},
    the 6th is because of Lemma~\ref{lem:Lipschitz_q:x} and the last step follows from Definition of $M$ in Lemma statement.

Thus, we have shown that
\begin{align*}
\| \delta_q \|_2 \leq M \cdot \| x_{t+1} - x_t \|_2
\end{align*}
        
The proof is complete.
\end{proof}
\subsection{For the \texorpdfstring{$A$}{} case}\label{sec:application_rescaled:A}
\begin{definition}\label{def:delta_b:A}
Let $\delta_q \in \R^{n^2}$ be the vector that fulfills the following condition
\begin{align*}
    \| \exp(\A_{t+1}) -D(\A_{t+1}) b \|_2^2 = \|   \exp(\A_{t}) - D(\A_{t})(b-\delta_q) \|_2^2
\end{align*}
\end{definition}

\begin{lemma} \label{lem:upper_bound:A}
Provided that the subsequent requirement are satisfied
\begin{itemize}
    \item Let $\delta_q$ be defined in Definition~\ref{def:delta_b:A}.
    \item Let $M = \exp(O(R^2 + \log n))$
\end{itemize}
We have
\begin{align*}
    \| \delta_q \|_2 \leq M \cdot \| A_{t+1} - A_t \|_2
\end{align*}
\end{lemma}

\subsection{Main Result}\label{sec:application_rescaled:main_result}

Now, we will apply the results mentioned above to in-context learning. The formal version of Theorem~\ref{thm:in_context_rescaled} is presented as follows.

\begin{theorem}[Learning in-context for Rescaled Version]\label{thm:in_context_rescaled}
Provided that the subsequent requirement are satisfied
\begin{itemize}
    \item Let $R \geq 4$.
    \item Let $x_t \in \R^{d^2}, x_{t+1} \in \R^{d^2}$ satisfy $\| x_{t} \|_2 \leq R$ and $\| x_{t+1} \|_2 \leq R$
    \item Let $\A \in \R^{n^2 \times d^2}$
    \item Let $\max_{j_1 \in [n]} \| \A_{[j_1],*} (x_{t}-x_{t+1}) \|_{\infty} < 0.01$
    \item Let $\max_{j_1 \in [n]}\| \A_{[j_1],*} \| \leq R$
    \item Let $\max_{j_1 \in [n]} \| b_{[j_1]} \|_2 \leq 1$
    \item Let $M = \exp(O(R^2 + \log n))$.
\end{itemize}
We consider the matrix formulation for attention regression (Definition~\ref{def:intro_normalized_matrix}) problem
\begin{align*}
     \min_{x \in \R^{d^2}} \|  \exp(\A x) - D(x) b \|_2^2.
\end{align*}
\begin{itemize}
\item {\bf Part 1.} By transitioning $x_t$ to $x_{t+1}$, we are effectively addressing a fresh rescaled softmax regression problem involving
\begin{align*}
      \min_{x \in \R^{d^2}} \|  \exp(\A x) - D(x) \wt{b} \|_2^2
\end{align*}
where 
\begin{align*}
    \| \wt{b} - b \|_2 \leq M \cdot \| x_{t+1} - x_t \|_2
\end{align*}
\item {\bf Part 2.}  By transitioning $A_t$ to $A_{t+1}$, we are effectively addressing a fresh rescaled softmax regression problem involving
\begin{align*}
       \min_{x \in \R^{d^2}} \|\exp(\A x) -  D(x)\hat{b} \|_2^2
\end{align*}
where 
\begin{align*}
    \| \wh{b} - b \|_2 \leq M \cdot \| A_{t+1} - A_t \|
\end{align*}
\end{itemize}
\end{theorem}

\begin{proof}
The proof straightforwardly derives from Lemma~\ref{lem:upper_bound:x} and Lemma~\ref{lem:upper_bound:A}.
\end{proof}

\section{In-Context Learning for Normalized Version}\label{sec:application_normalized}
In this section, we introduce our analysis about in-context learning for normalized version. In Section~\ref{sec:application_normalized:x} and Section~\ref{sec:application_normalized:A}, Lipschitz result for $A$ and $x$ are presented. In Section~\ref{sec:application_normalized:main_result}, the main result is presented by us. 
\begin{definition}\label{def:delta_c:x}
Let $\delta_c \in \R^{n^2}$ be the vector that fulfills the following condition
\begin{align*}
    \|  D(x_{t+1})^{-1}\exp(\A x_{t+1}) -  b \|_2^2 = \|   D(x_{t})^{-1}\exp(\A x_{t}) -  (b-\delta_c) \|_2^2
\end{align*}
\end{definition}

\subsection{For the \texorpdfstring{$x$}{} case}\label{sec:application_normalized:x}
\begin{lemma} \label{lem:upper_bound_normalized:x}
Provided that the subsequent requirement are satisfied
\begin{itemize}
    \item Let $\delta_c$ be defined in Definition~\ref{def:delta_c:x}.
    \item Let $M = \exp(O(R^2 + \log n))$
    \item Let $R \geq 4$.
    \item Let $x_t \in \R^{d^2}, x_{t+1} \in \R^{d^2}$ satisfy $\| x_{t} \|_2 \leq R$ and $\| x_{t+1} \|_2 \leq R$
    \item Let $\A \in \R^{n^2 \times d^2}$
    \item Let $\max_{j_1 \in [n]} \| \A_{[j_1],*} (x_{t}-x_{t+1}) \|_{\infty} < 0.01$
    \item Let $\max_{j_1 \in [n]}\| \A_{[j_1],*} \| \leq R$
    \item Let $\max_{j_1 \in [n]} \| b_{[j_1]} \|_2 \leq 1$
    \item Let $M = \exp(O(R^2 + \log n))$.
\end{itemize}
We have
\begin{align*}
   \| \delta_c \|_2 \leq M \cdot \| x_{t+1} - x_t \|_2
\end{align*}
\end{lemma}
\begin{proof}
    We have 
    \begin{align*}
        D(x_{t+1})^{-1} \cdot \exp(\A x_{t+1}) -   b  = (  D(x_{t})^{-1} \cdot \exp(\A x_{t}) -(b-\delta_c) ) \circ \{ -1 , +1 \}^{n^2}
    \end{align*}
    Since all solutions have the same norm, we can simplify the problem by focusing on a single solution, namely the one presented below.
    \begin{align*}
         \underbrace{ D(x_{t+1})^{-1} }_{n^2 \times n^2 \mathrm{~matrix}}
         \cdot
         \underbrace{ \exp(\A x_{t+1}) }_{n^2 \times 1 \mathrm{~vector}} -  \underbrace{ b }_{n^2 \times 1 \mathrm{~vector}}  = D(x_{t})^{-1}\exp(\A x_{t}) - (b-\delta_c)
    \end{align*}

    One possible choice for $\delta_c \in \R^{n^2}$ is
    \begin{align}\label{eq:delta_c}
     \delta_c = D(x_{t+1})^{-1}\cdot \exp(\A x_{t+1})-D(x_{t})^{-1}\exp(\A x_t)
    \end{align}

In particular, we have
\begin{align}\label{eq:delta_c_j}
   \underbrace{ (\delta_c)_{[j]} }_{n \times 1 \mathrm{~vector}} =
    \alpha(x_{t+1})^{-1}_j u(x_{t+1})_j - \alpha(x_t)^{-1}_j u(x_t)_j
\end{align}

    Thus,
    \begin{align*}
        \| \delta_c\|_2^2
        = & ~ \| D(x_{t+1})^{-1}\cdot \exp(\A x_{t+1})-D(x_{t})^{-1}\exp(\A x_t) \|_2^2 \\
        = & ~ \sum_{j=1}^n \| \alpha(x_{t+1})^{-1}_j u(x_{t+1})_j - \alpha(x_t)^{-1}_j u(x_t)_j \|_2^2 \\ 
        = & ~ \sum_{j=1}^n \| c(x_{t+1}) - c(x_{t}) \|_2^2 \\
        \leq & ~ \sum_{j=1}^n (4 \beta^{-2} n^{1.5} R)^2 \exp(2R^2) \cdot \| x_{t+1} - x_t \|_2^2 \\ 
        \leq & ~ \sum_{j=1}^n (4  n^{1.5} R \exp(4R^2))^2 \cdot \| x_{t+1} - x_t \|_2^2 \\ 
        \leq & ~  ( 4 n^{2} R \exp(4R^2))^2 \cdot \| x_{t} - x_{t+1} \|_2 \\ 
        \leq & ~ M^2 \cdot \| x_{t+1} - x_t \|_2^2   
    \end{align*}
    where the first step is based on Eq.~\eqref{eq:delta_c}, 
    the 2nd step uses Eq.~\eqref{eq:delta_c_j}, 
    the 3rd step is due to Definition~\ref{def:c}, 
    the 4th step is based on Lemma~\ref{lem:lipschitz_c:x},
    the 5th step is because of Lemma~\ref{lem:lower_bound:beta},
    the 6th step follows from simple algebra,
    the 7th is from the definition of $M$ in Lemma statement.

Thus, we have shown that
\begin{align*}
\| \delta_c \|_2 \leq M \cdot \| x_{t+1} - x_t \|_2
\end{align*}
        
The proof is complete.
\end{proof}

\subsection{For the \texorpdfstring{$A$}{} case}\label{sec:application_normalized:A}
\begin{definition}\label{def:delta_c:A}
We define $\delta_c \in \R^{n^2}$ be to the vector satisfies 
\begin{align*}
   \|  D(\A_{t+1})^{-1}\exp(\A_{t+1}) -  b \|_2^2 = \|   D(\A_{t})^{-1}\exp(\A_{t}) -  (b-\delta_c) \|_2^2
\end{align*}
\end{definition}

\begin{lemma} \label{lem:upper_bound_normalized:A}
Provided that the subsequent requirement are satisfied
\begin{itemize}
    \item Let $\delta_c$ be defined in Definition~\ref{def:delta_b:A}.
    \item Let $M = \exp(O(R^2 + \log n))$
\end{itemize}
We have
\begin{align*}
    \| \delta_c \|_2 \leq M \cdot \| A_{t+1} - A_t \|_2
\end{align*}
\end{lemma}
\subsection{Main Result}\label{sec:application_normalized:main_result}
\begin{theorem}[Learning in-context for Normalized Version]\label{thm:in_context_normalized}
Provided that the subsequent requirement are satisfied
\begin{itemize}
    \item Let $R \geq 4$.
    \item Let $x_t \in \R^{d^2}, x_{t+1} \in \R^{d^2}$ satisfy $\| x_{t} \|_2 \leq R$ and $\| x_{t+1} \|_2 \leq R$
    \item Let $\A \in \R^{n^2 \times d^2}$
    \item Let $\max_{j_1 \in [n]} \| \A_{[j_1],*} (x_{t}-x_{t+1}) \|_{\infty} < 0.01$
    \item Let $\max_{j_1 \in [n]}\| \A_{[j_1],*} \| \leq R$
    \item Let $\max_{j_1 \in [n]} \| b_{[j_1]} \|_2 \leq 1$
    \item Let $M = \exp(O(R^2 + \log n))$.
\end{itemize}
We consider the matrix formulation for attention regression (Definition~\ref{def:intro_rescaled_matrix}) problem
\begin{align*}
     \min_{x \in \R^{d^2}} \|  D(x)^{-1}\exp(\A x) -  b \|_2^2.
\end{align*}
\begin{itemize}
\item {\bf Part 1.} 
By transitioning $x_t$ to $x_{t+1}$, we are effectively addressing a fresh normalized softmax regression problem involving
\begin{align*}
      \min_{x \in \R^{d^2}} \| D(x)^{-1} \exp(\A x) -  \wt{b} \|_2^2
\end{align*}
where 
\begin{align*}
    \| \wt{b} - b \|_2 \leq M \cdot \| x_{t+1} - x_t \|_2
\end{align*}
\item {\bf Part 2.} By transitioning $A_t$ to $A_{t+1}$, we are effectively addressing a fresh normalized softmax regression problem involving
\begin{align*}
       \min_{x \in \R^{d^2}} \|D(x)^{-1} \exp(\A x) -  \hat{b} \|_2^2
\end{align*}
where 
\begin{align*}
    \| \wh{b} - b \|_2 \leq M \cdot \| A_{t+1} - A_t \|
\end{align*}
\end{itemize}
\end{theorem}

\begin{proof}
The proof straightforwardly derives from Lemma~\ref{lem:upper_bound_normalized:x} and Lemma~\ref{lem:upper_bound_normalized:A}.
\end{proof}

\ifdefined\isarxiv
\bibliographystyle{alpha}
\bibliography{ref}
\else
\bibliography{ref}
\bibliographystyle{alpha}

\fi

\newpage
\onecolumn
\appendix




\end{document}